\documentclass{article}

\usepackage[nonatbib,final]{neurips_2021}

\usepackage[utf8]{inputenc} % allow utf-8 input
\usepackage[T1]{fontenc}    % use 8-bit T1 fonts
\usepackage{hyperref}       % hyperlinks
\usepackage{url}            % simple URL typesetting
\usepackage{booktabs}       % professional-quality tables
\usepackage{amsfonts}       % blackboard math symbols
\usepackage{nicefrac}       % compact symbols for 1/2, etc.
\usepackage{microtype}      % microtypography
\usepackage{xcolor}         % colors

\usepackage{comment}
\usepackage[ruled,vlined]{algorithm2e}
\usepackage{mathtools}
\usepackage{multirow}
\usepackage{amsthm}
\usepackage[colorinlistoftodos]{todonotes}

\DeclarePairedDelimiter\norm{\lVert}{\rVert}

\title{Adversarial Feature Desensitization}

\author{%
  Pouya Bashivan$^{1,2,*}$ \\
   \And
   Reza Bayat$^{2}$
   \And
   Adam Ibrahim$^{2}$
   \And
   Kartik Ahuja$^{2}$
   \And
   Mojtaba Faramarzi$^{2}$
   \And
   Touraj Laleh$^{2}$
   \And
   Blake Richards$^{1,2}$ \\
   \AND
   Irina Rish$^{2,*}$ \\
   \\
   $^1$ McGill University, Montreal, Canada \\ 
   $^2$ MILA, Université de Montréal, Montreal, Canada \\
   *Correspondence to: \texttt{\{bashivap,irina.rish\}@mila.quebec} \\
}

\begin{document}

\maketitle

\begin{abstract}
%  Neural networks are known to be vulnerable to adversarial attacks -- slight but carefully constructed perturbations of the inputs which drastically impair the model's performance. Various defense methods have been proposed to improve the robustness of neural net models against adversarial attacks. However, robust models trained with these methods remain vulnerable to other forms of attacks, or even to the same attacks with slightly larger perturbations. Here we view the adversarial robustness problem through the lens of domain generalization and propose a novel way to improve neural network robustness against adversarial attacks by minimizing the divergence between the natural and adversarial representations. Our proposed method, called \emph{Adversarial Feature Desensitization (AFD)}, encourages the model to learn representations invariant under adversarial perturbations via an adversarial game between a feature-learning network and an domain discriminator network that is trained to distinguish between the clean and adversarial inputs. 
%  Empirical results on several benchmarks validate the effectiveness of our proposed approach in learning robust representations that generalize to novel and familiar attacks across a wide range of attack types and attack strengths. These results indicate that reducing feature sensitivity to adversarial perturbations is a promising approach towards improving the robustness of deep models. 
 %Our code will be available at [URL] upon publication. 
 
 Neural networks are known to be vulnerable to adversarial attacks -- slight but carefully constructed perturbations of the inputs which can drastically impair the network's performance. Many defense methods have been proposed for improving robustness of  deep networks by training them on adversarially perturbed inputs. However, these models often remain vulnerable to new types of attacks not seen during training, and even to slightly stronger versions of previously seen  attacks. In this work, we propose a novel approach to  adversarial robustness,  which builds upon the insights from the  domain adaptation field.
Our method, called \emph{Adversarial Feature Desensitization (AFD)}, aims at learning  features that are invariant towards adversarial perturbations of the inputs. This is achieved through a game where we learn features that are both predictive and robust (insensitive to adversarial attacks), i.e. cannot be used to discriminate between natural and adversarial data.
Empirical results on several benchmarks  demonstrate the effectiveness of the proposed approach against a wide range of attack types and attack strengths. Our code is available at \url{https://github.com/BashivanLab/afd}.
\end{abstract}

\vspace{-0.05in}
\section{Introduction}
\vspace{-0.05in}
When training a classifier, it is common to assume that the training and test samples are drawn from the same underlying distribution. In adversarial machine learning, however, this assumption is intentionally violated by using the classifier itself to perturb the samples from the original (natural) data distribution towards a new distribution over which the classifier's error rate is increased \cite{Szegedy2013}. As expected, when tested on such adversarially generated input distribution, the classifier severely underperforms. 
%[Motivation] 
To date, various methods have been proposed to defend the neural networks against adversarial attacks \cite{madry2017towards,Athalye2018}, additive noise patterns and corruptions \cite{Hendrycks2019,Hendrycks2019pre,Rusak2020}, and transformations \cite{Engstrom2019}. Among these methods, two of the most successful adversarial defense methods to date are adversarial training \cite{madry2017towards}, which trains the neural network with examples that are perturbed to maximize the loss on the target model, and TRADES \cite{Zhang2019}, which regularizes the classifier to push the decision boundary away from the data. % (Figure \ref{fig_comp_graphics}b). 
While past adversarial defence methods have successfully improved the neural network robustness against adversarial examples, it has also been shown that these robust networks remain susceptible to even slightly larger adversarial perturbations or other forms of attacks \cite{Goodfellow2018ensemble,schott2018towards,Sitawarin2020}. 

% Despite the recent progress in deep learning that allowed neural networks to achieve  a near human-level performance across a wild range of fields \citep{he2016deep,mnih2015human,silver2017mastering,vinyals2019grandmaster}, systems deploying Deep Networks still face a number of important open challenges. Most notably, deep networks are highly vulnerable to imperceptible noise known as \emph{adversarial attacks} \cite{Szegedy2013}. 

% It has been shown that many adversarial perturbations that are often small in magnitude lead to large deviations in the high-level features of deep neural networks \cite{liao2018defense,Yoon2019}. 
% Previous work \cite{ilyas2019adversarial} demonstrated that adversarial patterns often rely on specific learned features which generalize even on large datasets such as ImageNet \cite{deng2009imagenet}. Although, it has been shown that non-feature adversarial examples can also be constructed by designing a suitable optimization goal <<ref to distill publication>>. 

In this paper, we propose to view the problem of adversarial robustness through the lens of domain adaptation, and to consider distributions of natural and adversarial images as distinct input domains that a classifier is expected to perform well on. We then focus our attention on learning features that are invariant under such domain shifts. Building upon domain adaptation literature \cite{Ben-David2010}, 
%Instead of estimating the divergence between the two distributions using common measures such as Kullback-Leibler or L1, 
we use the classification-based $\mathcal{H}\Delta\mathcal{H}$-divergence
to quantify the distance between the natural and adversarial  domains. 
%In our work, we use a domain discriminator trained on the output of a feature extractor network (shared with the task classifier) to classify between the two domains (Figure-\ref{fig_comp_graphics}a). 
The theory of domain adaptation allows us to formulate a bound on the adversarial classification error (i.e. the error under the distribution of adversarial examples) in terms of the classification error on natural images and the divergence between the natural and adversarial features.

We further propose an algorithm for %variational
minimizing the adversarial error using this bound. For this, we train a classifier and a domain discriminator to respectively minimize their losses on the label classification and domain discrimination tasks. The feature extractor is trained to minimize the label classifier's loss and maximise the discriminator's loss. In this way, the feature extractor network is encouraged to learn features that are both predictive for the classification task and insensitive to the adversarial attacks. The proposed setup is conceptually similar to prior work in adversarial domain adaptation \cite{Ganin2015,Tzeng2017}, where domain-invariant features are learned through an adversarial game between the domain discriminator and a feature extractor network. 
% we propose \emph{to learn robust representations via an adversarial game between two agents that augments the regular training of the classifier on natural samples. The adversarial game consists of alternate training of a domain discriminator which estimate the divergence between the distribution of natural and adversarial representations and a feature extractor which adversarially minimizes this distance} (Figure \ref{fig_comp_graphics}b).
%i) an attacker that searches for performance-degrading perturbations given the embedding function and ii) a discriminator function that distinguishes between the clean and perturbed inputs from their high-level representations.} The parameters of the embedding and the adversarial discriminator functions are then tuned via an adversarial game between the two (Figure \ref{fig_comp_graphics}b). 

This setup is similar to the adversarial learning paradigm widely used in image generation and transformation \cite{goodfellow2014generative,Karras2018,Zhu2017}, unsupervised and semi-supervised learning \cite{Miyato2018}, video prediction \cite{Mathieu2016, Lee2018}, active learning \cite{Sinha2019}, and continual learning \cite{ebrahimi2020adversarial}.
Some prior work have also considered adversarial learning to tackle the problem of adversarial examples \cite{wang2019direct, matyasko2018improved,Chan2020,chan2020thinks}. These methods used generative models to learn the distribution of the adversarial images\cite{wang2019direct,matyasko2018improved}, or to learn the distribution of input gradients\cite{Chan2020,chan2020thinks}.  Unlike our method which learns a discriminator function between distributions of adversarial and natural features and updates the feature extractor to reduce the discriminability of those distributions.

The main contributions of this work are as follows:
\begin{itemize}
    \item We apply domain-adaptation theory to the problem of adversarial robustness; this allows to bound the adversarial error
    %robustness against adversarial attacks and show how the theoretical domain adaptation work could be adapted for this problem.
  % Using this framework, we introduce a bound on the adversarial error of the classifier
   in terms of the error on the natural inputs and the divergence between the feature (representation) distributions of adversarial and natural domains.
    
    \item  Aiming to minimize this bound, we propose a method which learns  adversarially robust features
    %through an adversarial game between a feature learning function and a domain discriminator that distinguishes between the natural and adversarial representations. 
   %through a game where we learn features 
   that are both predictive and  insensitive to adversarial attacks, i.e. cannot be used to discriminate between natural and adversarial data.
    % \item We theoretically show that our proposed adversarial approach leads to a flat loss function in the vicinity of the training samples, thereby making the overall representation more stable against adversarial attacks.
    \item We empirically demonstrate the effectiveness of the proposed method in learning robust models against a wide range of attack types and attack strengths, and show that our proposed approach often significantly outperforms most previous defense methods. 
    
\end{itemize}

\vspace{-0.05in}
\section{Related Work}
\vspace{-0.05in}
% \todo{discuss IRM, self-supervised etc other examples of 'robust' representation}
There is an extensive literature on mitigating susceptibility to adversarial perturbations \cite{madry2017towards,Zhang2019,Dong2020ADT,Zhang2021geo,Bai2021,Gowal2020,Carmon2019}. Adversarial training \cite{madry2017towards} is one of the earliest successful attempts to improve robustness of the learned representations to potential perturbations to the input pattern by solving a min-max optimization problem. 
TRADES \cite{Zhang2019} adds a regularization term to the cross-entropy loss which penalizes the network for assigning different labels to natural images and their corresponding perturbed images. 
\cite{Qin2019} proposed an additional regularization term (local linearity regularizer) that encourages the classification loss to behave linearly around the training examples. \cite{Wu2019,Sriramanan2020guided} proposed to regularize the flatness of the loss to improve adversarial robustness. 

Our work is closely related to the domain adaptation literature in which adversarial optimization has recently gained much attention \cite{Ganin2015,Liu2019,Tzeng2017}. From this viewpoint one could consider the clean and perturbed inputs as two distinct domains for which a network aims to learn an invariant feature set. Although in our setting, i) the perturbed domain continuously evolves while the parameters of the feature network are tuned; ii) unlike the usual setting in domain-adaptation problems, here we have access to the labels associated with some samples from the perturbed (target) domain. Recent work\cite{Song2019} regularized the network to have similar logit values in response to clean and perturbed inputs and showed that this additional term leads to better robust generalization to unseen perturbations. Related to this, Adversarial Logit Pairing \cite{Kannan2018} increases robustness by directly matching the logits for clean and adversarial inputs. JARN \cite{Chan2020}
Another line of work is on developing certified defenses which consist of methods with provable bounds over which the network is \emph{certified} to operate robustly \cite{Zhang2019robust,Zhai2020,Cohen2019}. While these approaches provide a sense of guarantee about the proposed defenses, they are usually prohibitively expensive to train, drastically reduce the performance of the network on natural images, and the empirical robustness gained against standard attacks is low.

\begin{figure}[t]
\centering
\raisebox{-0.5\height}{\includegraphics[width=.25\linewidth]{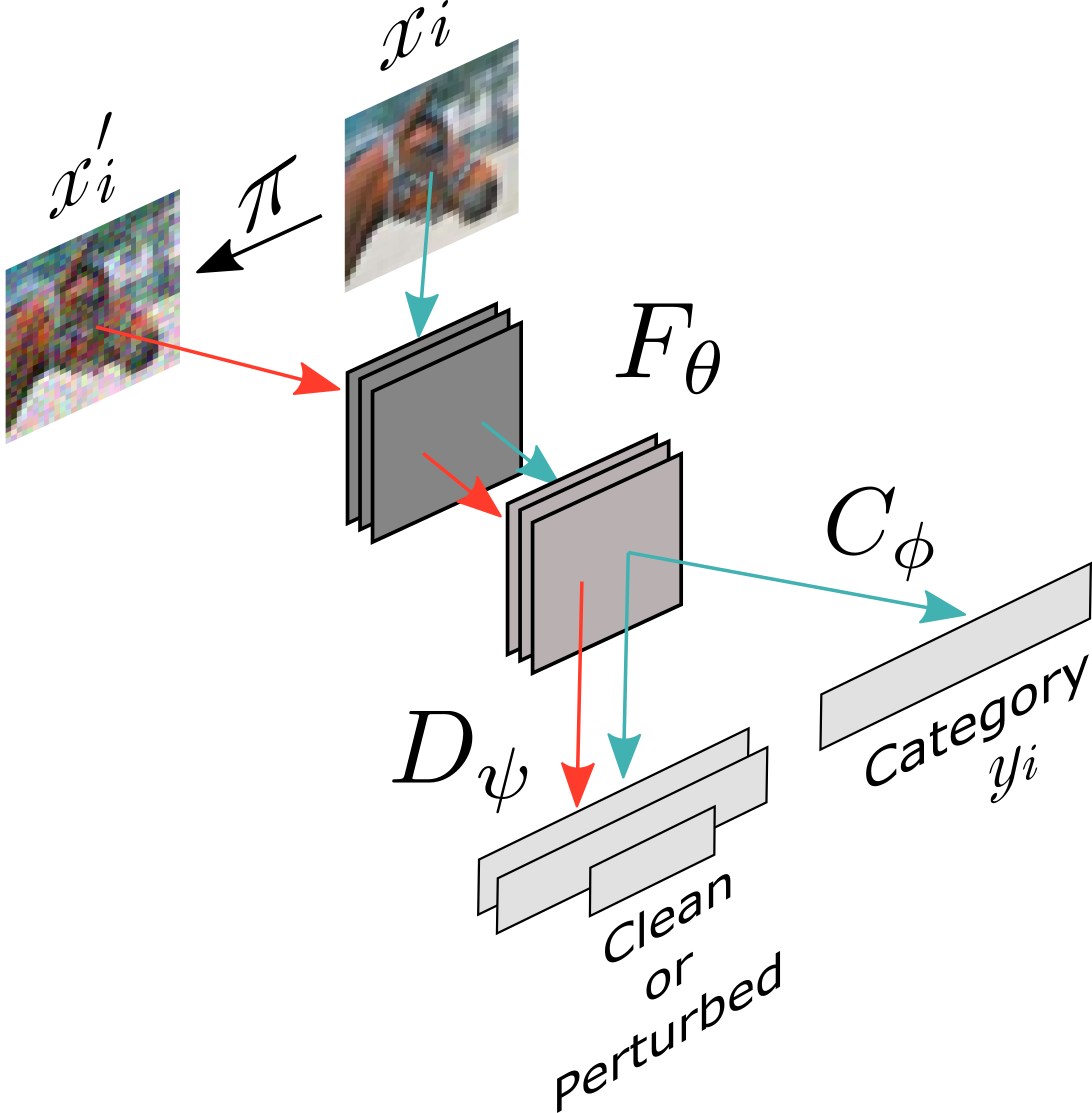}}
\hspace{5mm}
\raisebox{-0.5\height}{\includegraphics[width=.65\linewidth]{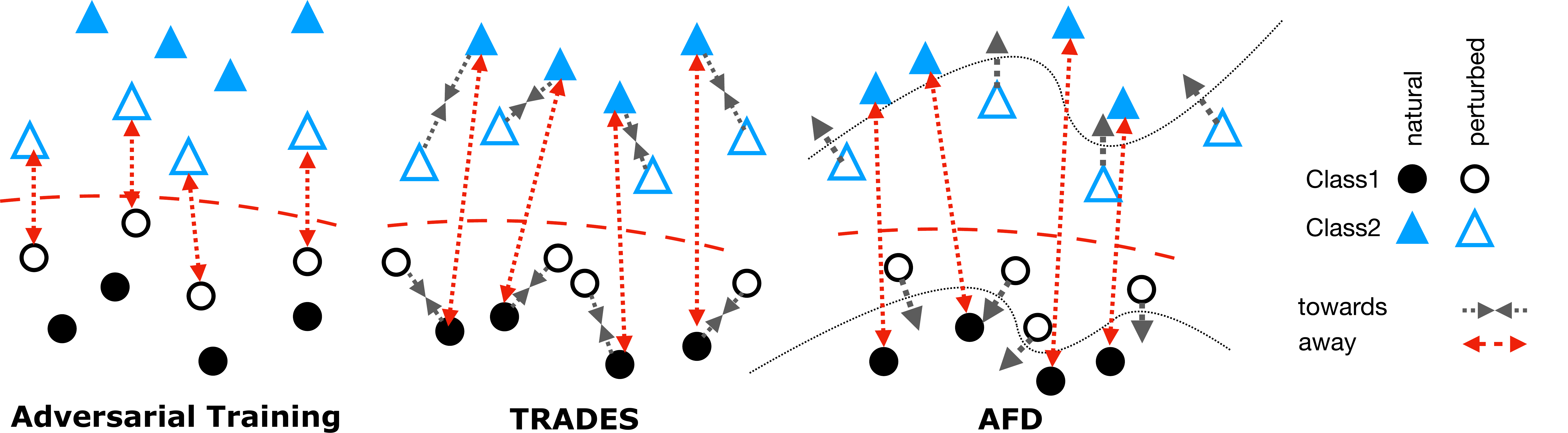}}
\vspace{-0.12in}
\centerline{(a) \hspace{2in} (b) }
\caption{  (a) Overview of the proposed AFD approach;  (b) Visual comparison of adversarial robustness methods (Adversarial training \cite{madry2017towards}, TRADES \cite{Zhang2019}, and AFD). The dashed red and dotted black lines correspond to the decision boundary of classes and the domain discriminator respectively.}
\label{fig_comp_graphics}
\vspace{-0.05in}
\end{figure}

\vspace{-0.05in}
\section{Our approach}
\vspace{-0.05in}
We will now make a connection between the domain adaptation and adversarial robustness, and build upon this connection to develop an approach for improving  the network's robustness against adversarial attacks.

\vspace{-0.05in}
\subsection{Preliminaries}
\vspace{-0.05in}
Let  $F_{\theta}(x): \mathcal{X}  \rightarrow \mathcal{Z}$, where $\mathcal{X} \subseteq \mathbb{R}^n$, $\mathcal{Z} \subseteq \mathbb{R}^m$,  be a {\em feature extractor} (e.g. a neural network with parameters $\theta$) mapping the input $x \in \mathcal{X}$  into the feature vector (representation) $z \in \mathcal{Z}$, and let  $C_{\phi}: \mathcal{Z} \rightarrow \mathcal{Y}$, where  $\mathcal{Y} = \{1, \dots, K\}$ are the class labels, be a \textit{classifier}, with parameters $\phi$ (e.g., the last linear layer of a neural network plus the softmax function, on top of the extracted features). 
% The likelihood of each class $i$ from a set of $N_c$ classes,  $C=\{1,...,N_c\}$, given the input $x$, is computed as follows:
% $l_i(x)=softmax\big(Dc_{\phi}(E_{\theta}(x))\big)_i, i\in C$. 

\textbf{Adversarial attack}: Let  $\pi(x, \epsilon)$ denote  a perturbation function (an adversarial attack) which, for a given $(x, y)\in \mathcal{X}\times \mathcal{Y}$, generates a perturbed sample $x^\prime \in \mathcal{B}(x,\epsilon)$ within the $\epsilon$-neighborhood of $x$, $\mathcal{B}(x, \,\epsilon)=\{x^\prime\in\mathcal{X}: \norm{x^\prime-x}{}<\epsilon\}$, by solving the following maximization problem 
% \begin{equation}
% \label{eq_attack}
%     \forall x\in \mathcal{X}:  \pi(x,\, \epsilon)=x^\prime\in \mathcal{B}(x,\,\epsilon);\;
%     \mathcal{B}(x, \,\epsilon)=\{x^\prime\in\mathcal{X}: \norm{x^\prime-x}{}<\epsilon\},
% \end{equation}
% and
\begin{equation}
\label{eq_attack}
     \max_{t \in \mathcal{B}(x,\epsilon)}\mathcal{L}(C_\phi(F_\theta(t)), y),
    %\mathcal{L}(C_\phi(F_\theta(x')), y),
\end{equation}
where $\mathcal{L}$ is the task classification loss function. In practice, however, the perturbed sample $x'$ found by an attacker is typically an approximate rather than the exact  solution to this maximization problem.  

%Let $\mathcal{D}_\mathcal{X}$ be a probability distribution over the instances in $\mathcal{X}$, which we call a {\em natural} data distribution. The attack induces a distribution $\mathcal{D}_\mathcal{X}$
%In adversarial robustness setting, we consider two domains --  the natural and the adversarial domains, corresponding to the target and source in domain adaptation. We denote by $\mathcal{D}_\mathcal{X}$ and $\mathcal{D}_{\mathcal{X}^\prime}$ the natural and adversarial distributions of instances, respectively, and by $\mathcal{D}_\mathcal{Z}$ and $\mathcal{D}_\mathcal{Z'}$ their corresponding induced distributions over the feature space $\mathcal{Z}$. As in domain adaptation, we assume that $f: \mathcal{X} \rightarrow \mathcal{Y}$ is a labeling function common to both domains.

% \ir{Maybe these definitions can be moved to the appendix? Also, the set of $H \delta H$  is defined in eq. (3) but NOT used in eq. (4) - maybe remove (3)?}
In order to characterize the distance between the natural and adversarial data distributions, the following notion of distance between  two probability distributions, defined in  \cite{Ben-David2010,Ganin2015}, will be used later to make a connection with domain adaptation theory.

\textbf{$\mathcal{H}\Delta\mathcal{H}$-distance:} Let  $\mathcal{H}$ be a set of binary classifiers (hypotheses), called a hypothesis space; then the symmetric difference hypothesis space $\mathcal{H}\Delta\mathcal{H}$ defines the set of hypotheses that capture the disagreements between two hypotheses in $\mathcal{H}$, as in  \cite{Ben-David2010}:
\begin{equation}
    g\in\mathcal{H}\Delta\mathcal{H} \Longleftrightarrow g(x) = h(x) \oplus h^\prime(x) \text{} \quad \text{for some } h, h^\prime \in \mathcal{H},
\end{equation}
where $\oplus$ denotes the XOR function. 
Then the  $\mathcal{H}\Delta\mathcal{H}$-distance \cite{Ben-David2010,Ganin2015}   between two data distributions (domains) $\mathcal{S}$ and $\mathcal{T}$, with respect to the hypothesis space $\mathcal{H}$, is defined as:
% \begin{equation}
%     d_{\mathcal{H}\Delta\mathcal{H}}(\mathcal{S}, \mathcal{T})=2 \underset{h, h^\prime\in\mathcal{H}}{\textrm{ sup}}\vert P_{x\sim\mathcal{S}}\big[h(x)\neq h^\prime(x)\big] - P_{x\sim\mathcal{T}}\big[h(x)\neq h^\prime(x)\big]\vert.
% \end{equation}

\begin{equation}
    d_{\mathcal{H}\Delta\mathcal{H}}(\mathcal{S}, \mathcal{T})=2 \underset{h \in\mathcal{H}\Delta\mathcal{H}}{\textrm{ sup}}\vert P_{x\sim\mathcal{S}}\big[h(x)= 1\big] - P_{x\sim\mathcal{T}}\big[h(x)= 1\big]\vert.
\end{equation}

This equation turns into an inequation when the supremum is taken over the hypothesis space $\mathcal{H}$ instead of $\mathcal{H}\Delta\mathcal{H}$ \cite{Ganin2015}.
% \begin{equation}
%     d_{\mathcal{H}\Delta\mathcal{H}}(\mathcal{D}_\mathcal{Z}, \mathcal{D}_{\mathcal{Z}}') \leq
%      2 \underset{h \in\mathcal{H}}{\textrm{ sup}}\vert P_{z\sim\mathcal{\mathcal{D}_\mathcal{Z}}}\big[h(z)= -1\big] - P_{z\sim\mathcal{\mathcal{D}_{\mathcal{Z}}'}}\big[h(z)= -1\big]\vert
% \end{equation}

\subsection{A Domain Adaptation View of Adversarial Robustness}
\label{sec_da}
%We will now  restate some of the existing results on  domain adaptation (cross-domain generalization), in applicationto the   adversarial robustness setting.
%A {\em domain} is defined as a pair $\langle\mathcal{D}, f\rangle$ consisting of a data distribution $\mathcal{D}$ on inputs $\mathcal{X}$ and a labeling function $f: \mathcal{X} \rightarrow \mathcal{Y}$. 

A {\em domain} is defined as a data distribution $\mathcal{D}$ on the set of  inputs $\mathcal{X}$ \cite{Ben-David2007}. 
%In adversarial robustness setting,  we consider the classification tasks $\mathcal{X}\rightarrow\mathcal{Y}$ where the input $x$ is sampled from the natural distribution $\mathcal{D}_\mathcal{X}$ or the adversarial distribution $\mathcal{D}_{\mathcal{X}^\prime}$ with their respective feature distributions $\mathcal{D_{\mathcal{Z}}}$ and $\mathcal{D}_{\mathcal{Z}^\prime}$.
 In the adversarial robustness setting, we consider two domains --  the natural and the adversarial domains, corresponding respectively to the source and target domains in domain adaptation. We denote by $\mathcal{D}_\mathcal{X}$ and $\mathcal{D}'_{\mathcal{X}}$ the natural and adversarial distributions of input instances respectively and by $\mathcal{D}_\mathcal{Z}$ and $\mathcal{D}'_\mathcal{Z}$ their corresponding induced distributions over the feature space $\mathcal{Z}$. As in domain adaptation, we assume that $f: \mathcal{X} \rightarrow \mathcal{Y}$ is a labeling function common to both domains.   The expected classification error $\epsilon_\mathcal{Z}$ of the classifier $C_\phi$ over  $\mathcal{D}_\mathcal{Z}$  is defined as the probability that the classifier $C_\phi$ disagrees with the function $\tilde{f}$: 
% \begin{equation}
%     \epsilon_\mathcal{Z}(C_\phi) =  E_{z\sim\mathcal{D}_{\mathcal{Z}}}\Big[E_{y\sim\tilde{f}(z)}\big[y\neq C_\phi(z)\big]\Big],
% \end{equation}
\begin{equation}
    \epsilon_\mathcal{Z}(C_\phi) =  E_{z\sim\mathcal{D}_{\mathcal{Z}}}\big[y\neq C_\phi(z)\big],
\end{equation}
where $\tilde{f}: \mathcal{Z} \rightarrow \mathcal{Y}$ is a mapping from the  features to the class label such that
$f(x) = \tilde{f}(F_\theta(x))$.
We similarly define $\epsilon'_{\mathcal{Z}}$ as the expected error of  $C_\phi$ over   $\mathcal{D}_{\mathcal{Z}^\prime}$.
Using theorem 2 from \cite{Ben-David2010} that relates the source and  the target domain errors, we get an upper bound on the expected adversarial error $\epsilon_{\mathcal{Z}}'$  as:
\begin{equation}
\label{eq_hdh_bound}
    \epsilon'_{\mathcal{Z}}(h) \leq \epsilon_{\mathcal{Z}}(h) + \frac{1}{2}d_{\mathcal{H}\Delta\mathcal{H}}(\mathcal{D}_\mathcal{Z}, \mathcal{D}'_{\mathcal{Z}}) + c,
\end{equation}
where $c$  is a constant term w.r.t. $h$. Eq. \ref{eq_hdh_bound} essentially gives a bound   on the adversarial error $\epsilon_{\mathcal{Z}}'$ in terms of the natural error $\epsilon_{\mathcal{Z}}$ and a divergence $d_{\mathcal{H}\Delta\mathcal{H}}$ between the natural and adversarial domains with respect to their induced representation distributions $\mathcal{D}_\mathcal{Z}$ and $\mathcal{D}_\mathcal{Z}'$. In the next section, we  will describe an algorithm for improving adversarial robustness of a model by iteratively estimating and minimizing these two components of the error bound.

\vspace{-0.05in}
\subsection{Adversarial Feature Desensitization}
\vspace{-0.05in}
%In this section, we  describe an algorithm for improving adversarial robustness of a model by iteratively estimating and minimizing the $\mathcal{H}\Delta\mathcal{H}$-distance between the natural and adversarial representation distributions. %As noted before,   adversarial examples $x^\prime=\pi(x, \epsilon)$  maximizes the loss function in Eq. \ref{eq_attack} only approximately,  and thus the corresponding adversarial error  $\epsilon'_{\mathcal{Z}}$ is rather a  lower bound on the optimal one. Thus, defense methods such as adversarial training essentially minimize the lower bound on the adversarial error.

%As stated in Eq. \ref{eq_hdh_bound}, the upper bound on the expected adversarial error can be written in terms of the natural error, the divergence between the natural and adversarial domains, and a constant term $C$.
Based on Eq. \ref{eq_hdh_bound},  the expected adversarial error could be reduced by jointly minimizing the natural error and the divergence between the distributions of natural and adversarial representations $d_{\mathcal{H}\Delta\mathcal{H}}(\mathcal{D}_\mathcal{Z}, \mathcal{D}_{\mathcal{Z}}')$. While minimizing the natural error $\epsilon_X$ is straightforward, minimizing the cross-domain divergence requires us to estimate $d_{\mathcal{H}\Delta\mathcal{H}}(\mathcal{D}_\mathcal{Z}, \mathcal{D}_{\mathcal{Z}}')$. As was shown before \cite{Ganin2015}, training a {\em domain discriminator} $D_{\psi}$  is closely related to estimating the $d_{\mathcal{H}\Delta\mathcal{H}}(\mathcal{D}_\mathcal{Z}, \mathcal{D}_{\mathcal{Z}}')$. The domain discriminator is a classifier trained to assign a label of 1 to samples from $\mathcal{D}_\mathcal{Z}$, and -1 to samples from $\mathcal{D}_\mathcal{Z}'$. Namely, it is shown \cite{Ganin2015} that
\begin{equation}
\label{dist_est}
    d_{\mathcal{H}\Delta\mathcal{H}}(\mathcal{D}_\mathcal{Z}, \mathcal{D}_{\mathcal{Z}}') \leq
    %  2 \underset{h \in\mathcal{H}}{\textrm{ sup}}\vert P_{z\sim\mathcal{\mathcal{D}_\mathcal{Z}}}\big[h(z)= -1\big] - P_{z\sim\mathcal{\mathcal{D}_{\mathcal{Z}}'}}\big[h(z)= -1\big]\vert = \\
    2 \underset{h \in\mathcal{H}}{\textrm{ sup}}\vert \alpha_{\mathcal{D}_\mathcal{Z},\mathcal{D}'_\mathcal{Z}}(h) - 1\vert,
\end{equation}
where $\alpha_{\mathcal{D}_\mathcal{Z},\mathcal{D}'_\mathcal{Z}}(h)=P_{z\sim\mathcal{\mathcal{D}_\mathcal{Z}}}\big[h(z)= 1\big] + P_{z\sim\mathcal{\mathcal{D}_{\mathcal{Z}}'}}\big[h(z)= -1\big]$ combines the  true positives and true negatives, and is thus  maximized   by the optimal domain discriminator $h=D_\psi$. Note that, if the domain distributions $\mathcal{D}_\mathcal{Z}$ and $\mathcal{D}'_\mathcal{Z}$ are the same, then even the best choice of domain discriminator $D_\psi$ will achieve chance-level accuracy, corresponding to $\alpha_{\mathcal{D}_\mathcal{Z},\mathcal{D}'_\mathcal{Z}}(D_\psi)=1$.
{\it Our approach will aim at minimizing this estimated distance $d_{\mathcal{H}\Delta\mathcal{H}}(\mathcal{D}_\mathcal{Z}, \mathcal{D}_{\mathcal{Z}}')$ by tuning 
the feature extractor network parameters $\theta$ in the direction that pushes the distributions $\mathcal{D}_\mathcal{Z}$ and $\mathcal{D}_{\mathcal{Z}}'$ closer together.} 
In parallel, we train the domain discriminator to estimate and guide the progress of the feature extractor's tuning.

%Eq. \ref{eq_hdh_bound} implies that for a trained classifier $C_\phi$ with $\epsilon_\mathcal{Z}$, the adversarial error bound remains constant as long as the domain discriminator error does not change. 
%In practice, a classifier trained on relatively weak adversarial examples could be used to correctly classify larger attacks of the same kind or completely novel types of attacks as long as they rely on same the feature combinations as the weak attack. 
%---

We now describe the proposed approach (see Algorithm \ref{algo_afd}) which essentially involves simultaneous training of the feature extractor $F_\theta$, the task classifier $C_\phi$ and the domain discriminator $D_\psi$ (see Figure 1a)\footnote{Note that we will somewhat abuse the notation, assuming that $C_{\phi}$ and $D_{\psi}$ below correspond to the logits (last-layer output) of the corresponding networks. Also, we will use class-conditional discriminators, $D_{\psi}(F_{\theta}(x,y))$, i.e. train different domain discriminator for different label values $y$.}.
One iteration 
%(corresponding to one mini-batch of training data) 
of the training procedure consists of the following three steps. 

First, parameters of the feature extractor  $F_\theta$ and classifier $C_\phi$ are updated aiming to minimize the natural error $\epsilon_\mathcal{X}$ using the cross-entropy loss on natural inputs:
% \begin{equation}
% \label{eq_lc}
%     \mathcal{L}_{C}=-\frac{1}{m}\sum_{i=1}^{m}\log\big(P(y^{(i)}\vert x^{(i)}\big)=-\frac{1}{m}\sum_{i=1}^{m}\log\Big(\textrm{softmax}(-C_{\phi}(F_\theta(x^{(i)})))_{\,y^{(i)}}\Big)
% \end{equation}
\begin{equation}
\label{eq_lc}
    \mathcal{L}_{C}=-\frac{1}{m}\sum_{i=1}^{m}\tilde{y}_i\cdot\log\Big(\textrm{softmax}(C_{\phi}(F_\theta(x_i)))\Big),
\end{equation}
where $\tilde{y}_i$ is a one-hot encoding of the true label of the $i$-th sample $ x_i$.
% \ir{I am not sure I understand this notation for cross-entropy loss, the $ y_i$ subscript? shouldn't this be (as well as the corresponding loss in the algorithm):
% \begin{equation}
%     \mathcal{L}_{C}=-\frac{1}{m}\sum_{i=1}^{m}y_i\cdot\log\Big(\textrm{softmax}(-C_{\phi}(F_\theta(x_i)))\Big)
% \end{equation}

% where $y_i$ is a one-hot encoding of the true label of $x_i$.
% }

Next, steps two and three  essentially implement a   two-player minimax game similar to that in Generative Adversarial Networks (GAN) \cite{goodfellow2014generative}, carried out between the feature extractor network $F_\theta$ and the domain discriminator $D_\psi$, with a value function %$V(F_\theta,\,D_\psi)$ 
\vspace{0.05in}
\begin{equation}
\label{eq_gan}
 V(F_\theta,\,D_\psi)=\mathbb{E}_{p(y)}\big[\mathbb{E}_{p(x|y)}[\mathcal{S}(-D_\psi(F_\theta(x),\,y))]\big] + \mathbb{E}_{q(y)}\big[\mathbb{E}_{q(x|y)}[\mathcal{S}(D_\psi(F_\theta(x),\,y))]\big],
\end{equation}
%\vspace{0.05in}
where $\mathcal{S}$ is the softplus function. In particular, parameters  of the domain discriminator $D_\psi$ are updated to minimize the cross-entropy loss associated with discriminating natural and adversarial inputs,
maximizing $\alpha(h)$ in Eq. \ref{dist_est}.  
\begin{equation}
\label{eq_ld}
    \mathcal{L}_{D}=\frac{1}{m}\sum_{i=1}^{m}\Big[\mathcal{S}(-D_{\psi}(F_\theta(x_i), \,y_i))+\mathcal{S}(D_{\psi}(F_\theta(x^{\prime}_i), \,y_i))\Big],
\end{equation}
while the  parameters of the feature extractor function $F_\theta$ are adversarially updated to maximize the domain discriminator's loss from Eq. \ref{eq_ld} 
% \begin{equation}
% \label{eq_lf}
%     \mathcal{L}_{F}=\frac{1}{m}\sum_{i=1}^{m}\log\Big( 1-D_\psi(F_\theta(x^\prime_i), y_i)\Big)
% \end{equation}
\begin{equation}
\label{eq_lf}
    \mathcal{L}_{F}=\frac{1}{m}\sum_{i=1}^{m}\mathcal{S}(-D_{\psi}(F_\theta(x^{\prime}_i),\,y_i)).
\end{equation}
%Algorithm \ref{algo_afd} summarizes the proposed approach. 
In Figure \ref{fig_comp_graphics}b, we visually compare the learning dynamics in adversarial training, TRADES and AFD. Essentially, the adversarial training solves the classification problem by pushing the representation of adversarial examples from different classes away. TRADES regularizes the normal classification loss on the natural inputs with an additional term that encourages the representation of adversarial and natural images to match. Similar to TRADES, in AFD, the regular classification loss on natural inputs is augmented but with an adversarial game which consists of training the domain discriminator that distinguishes between the adversarial and natural inputs for each class followed by updates to the feature extractor to make the representations for natural and adversarial examples to become indistinguishable from each other. 
Notably, because the parameter update for the feature extractor network is done to maximize the domain discriminator loss and not to decrease the loss for particular adversarial examples (as is done in adversarial training or TRADES), it potentially increases the network robustness against any perturbation that could be correctly classified using the same domain discriminator. This could potentially lead to a broader form of generalization learned by the network. 

%\subsection
\noindent{\bf Discussion: Relation to Adversarial Training.}
Adversarial training minimizes the expected error on adversarial examples (the perturbed versions of the natural samples), generated by an attacker in order to maximize the classification loss. The adversarial training procedure involves a minimax optimization problem consisting of an inner maximization to find adversarial examples that maximize the classification loss and an outer minimization to find model parameters that minimize the adversarial loss. 
From the domain adaptation point of view, the inner optimization of adversarial training is equal to a sampling procedure that generates samples from the target domain. Intuitively, direct training of the classifier on samples from the target domain would be the best way to improve the accuracy in that domain (i.e. adversarial classification accuracy). However, it's important to note that the adversarial examples found through the inner optimization only approximately maximize the classification loss, and therefore the adversarial error associated with these samples only act as a lower bound on the true adversarial error and therefore the outer loop of the adversarial training method essentially minimizes a lower bound on the adversarial classification error.
In contrast to this setup, our proposed method minimizes a conservative upper bound on the adversarial error and therefore is more likely to generalize to a larger set of unseen attacks, and to stronger versions of previously seen attacks (i.e. ones that generate higher-loss samples in the  inner optimization loop).

\begin{algorithm}[t]
\label{algo_afd}
\SetAlgoLined
\KwIn{Adversarial perturbation function (attack) $\pi$, %mini-batch $B$ of size $m$, 
feature extractor  $F_\theta$,   task classifier $C_{\phi}$, domain discriminator $D_{\psi}$,  learning rates $\alpha$, $\beta$, and $\gamma$.}
\Repeat{convergence}{
 input next mini-batch $\{(x_i,\,y_i), ..., (x_m,\,y_m)\}$ 
 
 for i=1 to m: $x^\prime_i \leftarrow \pi(x_i,\, \epsilon)$ 
 
%  $\mathcal{L}_{C}=-\frac{1}{m}\sum_{i=1}^{m}\log\Big(\textrm{softmax}(-C_{\phi}(F_\theta(x_i)))_{\,y_i}\Big)$
Compute $\mathcal{L}_{C}$ according to Eq. \ref{eq_lc}
 
%  $\mathcal{L}_{D}=\frac{1}{m}\sum_{i=1}^{m}\Big[\mathcal{S}(-D_{\psi}(F_\theta(x_i), \,y_i))+\mathcal{S}(D_{\psi}(F_\theta(x^\prime_i), \,y_i))\Big]$  \%  $\mathcal{S}$ is the softplus function
Compute $\mathcal{L}_{D}$ according to Eq. \ref{eq_ld}
 
%  $\mathcal{L}_{F}=\frac{1}{m}\sum_{i=1}^{m}\mathcal{S}(-D_{\psi}(F_\theta(x^\prime_i),\,y_i))$
Compute $\mathcal{L}_{F}$ according to Eq. \ref{eq_lf}
 
 ($\theta,\,\phi) \leftarrow (\theta,\,\phi) - \alpha \nabla_{\theta,\,\phi}\mathcal{L}_{C} \quad$  \% update feature extractor and task classifier
 
 $\psi \leftarrow \psi - \beta \nabla_{\psi}\mathcal{L}_{D} \quad$ \% update domain discriminator
 
 $\theta \leftarrow \theta - \gamma \nabla_{\theta}\mathcal{L}_{F} \quad$  \% update feature extractor
 
 }
 \caption{AFD training procedure}
\end{algorithm}

% While the assumption of convergence to global optimum is a strong assumption, in practice, it is possible to derive a bound on the classifier's adversarial error in terms of its error on clean inputs and a divergence measure between the clean and perturbed representations (see \ref{sec_apndx_bound} in the appendix).   

% {\bf ?? add a sentence highlighting the novelty}

\vspace{-0.05in}
\section{Experiments}
\vspace{-0.05in}
%In this section we empirically demonstrate the effectiveness of AFD in training neural networks that are robust against a range of attack types and magnitudes.
%\vspace{-0.05in}
\subsection{Experimental setup}
\vspace{-0.05in}
\textbf{Datasets.} We validated our proposed method on several common datasets including MNIST \cite{Lecun1998}, CIFAR10, CIFAR100 \cite{krizhevsky2009cifar}, and tiny-Imagenet \cite{Huang2017tiny}. The inputs for all datasets were used in their original resolution except for tiny-Imagenet where the inputs were resized to $32\times32$ to allow the experiments to finish within reasonable time on two GPUs. 

\textbf{Adversarial attacks.} To fairly assess the generalization ability of each defense method across attack types, we tested each network on 9 well-known adversarial attacks from the literature, using existing implementations from the Foolbox \cite{rauber2017foolbox} and Advertorch \cite{ding2019advertorch} Python packages. Namely, we tested the models against different variations of the Projected Gradient Descent (PGD) \cite{madry2017towards} ($L_{\infty},\,L_2,\,L_1$), Fast Gradient Sign Method (FGSM) \cite{Goodfellow2014fgsm}, Momentum Iterative Method (MIM) \cite{Dong2018}, Decoupled Direction and Norm (DDN) \cite{Rony2019}, Deepfool \cite{Moosavi-Dezfooli2016}, C\&W \cite{Carlini2017}, and AutoAttack \cite{croce2020reliable} attacks. Also to assess the generalization in robustness across stronger adversarial attacks, for each attack we also varied the $\epsilon$ value across a wide range and validated different models on each. Specific hyperparameters used for each attack are listed in Table-\ref{table_supp_attack_settings}. 

\textbf{Feature extractor network $F_\theta$ and classifier $C_\phi$.} We used the same network architecture, ResNet18 \cite{he2016deep} for the feature extractor and classifier networks in experiments on all datasets and only increased the number of features for more challenging datasets. The number of base filters in the ResNet architecture was set to 16 for MNIST and 64 for other datasets. We used the activations before the last linear layer as the the output of the feature extractor network ($\mathcal{Z}$) and the last linear layer as the classifier network $C_\phi$. We added an activation normalization layer to the output of feature extractor network to provide normalized features to both $C_\theta$ and $D_\psi$ networks. 

\textbf{Domain discriminator network $D_\psi$.} We compared several variations of the domain discriminator architecture and evaluated its effect on robust classification on MNIST dataset (Table \ref{tab_ablation}). Overall, we found that using deeper networks for domain discriminator and adding projection discriminator layer improves the robust classification accuracy. The number of hidden units in all layers of $D_{\psi}$ were equal (64 for MNIST and 512 for other datasets). Following the common design principles in Generative Adversarial Networks literature, we used the spectral normalization \cite{Miyato2018sn} on all layers of $D_{\psi}$. In all experiments, the domain discriminator ($D_{\psi}$) consisted of three fully connected layers with Leaky ReLU nonlinearity followed by a projection discriminator layer that incorporated the labels into the adversarial discriminator through a dot product operation \cite{miyato2018cgans}. Further details of training for each experiment are listed in Table-\ref{table_supp_hyperpars}. 

\textbf{Training parameters and baselines.}  All networks including baselines were trained on an adaptive version of PGD attack \cite{croce2020reliable} that adaptively tunes the step size during the attack with virtually no computational overhead compared to standard PGD attack. We used $\epsilon=0.3$, $0.031$, and $0.016$ for MNIST, CIFAR, and Tiny-Imagenet datasets respectively.
To find the best learning rates, we randomly split the CIFAR10 train set into a train and validation sets (45000 and 5000 images in train and validation sets respectively). We then carried out a grid-search using the train-validation sets and picked the learning rates with highest validation performance. Based on this analysis, we selected the learning rate $\gamma=0.5$ for tuning the feature extractor $F_{\theta}$, and $\alpha=\beta=0.1$ for tuning the parameters in domain discriminator $D_{\psi}$, and the task classifier $C_{\phi}$. 

In all experiments we trained two versions of the AFD model, one with losses $L_D$ and $L_F$ according to Eq. \ref{eq_ld} and \ref{eq_lf} which we call AFD-DCGAN and another version where we substitute the losses with those from the Wasserstein GAN \cite{Arjovsky2017} dubbed AFD-WGAN (see Eq. \ref{eq_wgan_ld} and \ref{eq_wgan_lf} in the Appendix). We mainly compared the performance of our proposed method with two prominent defense methods, adversarial training and TRADES. We used a re-implementation of adversarial training (AT) method \cite{madry2017towards} and the official code for TRADES\footnote{\url{https://github.com/yaodongyu/TRADES.git}} \cite{Zhang2019} and denoted these results with $\dagger$ in the tables. All experiments were run on NVIDIA V100 GPUs. We used one GPU for experiments on MNIST and 2 GPUs for other datasets.

\begin{table}[th]
\centering
\caption{Comparison of adversarial accuracy against various attacks on different datasets. For $PGD_{\infty}$ attack we used $\epsilon=0.3$, $0.031$, and $0.015$ for MNIST, CIFAR10/CIFAR100, and Tiny-Imagenet datasets respectively and for C\&W attack we used $\epsilon=1$ for all datasets. $\dagger$ indicates replicated results. NT: natural training; AT: adversarial training; AFD: adversarial feature desensitization; WB: white-box attack; BB: black-box attack where the adversarial examples were produced by running the attack on the NT ResNet18 model. Numbers reported with $\mu\pm\sigma$ denote mean and std values over three independent runs with different random initialization. * RST\cite{Carmon2019} additionally uses 500K unlabeled images during training.}
\label{table_accu}
\resizebox{\textwidth}{!}{%
\begin{tabular}{c|c|c|c|c|c|c|c|c}
\toprule
\textbf{Method}     & \textbf{Dataset}      & \textbf{Network}
& \textbf{Clean}
& \textbf{$\text{PGD}_{\infty}$} (WB)
& \textbf{$\text{C\&W}_{2}$} (WB) 
& \textbf{$\text{AA}_{\infty}$} (WB) 
& \textbf{$\text{PGD}_{\infty}$} (BB) 
& \textbf{$\text{C\&W}_{2}$} (BB)
\\ \midrule
% NT$\dagger$      & \multirow{9}{*}{MNIST}   & LeNet  & 98.88     & 0       & 0.45       & 0       & 0.44       \\
% AT\cite{madry2017towards}      & \multirow{8}{*}{MNIST}   & LeNet  & 98.8     & 93.2       & 95.6    & 96.0       & 96.8       \\
% TRADES\cite{Zhang2019}  &    & LeNet  & \textbf{99.48}     & 96.07       & -         & -       & -       \\
% ATES\cite{Sitawarin2020}    &    & LeNet     & 99.11     & 94.04         & -      & -       & -      \\
% ABS\cite{schott2018towards}    &    & LeNet     & 99.0     & 13         & 34      & -       & -      \\
% Defense-GAN\cite{samangouei2018defense}      &    & ConvNet  & 99.20     & -     & -       & -       & 93.23       \\
% \midrule
NT$\dagger$      & \multirow{4}{*}{MNIST}   & RN18  & 98.84     & 0.       & 62.43   & 0.0   & 50.82       & 96.48       \\
AT\cite{madry2017towards}$\dagger$      &    & RN18  & \textbf{99.35}     & 95.66      & 92.78  & 89.99  & \textbf{98.92}       & \textbf{98.95}       \\
TRADES\cite{Zhang2019}$\dagger$  &    & RN18  & 99.14     & 94.81       & 90.08     & 88.66    & 98.5       & 98.57       \\
AFD-DCGAN      &    & RN18  & 99.24     & 95.72     & 93.78    & 88.79   & 98.65       & 98.49       \\
AFD-WGAN      &    & RN18  & 99.14     & \textbf{97.68}     & \textbf{97.68}   & \textbf{90.12}   & 98.59       & 98.71       \\
\midrule
AT\cite{madry2017towards}      & \multirow{11}{*}{CIFAR10}   & RN18  & 87.3     & 45.8       & -   & -  & 86.0       & -       \\
TRADES\cite{Zhang2019}  &    & RN18  & 84.92     & 56.61       & -    & -    & 87.60       & -       \\
% ATES\cite{Sitawarin2020}    &    & WRN-34-10     & 86.84     & 55.06         & -      & -       & -      \\
RLFAT\cite{Song2020}    &    & WRN-32-10     & 82.72     & 58.75         & -   & -  & -       & -      \\
RST+\cite{Wu2019,Carmon2019}$^*$    &    & WRN-34-10     & 89.82     & 64.86         & -    & -   & -       & -      \\
LLR\cite{Qin2019}    &    & WRN-28-8     & \textbf{86.83}     & 52.99         & -   & -   & -       & -      \\
% YOPO\cite{Zhang2019yopo}    &    & RN18     & 83.99     & 44.72         & -      & -       & -      \\
JARN\cite{Chan2020}    &    & WRN-34-10     & 84.8     & 46.7         & -    & -  & -       & -      \\
% FS\cite{zhang2019defense}      &     & WRN-28-10     & 90.00     & 70.5     & -   & -       & - \\
NT$\dagger$      &    & RN18  & 94.89     & 0.55       & 0.31    & 0.0   & 17.93       & -       \\
AT\cite{madry2017towards}$\dagger$      &    & RN18  & 85.92     & 40.07       & 40.27   & 36.14  & 85.14       & 85.84       \\
TRADES\cite{Zhang2019}$\dagger$  &    & RN18  & 81.94     & 53.3       & 40.24     & \textbf{43.48}   & 80.82       & 81.74       \\
AFD-DCGAN      &    & RN18  & \textbf{86.82}     & 44.35     & 50.93    & 34.46   & \textbf{85.73}       & \textbf{86.68}       \\
AFD-WGAN      &    & RN18  & 85.95     & \textbf{59.38}     & \textbf{62.43}    & 37.33   & 84.74       & 85.79       \\

\midrule
% RLFAT\cite{Song2020}    & \multirow{6}{*}{CIFAR100}   & WRN-32-10     & 56.70     & 31.99          & -       & -      \\
% FS\cite{zhang2019defense}      &     & WRN-28-10     & 73.9     & \textbf{47.2}     & \textbf{61.0}   & -       & - \\
NT$\dagger$      & \multirow{5}{*}{CIFAR100}    & RN18  & 76.76     & 0.01       & 0.52     & 0.02   & -       & -       \\
AT\cite{madry2017towards}$\dagger$      &    & RN18  & 56.49     & 18.54       & 17.71   & 18.30   & 56.07       & 56.42       \\
TRADES\cite{Zhang2019}$\dagger$  &    & RN18  & 60.32     & \textbf{25.11}       & 20.52    & \textbf{21.10}     & 59.62       & 60.29       \\
AFD-DCGAN      &    & RN18  & \textbf{60.95}     & 18.06    & 21.47    & 16.31   & \textbf{60.31}       & \textbf{60.86}       \\ 
AFD-WGAN      &    & RN18  & 58.87     & 22.35    & \textbf{25.33}    & 18.00   & 58.15       & 58.75       \\ 
\midrule
NT$\dagger$      & \multirow{4}{*}{Tiny-IN}   & RN18  & 58.30     & 0.3       & 0.0    & 0.0    & -       & -       \\
AT\cite{madry2017towards}$\dagger$      &    & RN18  & 43.80     & 12.62       & 4.90   & 9.48 & 41.87       & 42.86       \\
TRADES\cite{Zhang2019}$\dagger$  &    & RN18  & 37.70     & \textbf{13.26}       & 4.11     & \textbf{12.57}   & 36.26       & 36.72       \\
AFD-WGAN      &    & RN18  & \textbf{47.70}     & 11.49    & \textbf{5.90}    & 9.45   & \textbf{43.5}
& \textbf{44.69}       \\ 
\bottomrule
\end{tabular}}
\end{table}

% \begin{figure}[bth]
% \centering
% \includegraphics[width=0.8\linewidth]{ICLR21/figs/fig2.pdf}
% \vspace{-0.18in}
% \caption{Robust accuracy for different strengths of PGD-$L_\infty$ attack on different datasets.}
% \label{fig_accu_eps}
% \end{figure}

\vspace{-0.05in}
\subsection{Robust classification against nominal attacks}
\vspace{-0.05in}
We first evaluated our method against adversarial attacks under similar settings to those used during training ($\epsilon=0.3$, $0.031$, and $0.015$ for MNIST, CIFAR, and Tiny-Imagenet datasets respectively). Table \ref{table_accu} compares the robust classification performance of AFD and several other defense methods against PGD-$L_\infty$, C\&W-$L_2$ and AutoAttack white-box and black-box attacks. The black-box attacks were carried out by constructing the adversarial examples using a ResNet18 architecture trained on the natural inputs $x\sim D_\mathcal{X}$. 
Overall both versions of AFD (AFD-DCGAN and AFD-WGAN) were highly robust against all five tested attacks while maintaining a higher "Clean" accuracy (on natural data) compared to strong baseline models like TRADES and Adversarial Training. AFD-WGAN was consistently at the top on MNIST and CIFAR10 datasets. On CIFAR100 and Tiny-Imagenet, AFD performed better than or similar to Adversarial Training on all the attacks and performed better than TRADES on most of the attacks, although it was occasionally behind TRADES (on PGD-$L_\infty$ and AA white-box attacks). Analysis of feature sensitivity showed that on MNIST and CIFAR10 datasets on which AFD outperformed the other baselines by a larger margin, the features were significantly more insensitive to adversarial perturbations and over a larger range of attack strengths (Figure-\ref{fig_supp_emb_sens}). In addition to these tests, we also evaluated the AFD model against transfer black-box attacks from Adversarial Training and TRADES models which further demonstrated AFD's higher robustness to those attacks too (Table-\ref{tab_supp_trasnfer_bb}). 

\begin{figure}[tbh]
\centering
\includegraphics[width=0.245\linewidth]{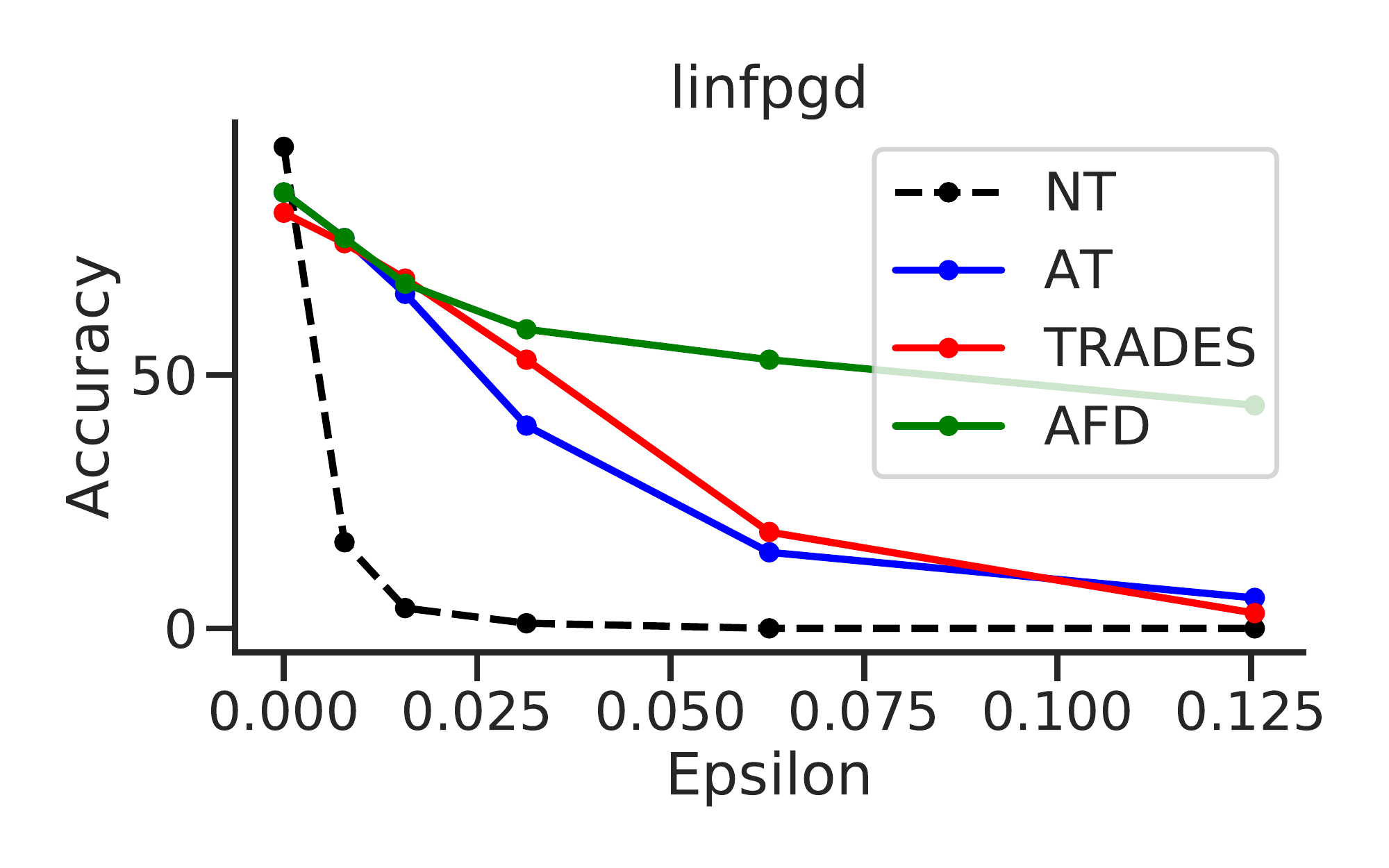}
\includegraphics[width=0.245\linewidth]{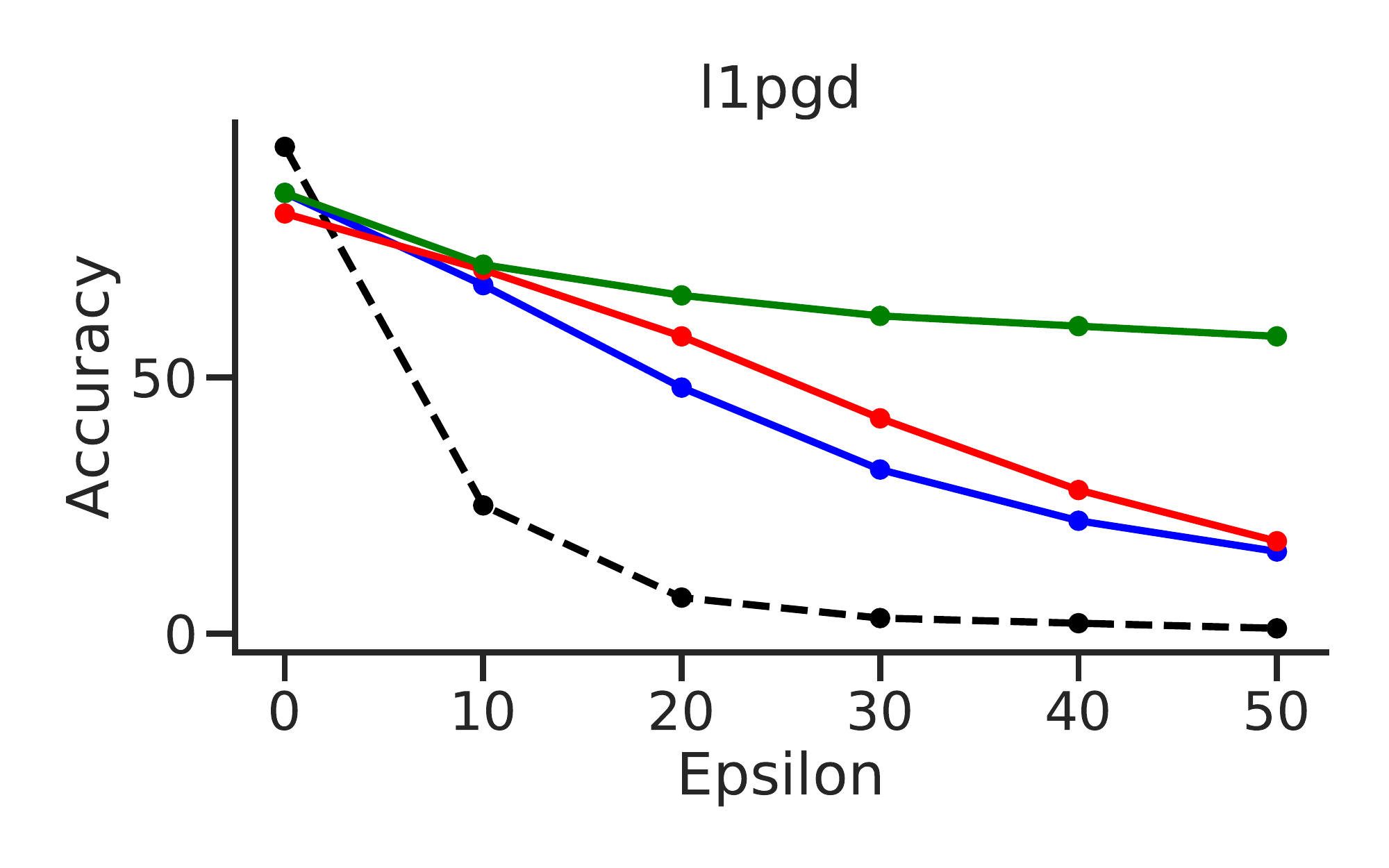}
\includegraphics[width=0.245\linewidth]{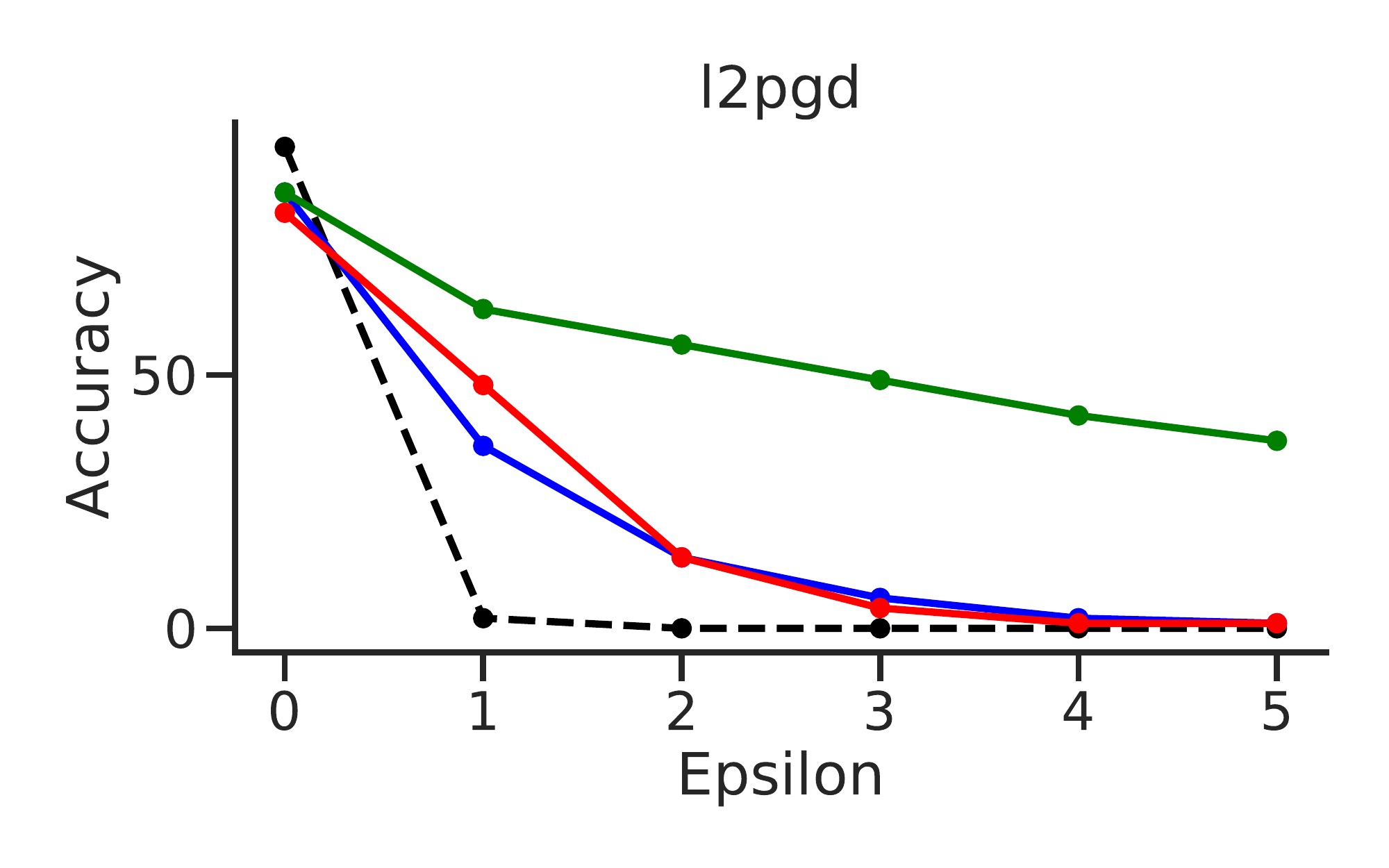}
\includegraphics[width=0.245\linewidth]{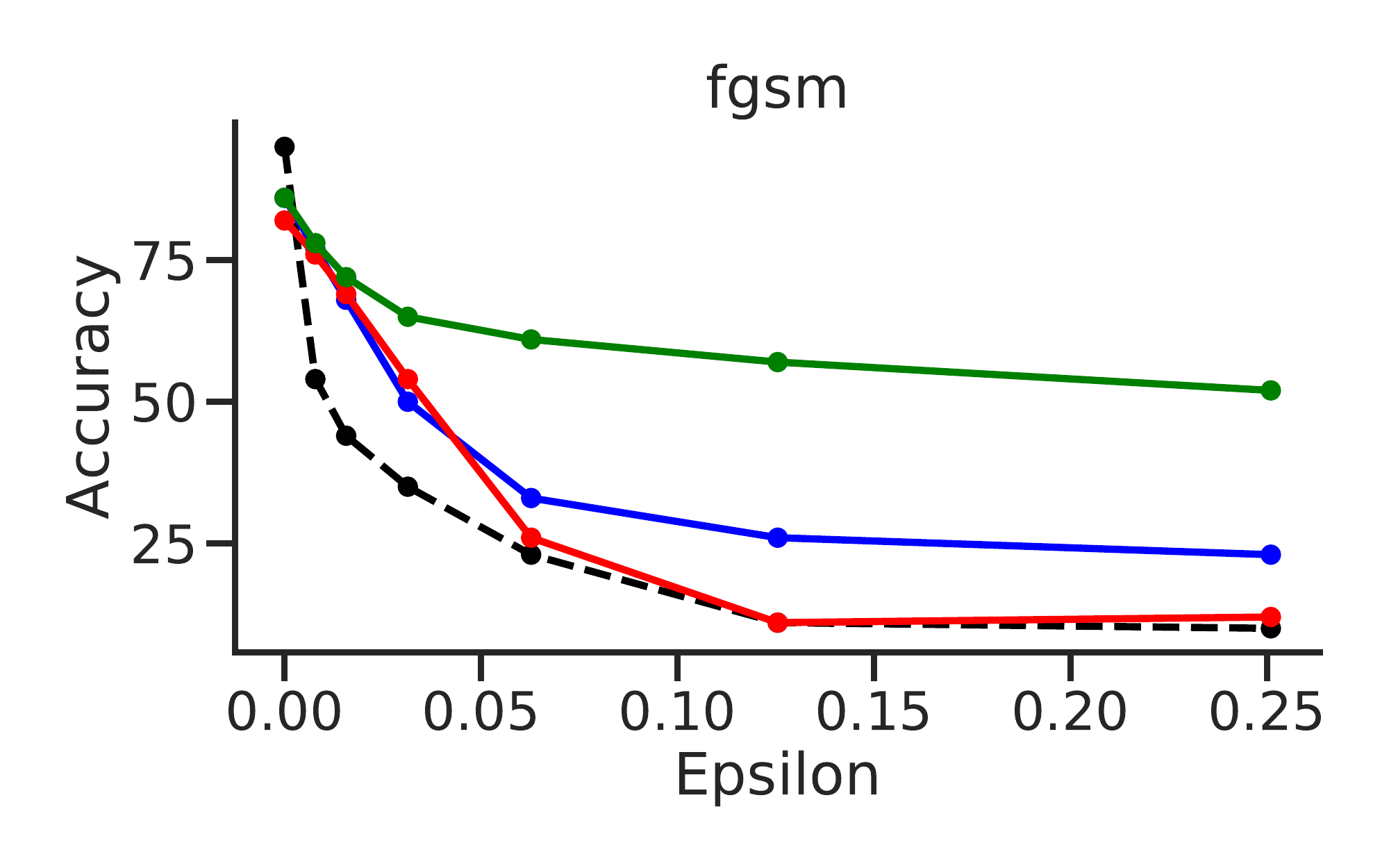}
\includegraphics[width=0.245\linewidth]{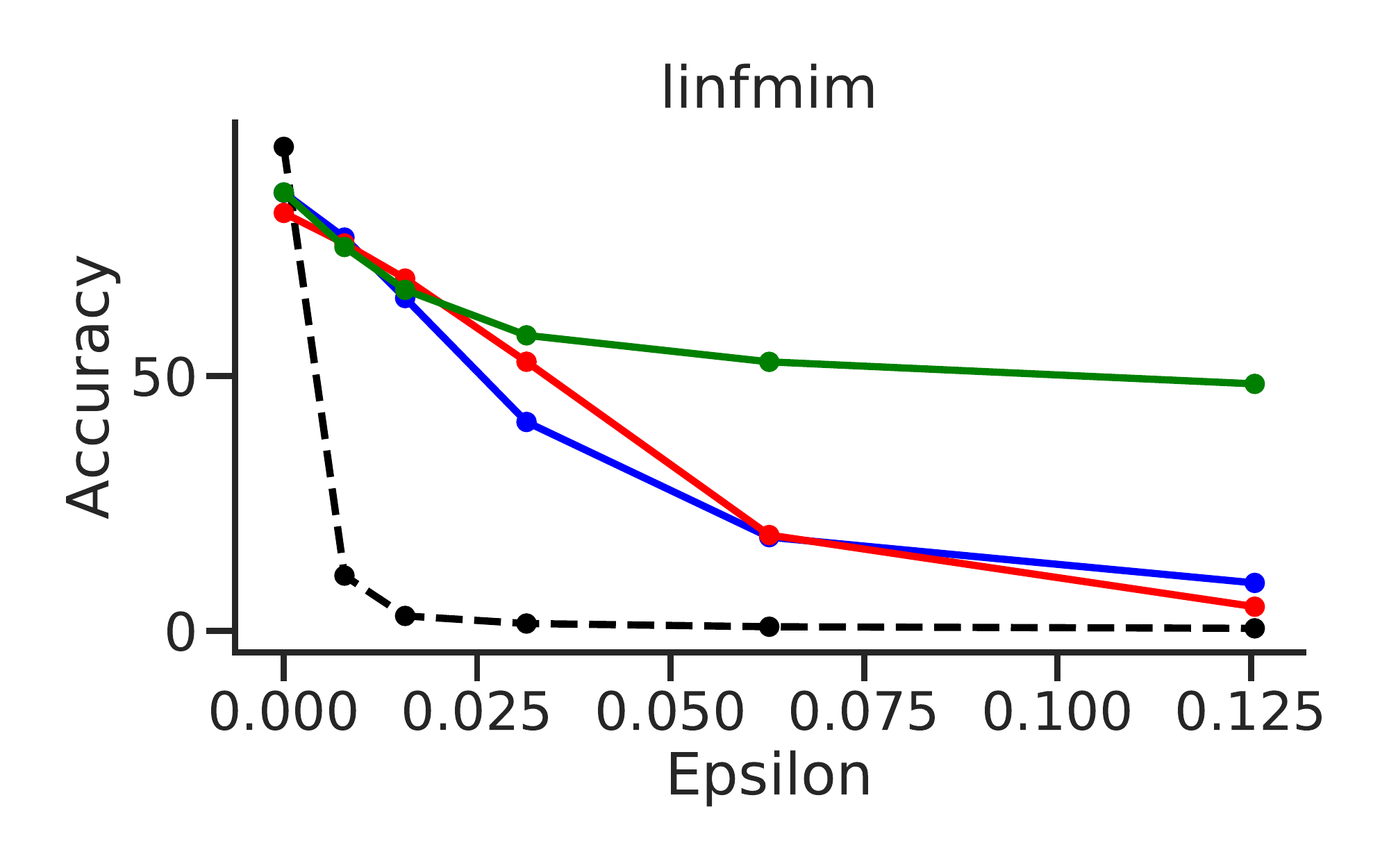}
\includegraphics[width=0.245\linewidth]{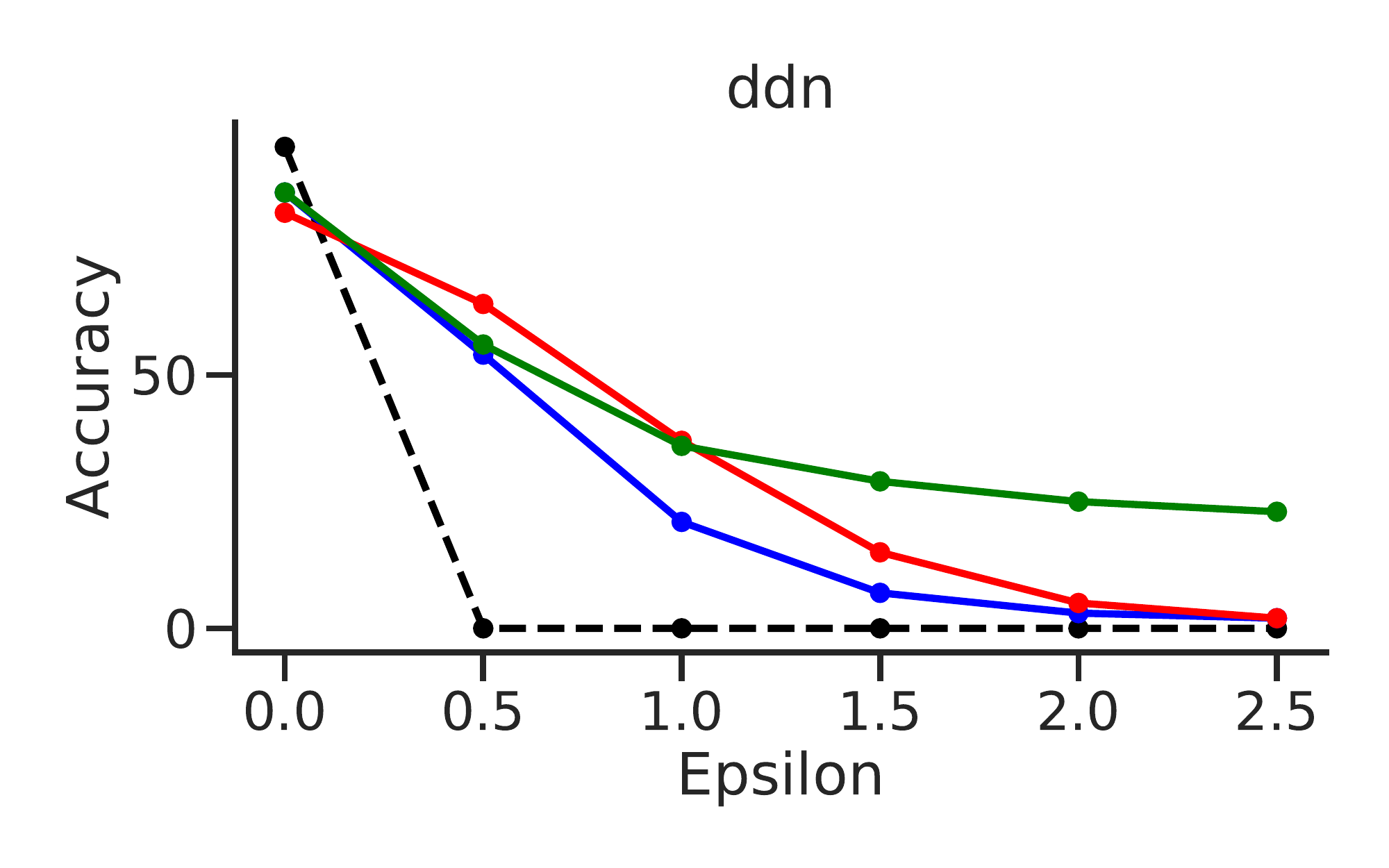}
\includegraphics[width=0.245\linewidth]{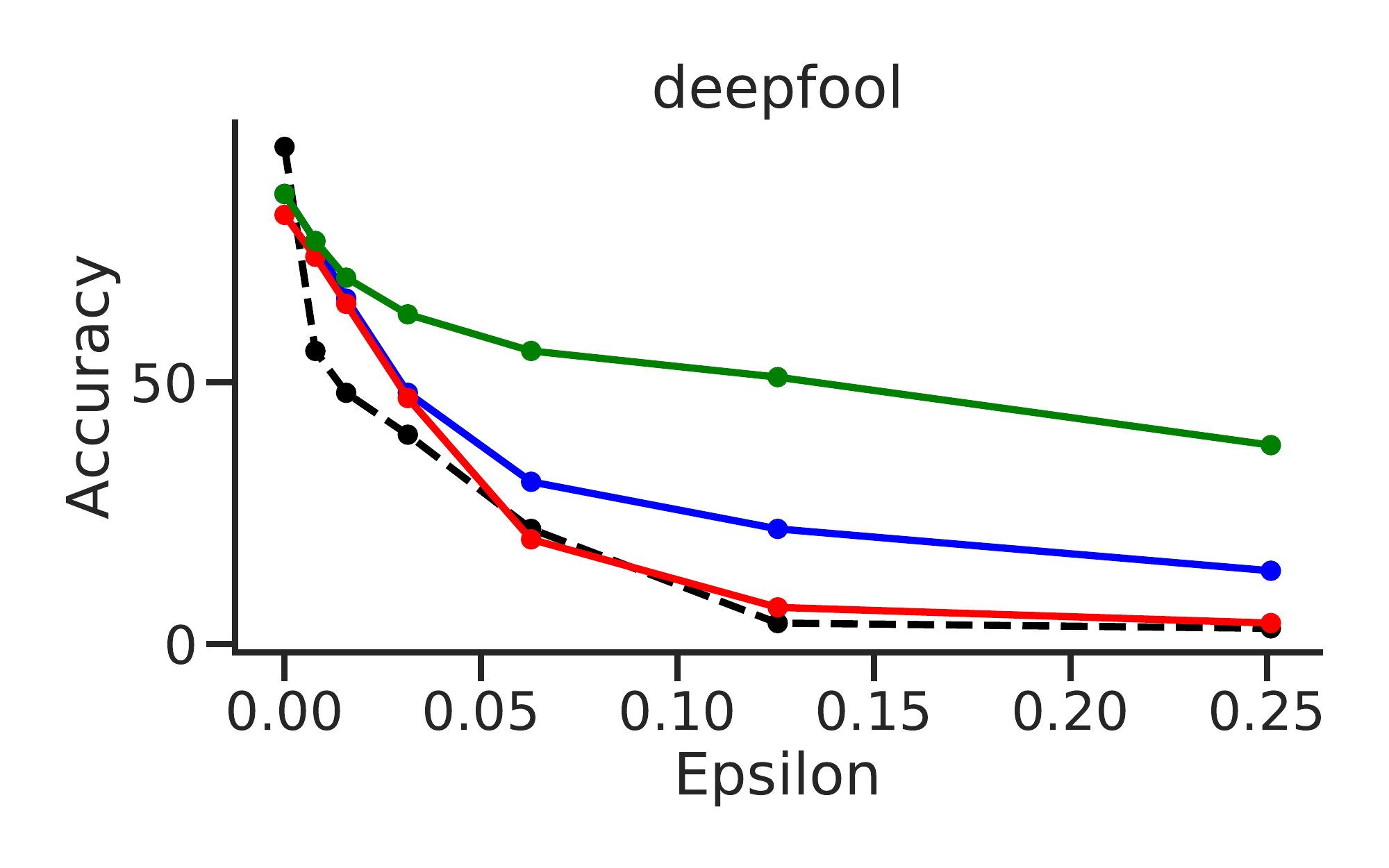}
\includegraphics[width=0.245\linewidth]{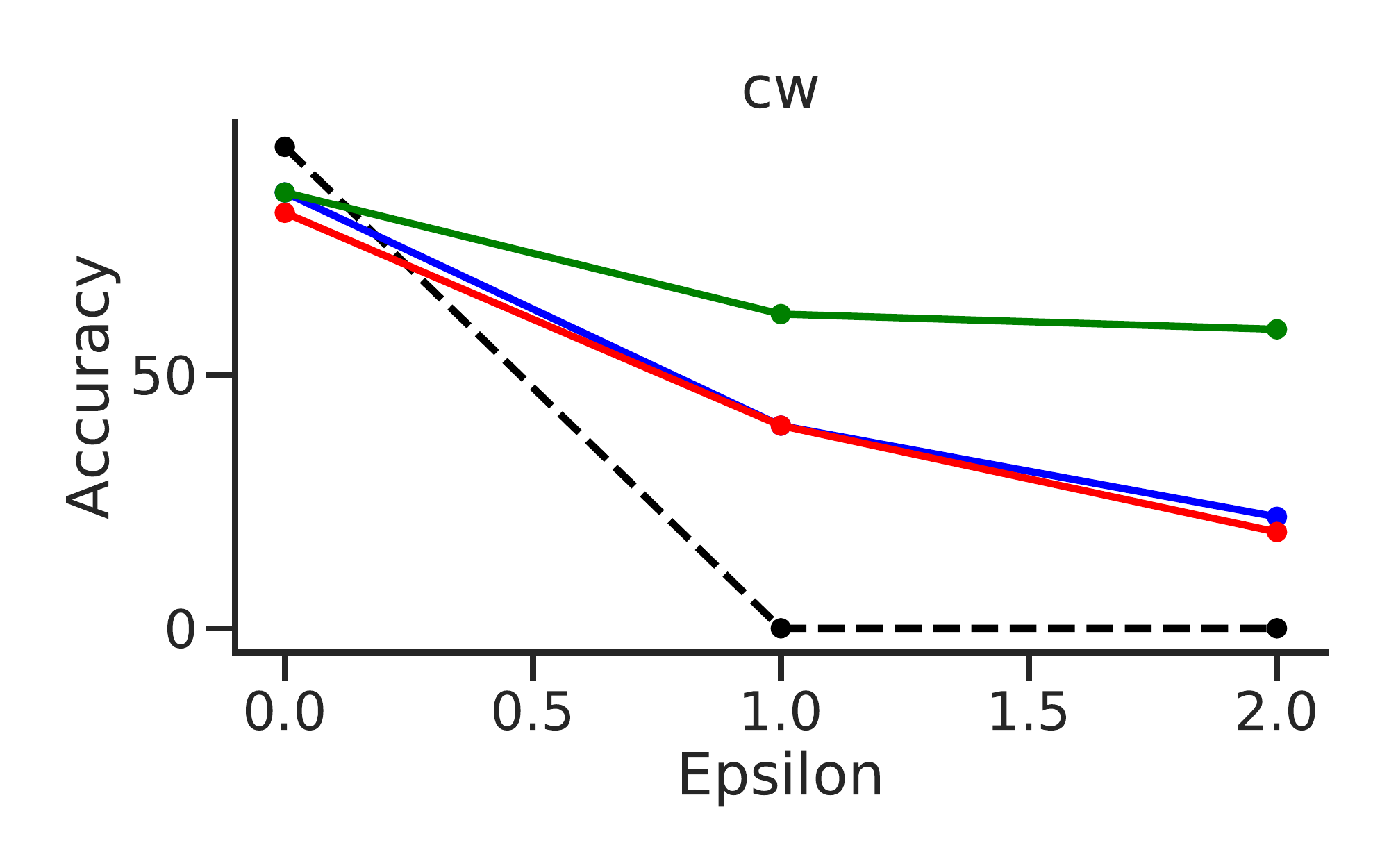}
\vspace{-0.18in}
\caption{Adversarial accuracy for different methods against white-box attacks on CIFAR10 dataset with ResNet18 architecture.}
\label{fig_cifar10_rn18_eps_lineplots}
\end{figure}

\begin{table}[t]
\centering
\caption{AUC measures for different perturbations and methods on MNIST, CIFAR10, CIFAR100, and tiny-imagenet datasets. AUC values are normalized to have a maximum allowable value of 1. Evaluations on AT and TRADES were made on networks trained using reimplemented or official code.\label{tab_auc}}
\resizebox{0.95\textwidth}{!}{%
\begin{tabular}{c|c|c|c|c|c|c|c|c|c|c}
\toprule
\textbf{Dataset}  & \textbf{Model}     
& \textbf{$\text{PGD}_{L_\infty}$}         & \textbf{$\text{PGD}_{L2}$}        & \textbf{$\text{PGD}_{L1}$} 
& \textbf{FGSM}                         & \textbf{MIM}                      & \textbf{DDN}              & \textbf{DeepFool}                       & \textbf{C\&W}   & \textbf{AA}     
\\ \midrule
\multirow{5}{*}{MNIST}
& NT     & 0.16              & 0.06              & 0.07              & 0.3              & 0.19                  & 0.09
    & 0.21              & 0.57         & 0.28    \\ 
& AT      & 0.74              & 0.29              & 0.19              & 0.83              & 0.95                  & 0.49
    & 0.55              & 0.87         & 0.89    \\ 
& TRADES   & 0.71              & 0.26              & 0.15              & 0.79              & 0.88                  & 0.42
    & 0.47              & 0.86          & 0.88   \\ 
& AFD-DCGAN      & 0.77              & 0.33              & 0.3              & 0.78              & 0.91                  & 0.51
    & 0.49              & 0.9            & 0.88  \\ 
& AFD-WGAN      & \textbf{0.92}              & \textbf{0.54}              & \textbf{0.55}              & \textbf{0.9}              & \textbf{0.98}                  & \textbf{0.68}
    & \textbf{0.63}              & \textbf{0.94}         & \textbf{0.90}    \\ 
\midrule
\multirow{5}{*}{CIFAR10}
& NT      & 0.05              & 0.1              & 0.17              & 0.19              & 0.05                  & 0.1
    & 0.16              & 0.1         & 0.12     \\ 
& AT      & 0.28              & 0.2              & 0.44              & 0.33              & 0.31                  & 0.26
    & 0.29              & 0.31        & 0.22     \\ 
& TRADES  & 0.32              & 0.22              & 0.5              & 0.24              & 0.32                  & 0.33
    & 0.18     & 0.28       & \textbf{0.25} \\ 
    & AFD-DCGAN      & 0.34     & \textbf{0.54}     & 0.43     & 0.4     & 0.31         & \textbf{0.4}
    & 0.43              & 0.47        & 0.22      \\
    & AFD-WGAN      & \textbf{0.56}     & \textbf{0.54}     & \textbf{0.66}     & \textbf{0.59}     & \textbf{0.56}         & \textbf{0.4}
    & \textbf{0.52}              & \textbf{0.62}            & 0.24  \\

\midrule
\multirow{5}{*}{CIFAR100}
& NT       & 0.03              & 0.08              & 0.1              & 0.07              & 0.03                  & 0.08
    & 0.06              & 0.08        & 0.09     \\ 
& AT       & 0.13              & 0.1              & 0.24              & 0.13              & 0.14                  & 0.14
    & 0.12              & 0.15        & 0.13     \\ 
& TRADES   & 0.16              & 0.13              & 0.31              & 0.12              & 0.17                 & \textbf{0.18}
    & 0.1     & 0.16    & \textbf{0.15} \\ 
& AFD-DCGAN       & 0.14     & 0.12     & 0.27     & 0.17     & 0.16         & 0.15
    & 0.16              & 0.18        & 0.13     \\ 
& AFD-WGAN       & \textbf{0.18}     & \textbf{0.16}     & \textbf{0.31}     & \textbf{0.22}     & \textbf{0.19}         & 0.16
    & \textbf{0.19}              & \textbf{0.23}         & 0.13    \\ 
\midrule
\multirow{4}{*}{Tiny-IN}
& NT       & 0.04              & 0.03              & 0.08              & 0.05              & 0.04                  & 0.06
    & 0.07              & 0.07         & 0.07     \\ 
& AT       & \textbf{0.10}              & 0.03              & 0.16              & \textbf{0.15}              & \textbf{0.09}                  & 0.14
    & 0.13              & 0.11         & 0.14     \\ 
& TRADES   & \textbf{0.10}              & 0.03              & 0.16              & 0.07              & \textbf{0.09}                 & \textbf{0.15}
    & 0.11     & 0.09       & \textbf{0.16} \\ 
& AFD-WGAN       & \textbf{0.10}     & \textbf{0.04}     & \textbf{0.19}     & 0.12     & \textbf{0.09}         & \textbf{0.15}
    & \textbf{0.16}              & \textbf{0.12}         & 0.15    \\ 

\bottomrule
\end{tabular}}
\end{table}

\vspace{-0.05in}
\subsection{Robust classification against stronger and unseen attacks} To evaluate how each network generalizes to unseen domains of adversarial inputs (i.e. adversarial attacks generated with unseen forms of adversarial attacks), we additionally validated the classification robustness against a range of possible $\epsilon$ values for several widely used attacks that were not used during training. To fairly compare different models while considering both attack types and $\epsilon$ values, we computed the area-under-the-curve (AUC) for accuracy vs. epsilon for each attack (similar to Figure-\ref{fig_cifar10_rn18_eps_lineplots}). Table-\ref{tab_auc} summarizes the AUC values for all 9 attack methods on four tested datasets. Compared with the baselines, we found that, AFD-trained networks consistently performed better on various datasets and on almost all the tested attacks even for substantially larger $\epsilon$ values (Figure \ref{fig_cifar10_rn18_eps_lineplots}, also see Figures \ref{fig_supp_mnist_eps_lineplots},\ref{fig_supp_cifar100_rn18_eps_lineplots} in the appendix). These results show that compared to other baselines, AFD-trained networks are robust against a wider range of attacks and attack strengths ($\epsilon$). This further suggests that the features learned through AFD generalize better across various forms of attacks and can sustain larger perturbations. 

We also observed that the AFD-WGAN performs better than AFD-DCGAN under most tested conditions. This is potentially due to: 1) WGAN’s ability to avoid vanishing gradients when the discriminator becomes too good compared to the generator (the feature extractor function in our case) [5]; 2) WGAN’s ability to avoid mode-collapses during training. In training GANs, mode collapses lead to the generator network to only output a limited set of patterns instead of learning to produce a diverse set of natural-looking images that fool the discriminator. Under our setting, WGAN potentially leads to learning a feature extractor that can produce a more diverse set of features for perturbed inputs, instead of focusing on a subset of latent dimensions. This suggests that applying more advanced GAN training algorithms could potentially further improve the robust performance in AFD-type models.

\vspace{-0.15in}
\subsection{Estimated $\mathcal{H}\Delta\mathcal{H}$-distance and adversarial-vs-natural generalization gap}
\begin{figure}[t]
\centering
\includegraphics[width=.45\linewidth]{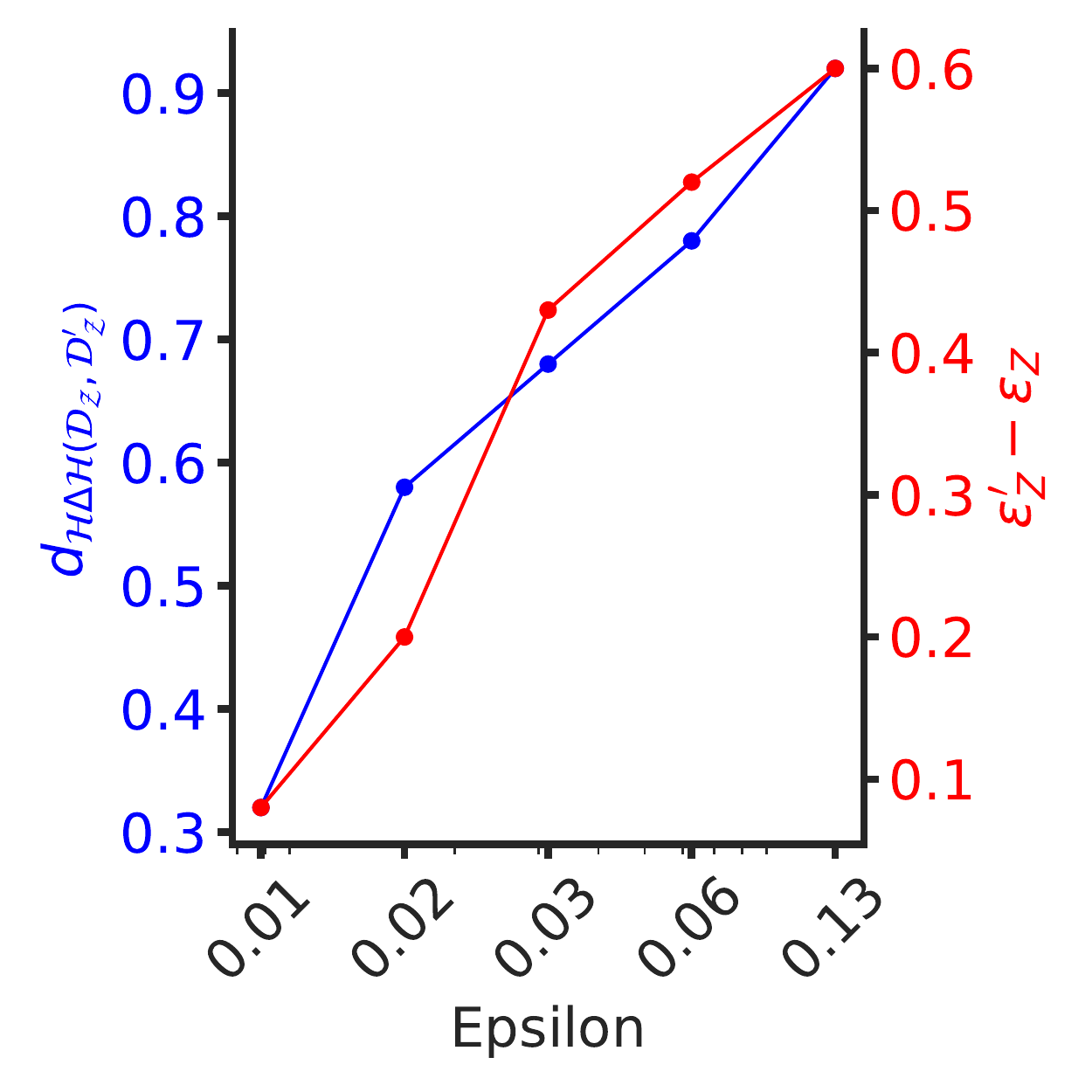}
\includegraphics[width=.45\linewidth]{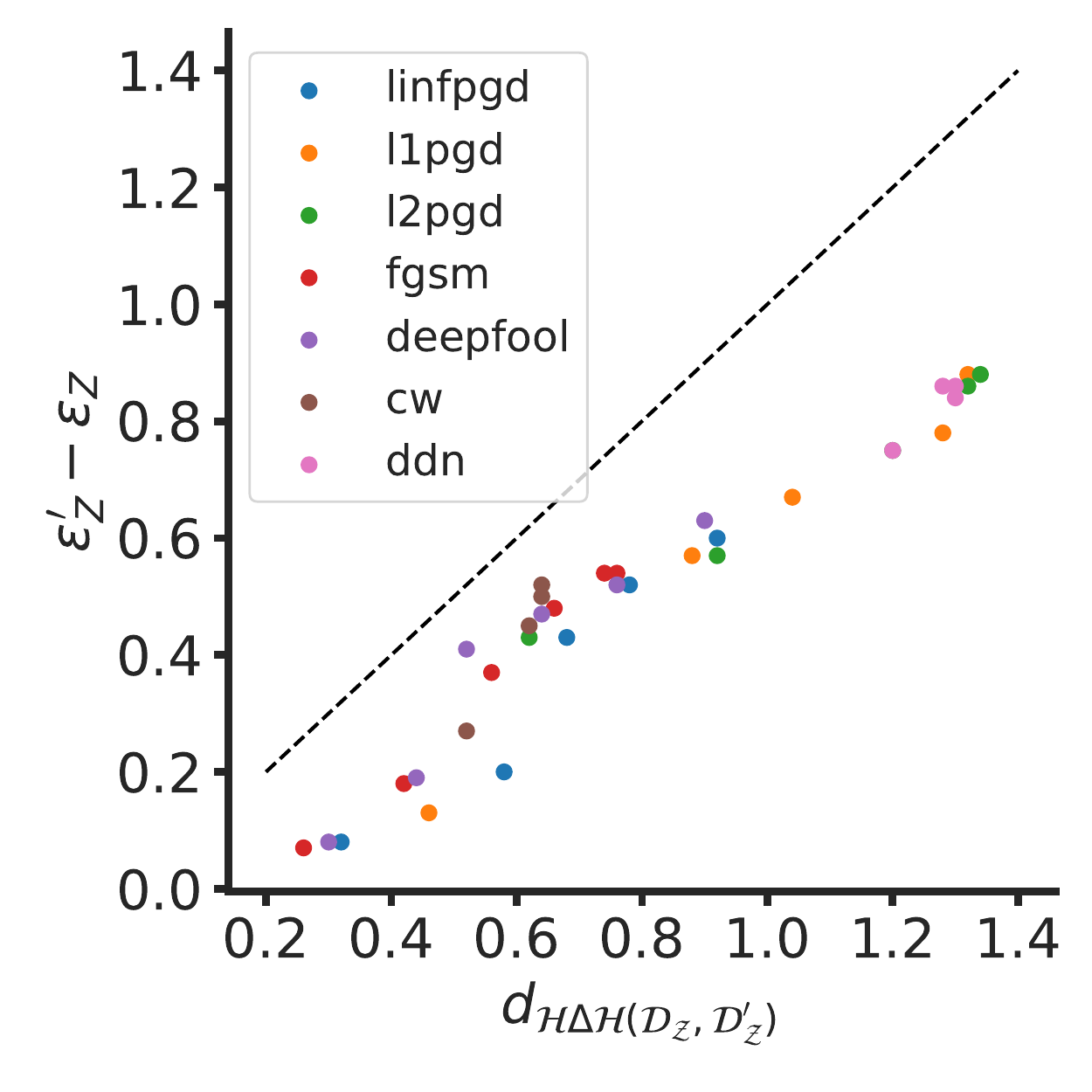}
\vspace{-0.05in}
\centerline{(a) \hspace{2in} (b) }
\vspace{-0.2in}

\caption{(a) Estimated $d_{\mathcal{H}\Delta\mathcal{H}}$ distance (based on empirical value of $\alpha$ in Eq. \ref{dist_est})  and generalization gap in adversarial and natural error $\epsilon_{Z}'-\epsilon_{Z}$ as a function of epsilon for $PGD-L_{\infty}$ attack; (b) scatter plot of the estimated $d_{\mathcal{H}\Delta\mathcal{H}}$ distance using the domain discriminator and the gap in adversarial and natural error across different attack types and magnitudes (i.e. $\epsilon$). Colors correspond to different attack types. Each dot corresponds to one attack evaluated at a particular epsilon value. Estimations of the $d_{\mathcal{H}\Delta\mathcal{H}}$ distance for all attacks and epsilons are made with the domain discriminator trained on PGD-$L_{\infty}$ with $\epsilon=0.031$.}
\vspace{-0.15in}
\label{fig_domain_clf_scatter}
\end{figure}

\vspace{-0.05in}
As stated in Eq. \ref{eq_hdh_bound}, the upper bound on the adversarial error can be stated in terms of the natural error, the divergence between the two domains, and a constant term. In practice, this means that the smaller the divergence term $d_{\mathcal{H}\Delta\mathcal{H}}$
is, the smaller the gap between the adversarial  and natural errors ($\epsilon_{Z}'-\epsilon_{Z}$) can be. We empirically tested this prediction using the domain discriminator trained on CIFAR10 dataset using the PGD-$L_\infty$ attack. Figure-\ref{fig_domain_clf_scatter}a shows that the estimated $d_{\mathcal{H}\Delta\mathcal{H}}$ using the domain discriminator (i.e., using the corresponding  empirical value of $\alpha$ in Eq. \ref{dist_est})  trained on $PGD-L_{\infty}$ with $\epsilon=0.031$ is closely related to the adversarial-vs-natural generalization gap over different $\epsilon$ values as predicted by Eq. \ref{eq_hdh_bound}. Moreover, estimations from the same domain discriminator also predicts the gap in generalization error attained for other forms of attacks (even ones not seen during AFD training) and $\epsilon$ values with high accuracy (Figure-\ref{fig_domain_clf_scatter}b). This further supports the proposal that minimizing the estimated distance between the natural and adversarial representations can be an efficient way to improve the model robustness against various adversarial attacks. 

\vspace{-0.05in}
\subsection{Learning a sparse representation} 
\vspace{-0.05in}
Because the AFD method aims to learn a representation that is insensitive to adversarial attacks, we expected the learned representational space to potentially be of lower dimensionality (i.e. less number of orthogonal features). To test this, we compared the dimensionality of the learned representation using two measures. i) number of non-zero features over the test set within each dataset and ii) number of Principal Component Analysis (PCA) dimensions that explains more than 99\% of the variance in the representation computed over the test-set of each dataset. We found that the same network architecture (i.e. ResNet18), when trained with AFD method learns a much sparser and lower dimensional representational space (Table \ref{table_supp_dimensions}) compared to the naturally trained, adversarial training and TRADES models. The representational spaces learned with AFD on MNIST, CIFAR10, and CIFAR100 datasets had only 6, 9, and 76 principal components respectively.

\vspace{-0.05in}
\subsection{Adversarial and norm-based desensitization}
\vspace{-0.05in}
To investigate whether the same level of robustness could be achieved by encouraging the network to produce similar representations in response to natural and adversarial inputs, we ran an additional experiment on the MNIST dataset in which we added a regularization term to the classification loss to directly minimize the representation sensitivity $S_e=\frac{1}{n} \sum_{x} \norm{F(x)-F(x^\prime)}, $ during training. We observed that although this augmented loss led to learning robustness representations, it achieved modest levels of robustness ($\sim80\%$) and showed only weak generalization to stronger and other unseen attacks (Figure-\ref{fig_supp_mnist_eps_lineplots_embmatch}). This result suggests that more direct forms of enforcing representational similarity may not lead to the same form of robustness with generalization properties similar to that achieved using an adversarial training with domain discriminator (e.g. as in AFD).

\vspace{-0.05in}
% \textbf{Non-obfuscated gradients} Recent literature have pointed out that many defense methods against adversarial perturbations could drive the network into a regime called \textit{obfuscated gradients} in which the network appears to be robust against common iterative adversarial attacks but could easily be broken using black-box or alternative attacks that do not rely on exact gradients \cite{Papernot2017,Athalye2018,carlini2019evaluating}. We believe that our results are not due to obfuscated gradients for several reasons. i) For most perturbations, the model performance continues to decrease with increased epsilon (Figures-\ref{fig_cifar10_rn18_eps_lineplots},\ref{fig_supp_mnist_eps_lineplots},\ref{fig_supp_cifar100_rn18_eps_lineplots}); ii) The iterative perturbations were consistently more successful than single-step ones (Table-\ref{table_accu}); iii) Black-box attacks were significantly less successful than white-box attacks (Table-\ref{table_accu}); 
% iv) The AFD-trained model performed similar or better than alternate methods against the Boundary attack \cite{Brendel2018} -- an attack which does not rely on the network gradients (Table-\ref{tab_supp_boundary}). In addition to these tests, we also evaluated the AFD performance on B\&B \cite{Brendel2018} and AutoAttack \cite{croce2020reliable}. On these attacks, AFD was consistently better than or equal to the baseline models on MNIST and CIFAR10 datasets but was less robust on the CIFAR100 dataset (Table-\ref{tab_supp_boundary}).

\vspace{-0.05in}
\section{Conclusion and limitations}
\vspace{-0.05in}
 Decreasing the input-sensitivity of features has long been desired in training neural networks \cite{Drucker1992} and has been suggested as a way to improve adversarial robustness \cite{Ros2018,zhu2020learning}. In this work we proposed an algorithm to decrease the sensitivity of neural network representations using an adversarial learning paradigm that involves joint training of a domain discriminator, a feature extractor, and a task classifier. Essentially, our proposed algorithm iteratively estimates a bound on the adversarial error in terms of the natural error and a classification-based measure of distance between the distributions of natural and adversarial features and then minimizes the adversarial error by concurrently reducing the natural error as well as the distance between the two feature distributions.

\textbf{Limitations}. The empirical results presented here suggest that AFD-trained models are robust against a wide range of adversarial attacks (distributions) even compared to strong baselines like Adversarial Training and TRADES. However, it is not guaranteed that the model would remain robust against any unseen attacks that we have not tested or may be invented in the future - as is the case in domain adaptation literature and the lack of theoretical guarantees for cross-domain generalization. With regards to the computational cost, when measuring the average per-epoch training time on the CIFAR10 dataset (using 2 NVIDIA V100 GPUs), we found that the AFD training time is 31\% longer than adversarial training and only 4\% longer than TRADES. This shows that while AFD requires three SGD updates per batch, the additional computational cost is not significantly higher than many prior methods when considering that most of the computational cost is associated with generating the adversarial examples during training. 
% We believe that these results could further be improved by i) using complementary attacks during training; ii) using larger neural networks such as wider or deeper networks as is shown in recent work \cite{Xie2020,madry2017towards}; iii) by applying the adversarial learning paradigm on multiple feature layers of the network;

\section{Acknowledgements}
We would like to thank Isabela Albuquerque, Joao Monteiro, and Alexia Jolicoeur-Martineau for their valuable comments on the manuscript. Pouya Bashivan was partially supported by the Unifying AI and Neuroscience – Québec (UNIQUE) Postdoctoral fellowship and NSERC Discovery grant RGPIN-2021-03035. Irina Rish acknowledges the support from Canada CIFAR AI Chair Program and from the Canada Excellence Research Chairs Program.

\bibliography{neurips2021_refs}
\bibliographystyle{plain}

\clearpage

\setcounter{figure}{0}
\setcounter{table}{0}
\setcounter{theorem}{0}
\renewcommand\thefigure{A\arabic{figure}}  
\renewcommand\thetable{A\arabic{table}} 
\renewcommand{\theHtable}{Supplement.\thetable}
\renewcommand{\theHfigure}{Supplement.\thefigure}

\appendix
\section{Appendix}

\subsection{Network architectures}
For all experiments, we trained the ResNet18 architecture \cite{he2016deep} using SGD optimizer with 0.9 momentum and learning rates as indicated in Table-\ref{table_supp_hyperpars}, weight decay of $10^{-4}$, batch size of 128. All learning rates were reduced by a factor of 10 after scheduled epochs.

\begin{table}[h]
\centering
\caption{Training hyperparameters for each dataset and network.\label{table_supp_hyperpars}}
\resizebox{\textwidth}{!}{%
\begin{tabular}{c|c|c|c|c|c|c|c|c|c}
\toprule
\textbf{Dataset}      & \textbf{Model}       
& \textbf{$\text{LR}_E$}  & \textbf{$\text{LR}_{Da}$}        & \textbf{$\text{LR}_{EDc}$}
& \textbf{Weight Decay}
& \textbf{Batch Size}
& \textbf{\# Epochs}
& \textbf{Scheduled Epochs}
& \textbf{WGAN GP}
\\ \midrule
% MNIST      & LeNet     & 0.01       & 0.01       & 0.01       & $10^{-4}$        & 50       & 200   \\
MNIST      & \multirow{4}{*}{ResNet18}     & \multirow{4}{*}{0.5}       & \multirow{4}{*}{0.1}       & \multirow{4}{*}{0.1}       & \multirow{4}{*}{$10^{-4}$}        & \multirow{4}{*}{128}       & 100   & [50, 80] & 0\\
CIFAR-10      &      &        &        &        &           &      & 300    & [150, 250] & 1.0\\
CIFAR-100  &      &        &       &        &         &      & 300   & [150, 250] & 0\\
Tiny-Imagenet  &      &        &       &        &         &      & 500   & [300, 450] & 0\\ \bottomrule
\end{tabular}}
\end{table}

\subsection{Adversarial attacks}
We used a range of adversarial attacks in our experiments. Hyperparameters associated with each attack are listed in the table below. Implementation of these attacks were adopted from Foolbox \cite{rauber2017foolbox}, AdverTorch \cite{ding2019advertorch} packages.

\subsection{Wasserstein GAN Loss}
For AFD-WGAN model, we used the generator and discriminator losses from \cite{Arjovsky2017} to adversarially train the feature extractor $F_\theta$ and domain discriminator $D_\psi$ respectively. 
\begin{equation}
\label{eq_wgan_ld}
    \mathcal{L}_{D}=\frac{1}{m}\sum_{i=1}^{m}\Big[-D_{\psi}(F_\theta(x_i), \,y_i)+D_{\psi}(F_\theta(x^{\prime}_i), \,y_i)\Big]
\end{equation}
\begin{equation}
\label{eq_wgan_lf}
    \mathcal{L}_{F}=\frac{1}{m}\sum_{i=1}^{m}-D_{\psi}(F_\theta(x_i),\,y_i)
\end{equation}

\section{Broader Impact}
As the application of deep neural networks becomes more common in everyday life, security and dependability of these networks becomes more crucial. While these networks excel at performing many complicated tasks under standard settings, they often are criticized for their lack of reliability under broader settings. One of the main points of criticism of today's artificial neural networks is on their vulnerability to adversarial patterns -- slight but carefully constructed perturbations of the inputs which drastically decrease the network performance. 

Our work presented here proposes a new way of addressing this important issue. Our approach could be used to improve the robustness of learned representation in an artificial neural network and as shown lead to a recognition behavior that is more aligned with the human judgement. More broadly, the ability to learn robust representations and behaviors is highly desired in a wide range of applications and disciplines including perception, control, and reasoning and we expect the presented work to influence the future studies in these areas.

\begin{table}[h]
\centering
\caption{Attack hyperparameters for each dataset and attack.}
\label{table_supp_attack_settings}
% \resizebox{\textwidth}{!}{%
\begin{tabular}{c|c|c|c|c|c}
\toprule
\textbf{Attack}      & \textbf{Dataset}       & \textbf{Steps}    & \textbf{$\epsilon$}        & \textbf{More}     & \textbf{Toolbox}
\\ \midrule
\multirow{3}{*}{FGSM}      
& MNIST     & \multirow{3}{*}{1}       & [0, 0.1, 0.3, 0.35, 0.4, 0.45, 0.5]       & -       & \multirow{3}{*}{Foolbox}           \\
& CIFAR     &       & [0, $\frac{2}{255}$, $\frac{4}{255}$, $\frac{8}{255}$, $\frac{16}{255}$, $\frac{32}{255}$, $\frac{64}{255}$]       & -           \\
& Tiny-IN     &       & [0, $\frac{2}{255}$, $\frac{4}{255}$, $\frac{8}{255}$, $\frac{16}{255}$, $\frac{32}{255}$]       & -           \\
\midrule
\multirow{3}{*}{PGD-$L_1$}      & MNIST     & \multirow{3}{*}{50}       & \multirow{2}{*}{[[0, 10, 50, 100, 200]]}       & \multirow{3}{*}{step=0.025}       & \multirow{3}{*}{Foolbox}             \\
& CIFAR     &       &        &        &              \\
& Tiny-IN     &       &  [0, 10, 50]      &        &              \\
\midrule

\multirow{3}{*}{PGD-$L_2$}  & MNIST     & \multirow{3}{*}{50}       & \multirow{2}{*}{[0, 2, 5, 10]}       & \multirow{3}{*}{step=0.025}       & \multirow{3}{*}{Foolbox}        \\ 
& CIFAR     &       &        &           &  \\   
& Tiny-IN     &       & [0, 2, 5]      &           &  \\   
\midrule

\multirow{3}{*}{PGD-$L_\infty$}  & MNIST     & 40       & [0, 0.1, 0.3, 0.35, 0.4, 0.45, 0.5]       & step=0.033       & \multirow{3}{*}{Foolbox}        \\ 
& CIFAR     & 20      & [0, $\frac{2}{255}$, $\frac{4}{255}$, $\frac{8}{255}$, $\frac{16}{255}$, $\frac{32}{255}$]       & \multirow{2}{*}{step=$\frac{2}{255}$}   & \\
& Tiny-IN     & 20      & [0, $\frac{2}{255}$, $\frac{4}{255}$, $\frac{8}{255}$, $\frac{16}{255}$]       &   & \\
\midrule

\multirow{3}{*}{MIM}  & MNIST     & \multirow{3}{*}{40}       & [0, 0.1, 0.3, 0.5, 0.8, 1]       & -       & \multirow{3}{*}{AdverTorch}        \\ 
& CIFAR     &       & [0, $\frac{2}{255}$, $\frac{4}{255}$, $\frac{8}{255}$, $\frac{16}{255}$, $\frac{32}{255}$]       & -    &  \\ & Tiny-IN     &       & [0, $\frac{2}{255}$, $\frac{4}{255}$, $\frac{8}{255}$, $\frac{16}{255}$]       & -    &  \\
\midrule

\multirow{3}{*}{DDN}  & MNIST     & \multirow{3}{*}{100}       & [0, 1, 2, 5]       & -       & \multirow{3}{*}{Foolbox}        \\ 
& CIFAR     &       & [0, 2, 5, 10, 15]       & -       & \\
& Tiny-IN     &       & [0, 0.2, 0.5, 1]       & -       & \\
\midrule

\multirow{3}{*}{Deepfool}  & MNIST     & \multirow{3}{*}{50}       & [0, 0.1, 0.3, 0.35, 0.4, 0.45, 0.5]       & -       & \multirow{3}{*}{Foolbox}        \\
& CIFAR     &       & [0, $\frac{2}{255}$, $\frac{4}{255}$, $\frac{8}{255}$, $\frac{16}{255}$, $\frac{32}{255}$, $\frac{64}{255}$]       & -       & \\
& Tiny-IN     &       & [0, $\frac{2}{255}$, $\frac{4}{255}$, $\frac{8}{255}$, $\frac{16}{255}$]       & -       & \\
\midrule

\multirow{3}{*}{C\&W}  & MNIST     & \multirow{3}{*}{100}       & \multirow{3}{*}{[0, 0.5, 1, 1.5, 2]}       & \multirow{3}{*}{stepsize=0.05}       & \multirow{3}{*}{Foolbox}        \\
& CIFAR     &       &       &        & \\
& Tiny-IN     &       &       &        & \\

\midrule
\multirow{3}{*}{AA}  & MNIST     & \multirow{3}{*}{100}       & [0, 0.2, 0.3, 0.35]       & \multirow{3}{*}{-}       & \multirow{3}{*}{AutoAttack}        \\
& CIFAR     &       & [0, 8/255., 16/255., 32/255.]      &        & \\
& Tiny-IN     &       & [0, 2/255., 4/255., 8/255.]      &        & \\
\bottomrule
\end{tabular}
\end{table}

% \begin{table}[h]
% \centering
% \caption{Comparison of adversarial accuracy against Boundary attack \cite{Brendel2018} with 5000 steps and $\epsilon=2$, and B\&B attack \cite{brendel2019accurate}. We tested the robust performance of each model on 100 random samples from each dataset's test-set.\label{tab_supp_boundary}}
% % \resizebox{\textwidth}{!}{%
% \begin{tabular}{c|c|c|c|c|c}
% \toprule
% \textbf{Dataset}  & \textbf{Model}  & \textbf{Method}   & \textbf{Boundary Attack}  & \textbf{B\&B}\\
% \midrule
% \multirow{4}{*}{MNIST}   & \multirow{4}{*}{RN18}       
% & NT             & 25    & 3\\
% &       & AT         & 63    & 92\\
% &       & TRADES          & 48    & 17\\
% &       & AFD        & 78   & 96\\
% \midrule
% \multirow{4}{*}{CIFAR10}   & \multirow{4}{*}{RN18}       
% & NT               & 0         & 0\\
% &       & AT            & 51        & 36\\
% &       & TRADES           & 58    & 54\\
% &       & AFD           & 68       & 41\\
% \midrule
% \multirow{4}{*}{CIFAR100}    & \multirow{4}{*}{RN18}       
% & NT                     & 2         & 0\\
% &       & AT             & 32    & 27\\
% &       & TRADES          & 35    & 30\\
% &       & AFD             & 32   & 10\\
% \bottomrule
% \end{tabular}
% % }
% \end{table}

\begin{table}[h]
\centering
\caption{Transfer black-box attack from ResNet18 network trained with adversarially training (AT) and TRADES on different datasets. \label{tab_supp_trasnfer_bb}}
% \resizebox{\textwidth}{!}{%
\begin{tabular}{c|c|c|c}
\toprule
\textbf{Dataset}  & \textbf{Method}  
& \textbf{AT Transfer}   & \textbf{TRADES Transfer}\\
\midrule
\multirow{3}{*}{MNIST}
& AT    & -     & 97.32\\
& TRADES & 96.64    & -\\
& AFD     & \textbf{97.41}   & \textbf{97.58}\\
\midrule
\multirow{3}{*}{CIFAR10}   
& AT        & -      & 64.34\\
& TRADES        & 78.43      & -\\
& AFD        & \textbf{86.36}      & \textbf{66.49}\\
\midrule
\multirow{3}{*}{CIFAR100}       
& AT        & -      & \textbf{42.54}    \\
& TRADES     & 39.22      & -    \\
& AFD        & \textbf{43.59}     & 42.26   \\
\bottomrule
\end{tabular}
\end{table}

\begin{table}[th]
\centering
\caption{Dimensionality of the learned representation space on various  datasets using different methods and measures. Units: number of non-zero feature dimensions over the test-set within each dataset. Dims: number of PCA dimensions that account for 99\% of the variance across all images within the test-set of each dataset.}
\label{table_supp_dimensions}
% \resizebox{\textwidth}{!}{%
\begin{tabular}{c|cc|cc|cc}
\toprule
\textbf{Dataset}    & \multicolumn{2}{c|}{\textbf{MNIST}}     &\multicolumn{2}{c}{\textbf{CIFAR10}} &\multicolumn{2}{c}{\textbf{CIFAR100}} \\
\midrule 
\multirow{2}{*}{\textbf{Network}}
& \multicolumn{2}{c|}{\textbf{ResNet18}}
& \multicolumn{2}{c|}{\textbf{ResNet18}}
& \multicolumn{2}{c}{\textbf{ResNet18}} \\
& Units    & Dims   & Units    & Dims    &Units     & Dims \\
\midrule 
NT    & 64     & \textbf{9}     & 512     & 70     & 512      & 376 \\
% & 184     & 9    \\
AT    & 64     & \textbf{9}     & 512     & 75     & 512     & 440 \\
% & 381     & 32   \\  
TRADES    & 64     & 14     & 512     & 70    &  512     & 339  \\
% & 296     & 27    \\
AFD    & \textbf{28}     & \textbf{9}     & \textbf{389}     & \textbf{12}       & \textbf{511}      & \textbf{304} \\
% & 28     & 7   \\
\bottomrule
\end{tabular}
\end{table}

\begin{table}[hbt!]
\centering
\caption{Comparison of adversarial accuracy on MNIST dataset against PGD-$L_{\infty}$ with $\epsilon=0.3$ for different domain discriminator architectures. FC1 and FC3 architectures refer to 1-layer and 3-layer fully connected networks respectively. PD refers to projection discriminator. \label{tab_ablation}}
% \resizebox{\textwidth}{!}{%
\begin{tabular}{c|c|c|c}
\toprule
\textbf{Dataset}  & \textbf{Model}  & \textbf{$Da$ Architecture}  & \textbf{Adversarial Acc.} \\
\midrule
\multirow{3}{*}{MNIST}   & \multirow{3}{*}{RN18}       & FC1-PD        & 85.96 \\
&   & FC3        & 90.73 \\
&   & FC3-PD     & \textbf{97.03} \\
\bottomrule
\end{tabular}
\end{table}

\begin{figure}[hbt!]
\centering
\includegraphics[width=.24\linewidth]{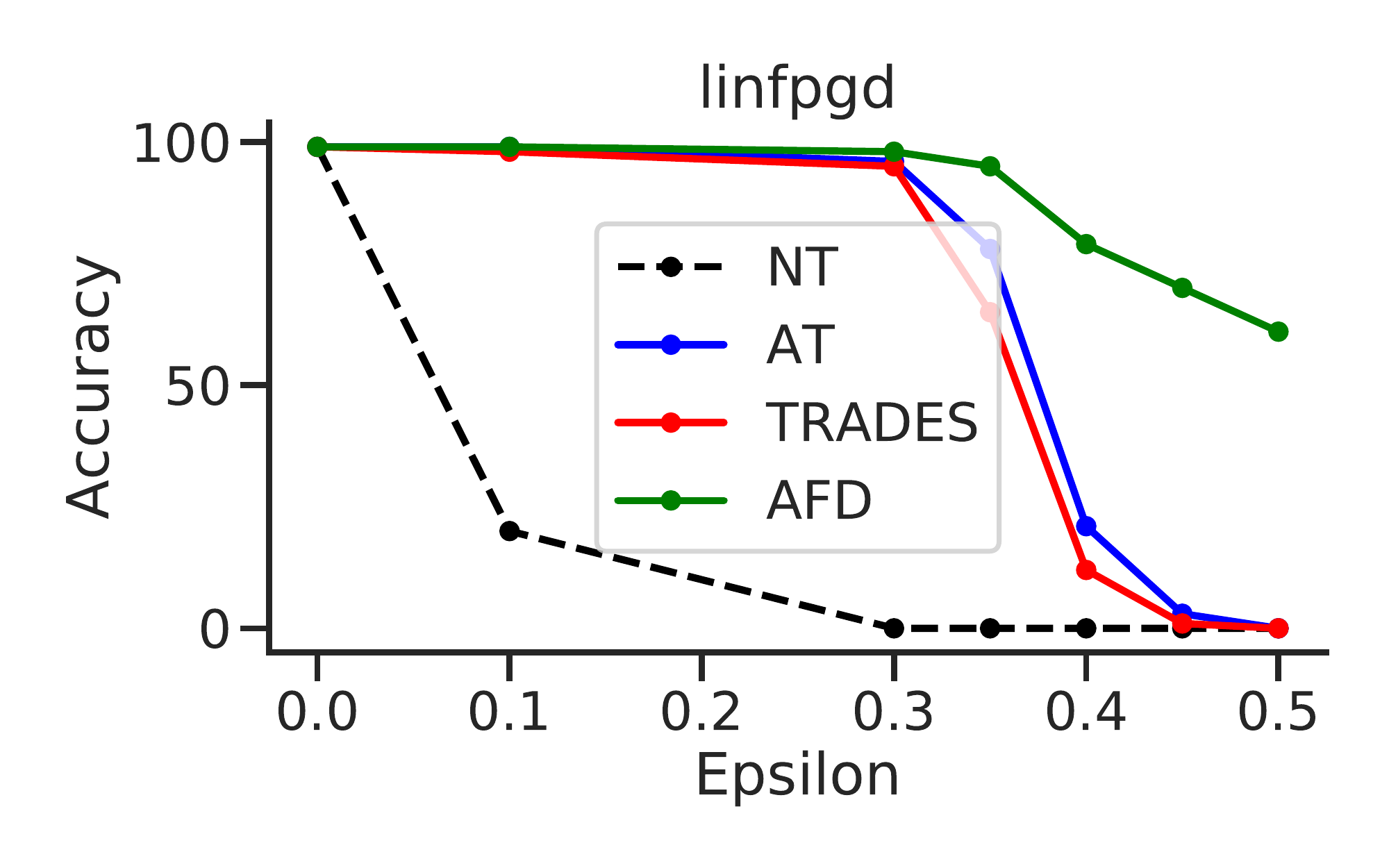}
\includegraphics[width=.24\linewidth]{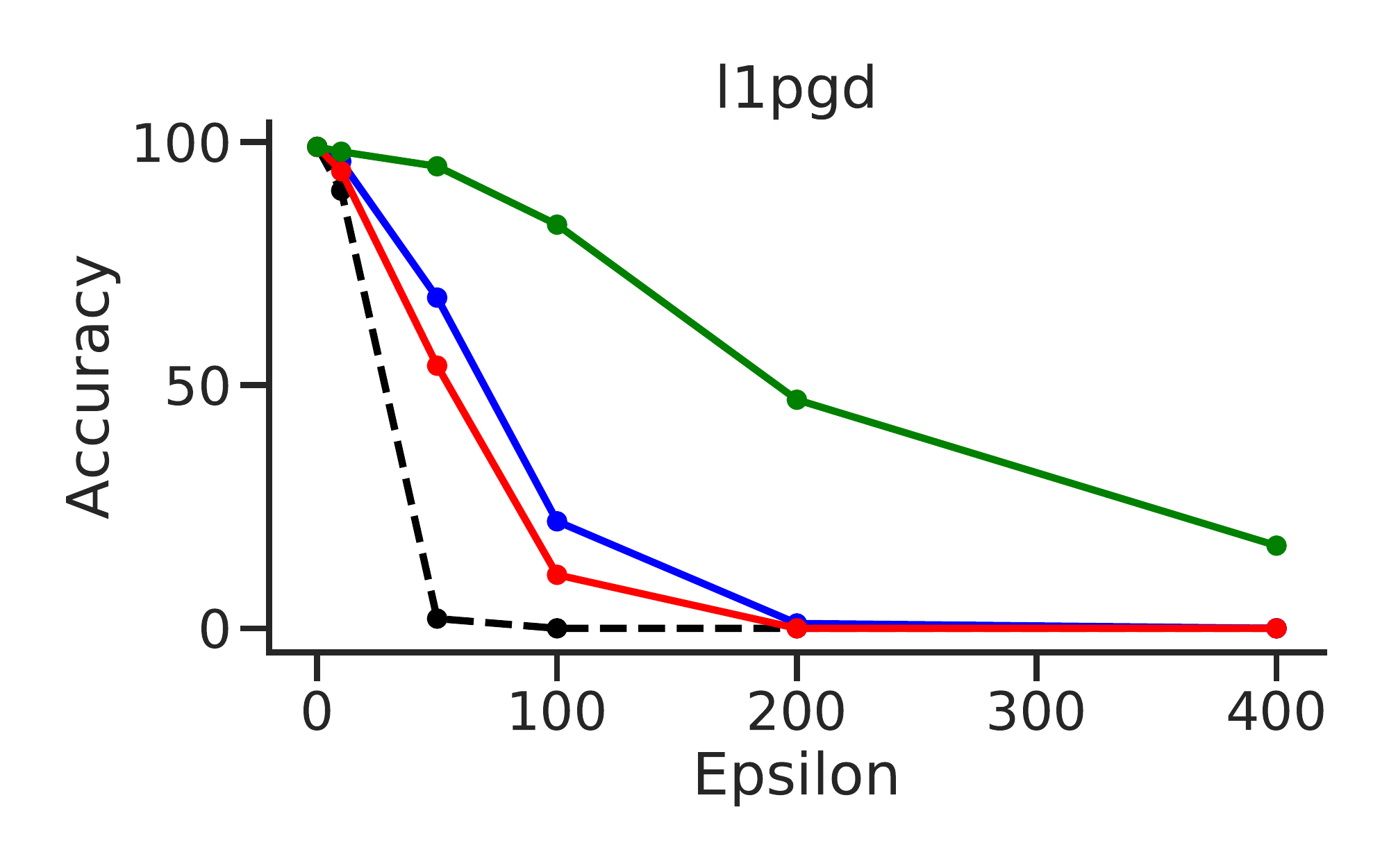}
\includegraphics[width=.24\linewidth]{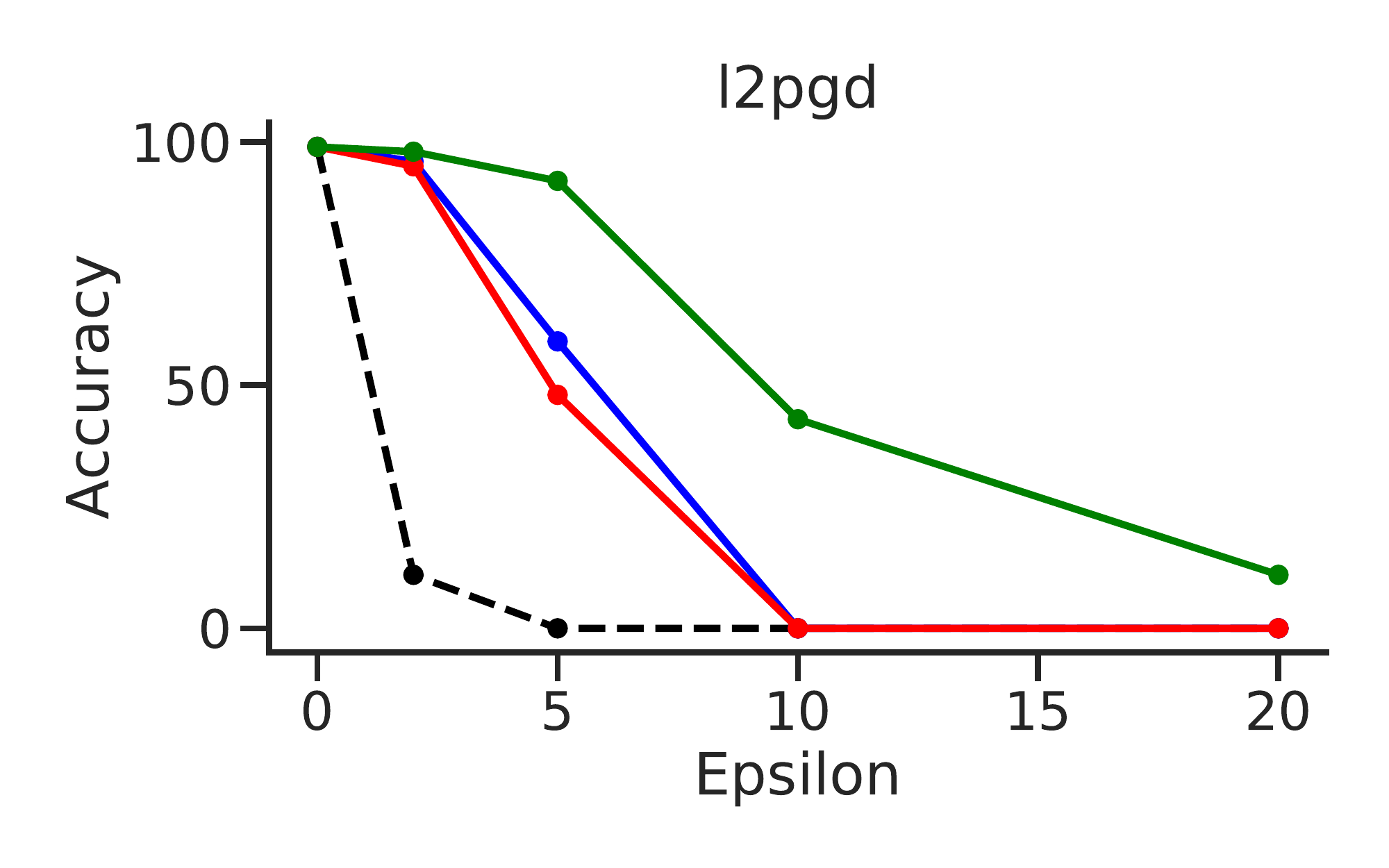}
\includegraphics[width=.24\linewidth]{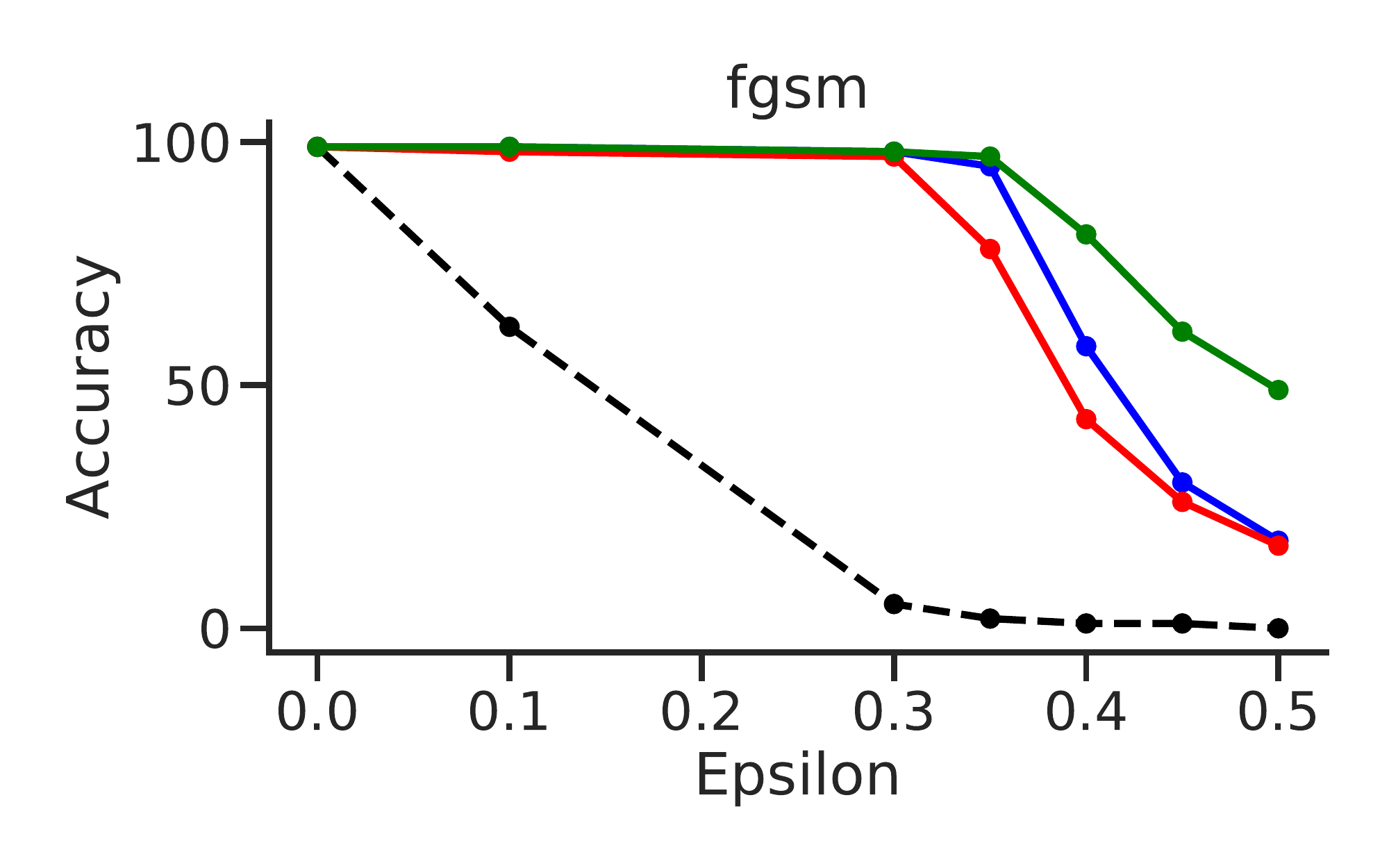}
\includegraphics[width=.24\linewidth]{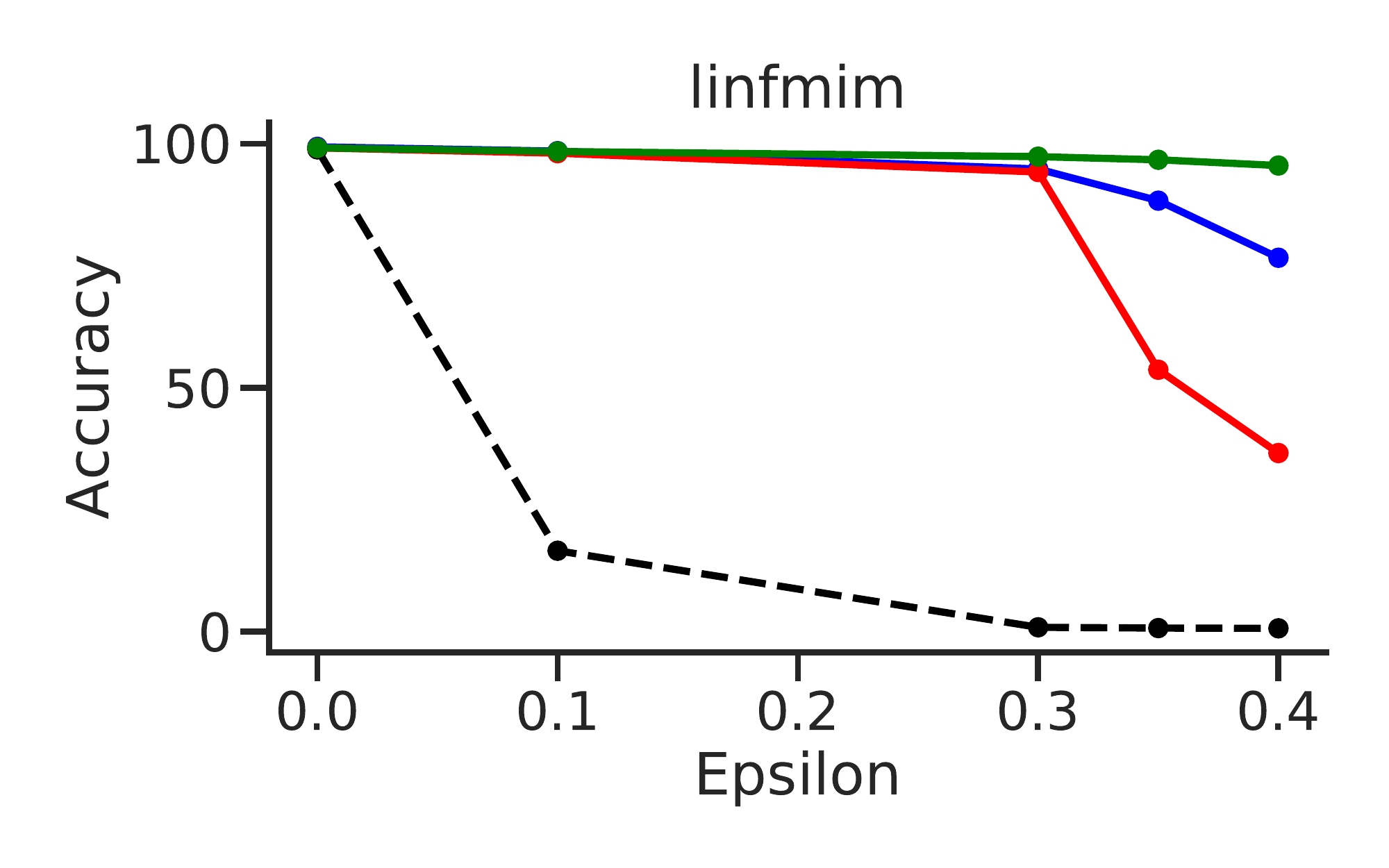}
\includegraphics[width=.24\linewidth]{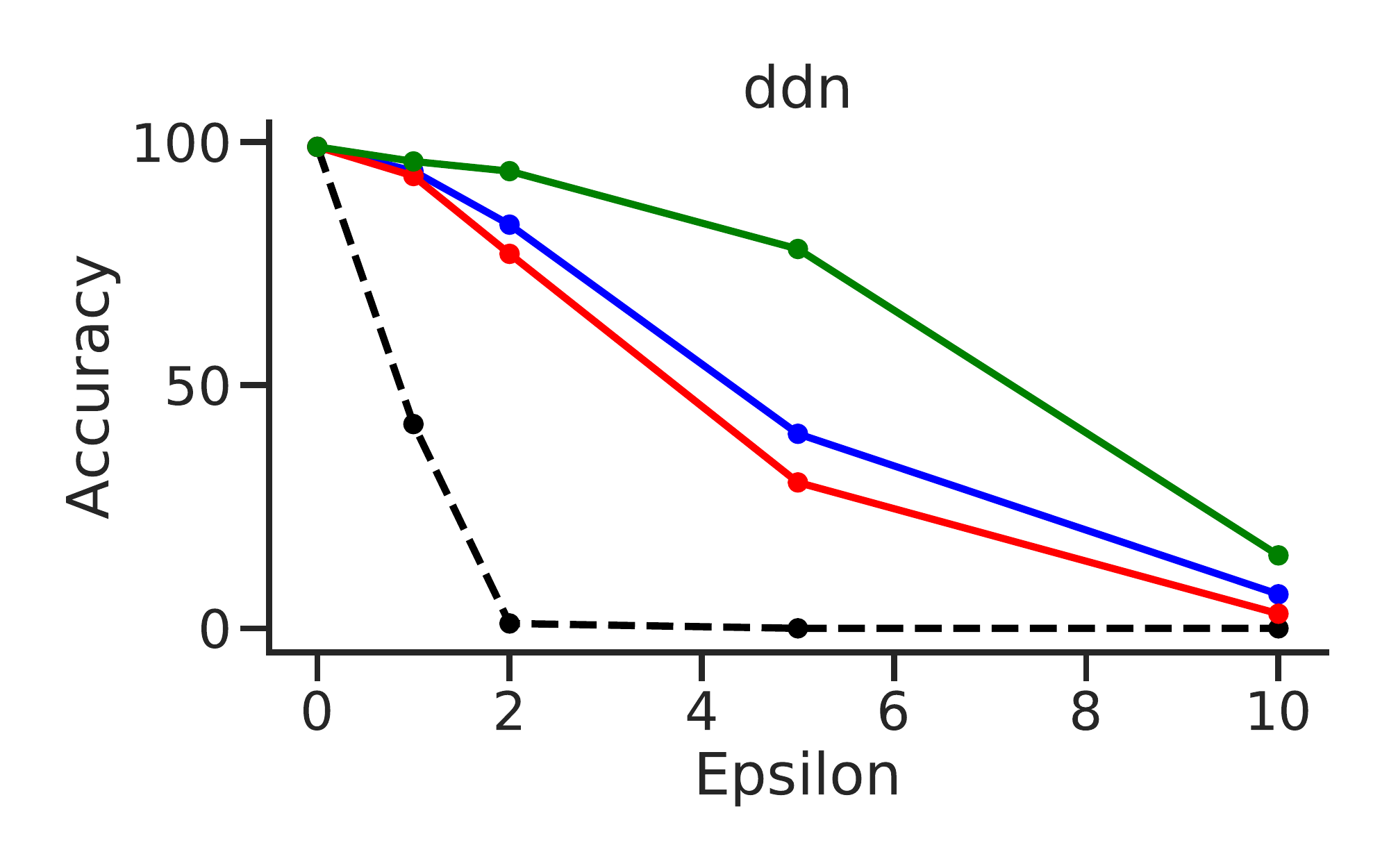}
\includegraphics[width=.24\linewidth]{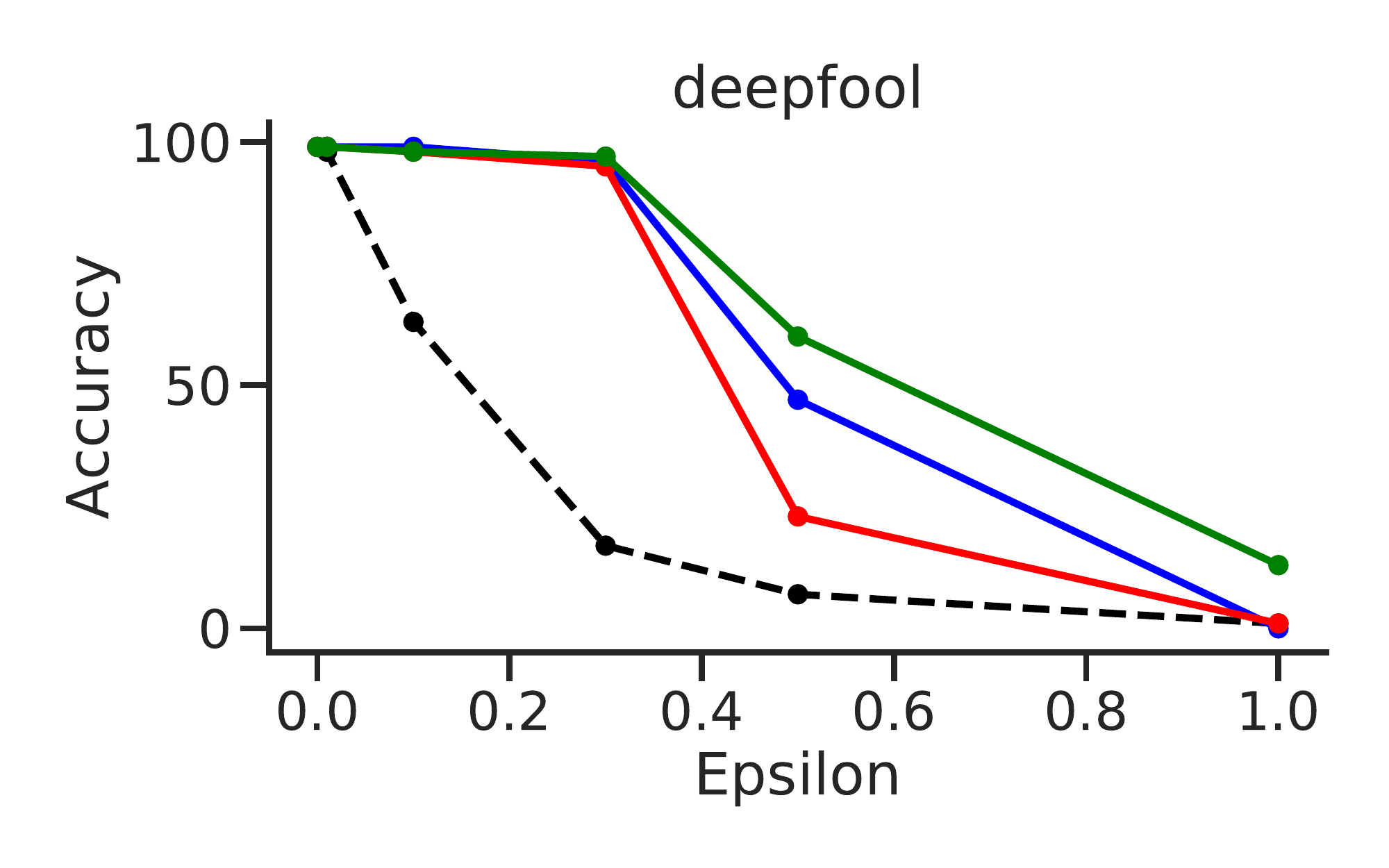}
\includegraphics[width=.24\linewidth]{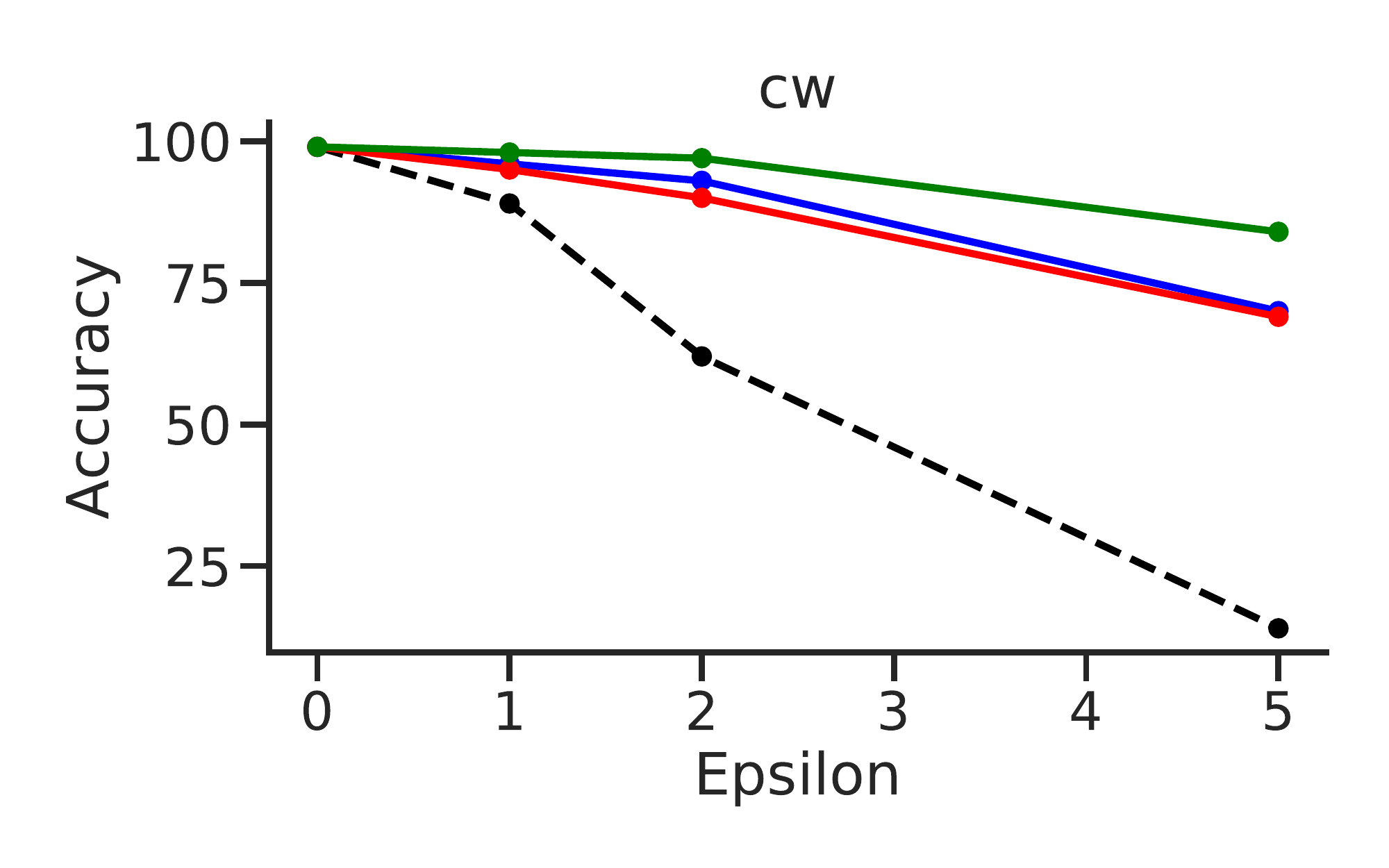}
\caption{Comparison of adversarial accuracy of different methods against white-box attacks on MNIST dataset with ResNet18 architecture.}
\label{fig_supp_mnist_eps_lineplots}
\end{figure}

\begin{figure}[hbt!]
\centering
\includegraphics[width=.24\linewidth]{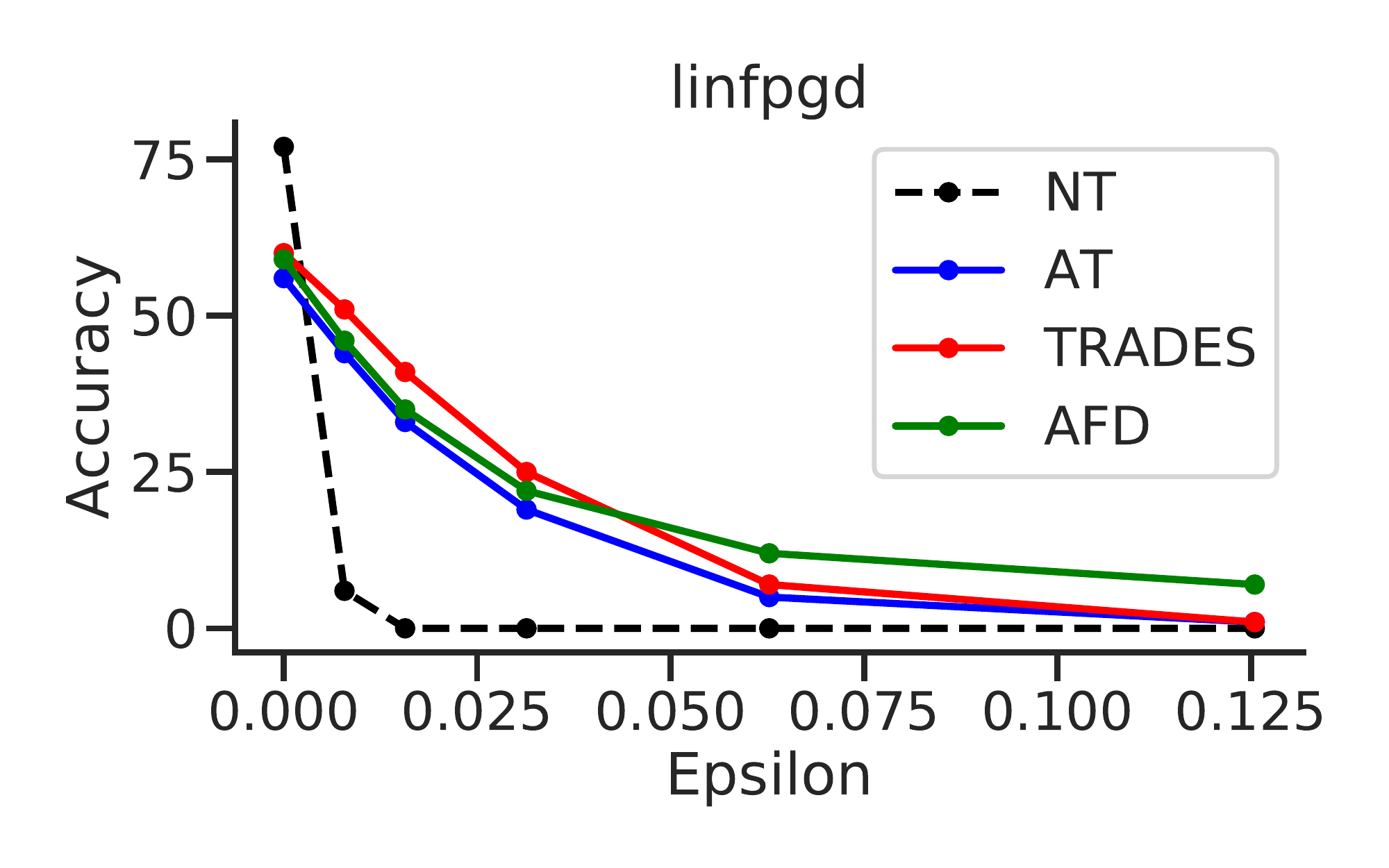}
\includegraphics[width=.24\linewidth]{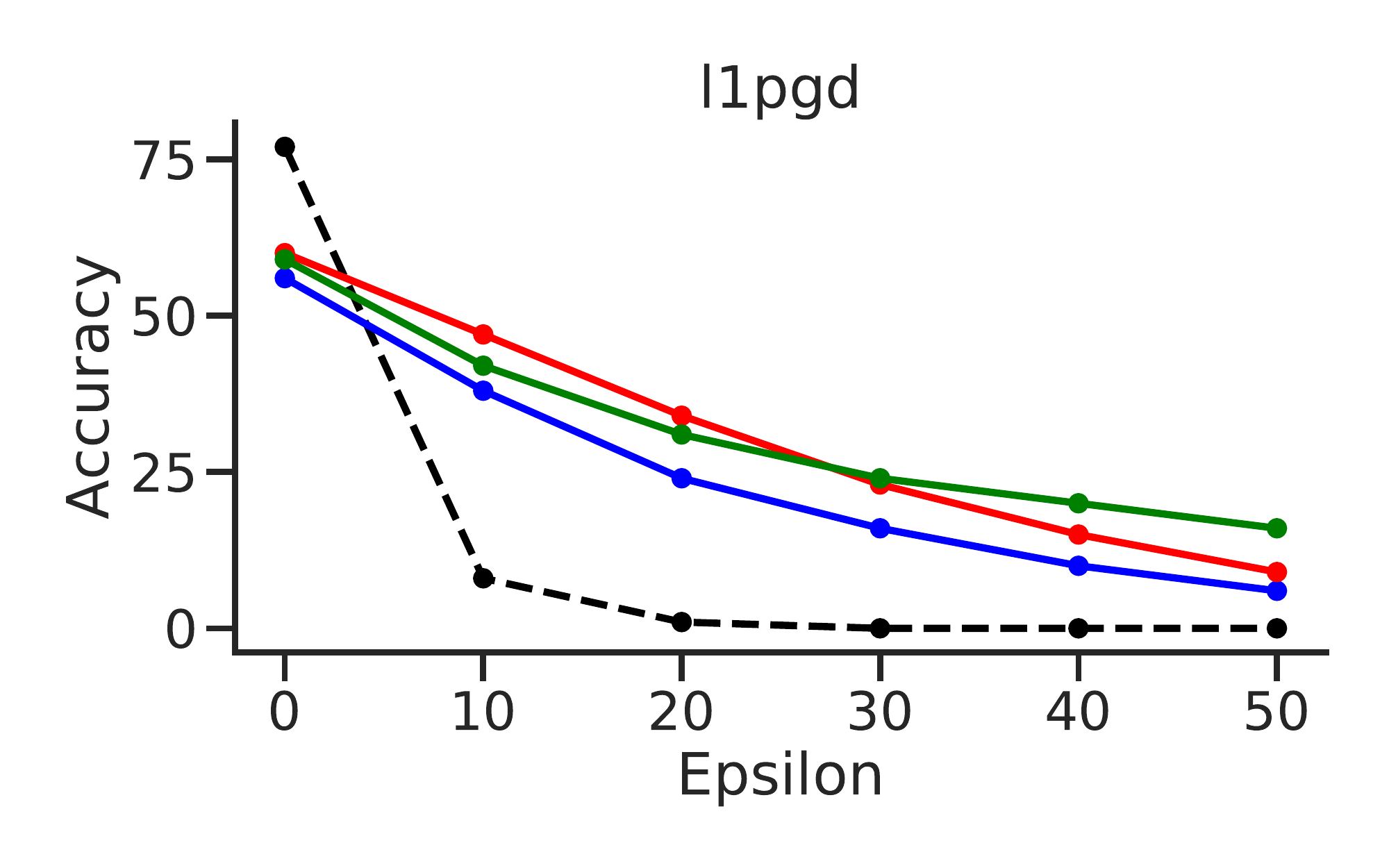}
\includegraphics[width=.24\linewidth]{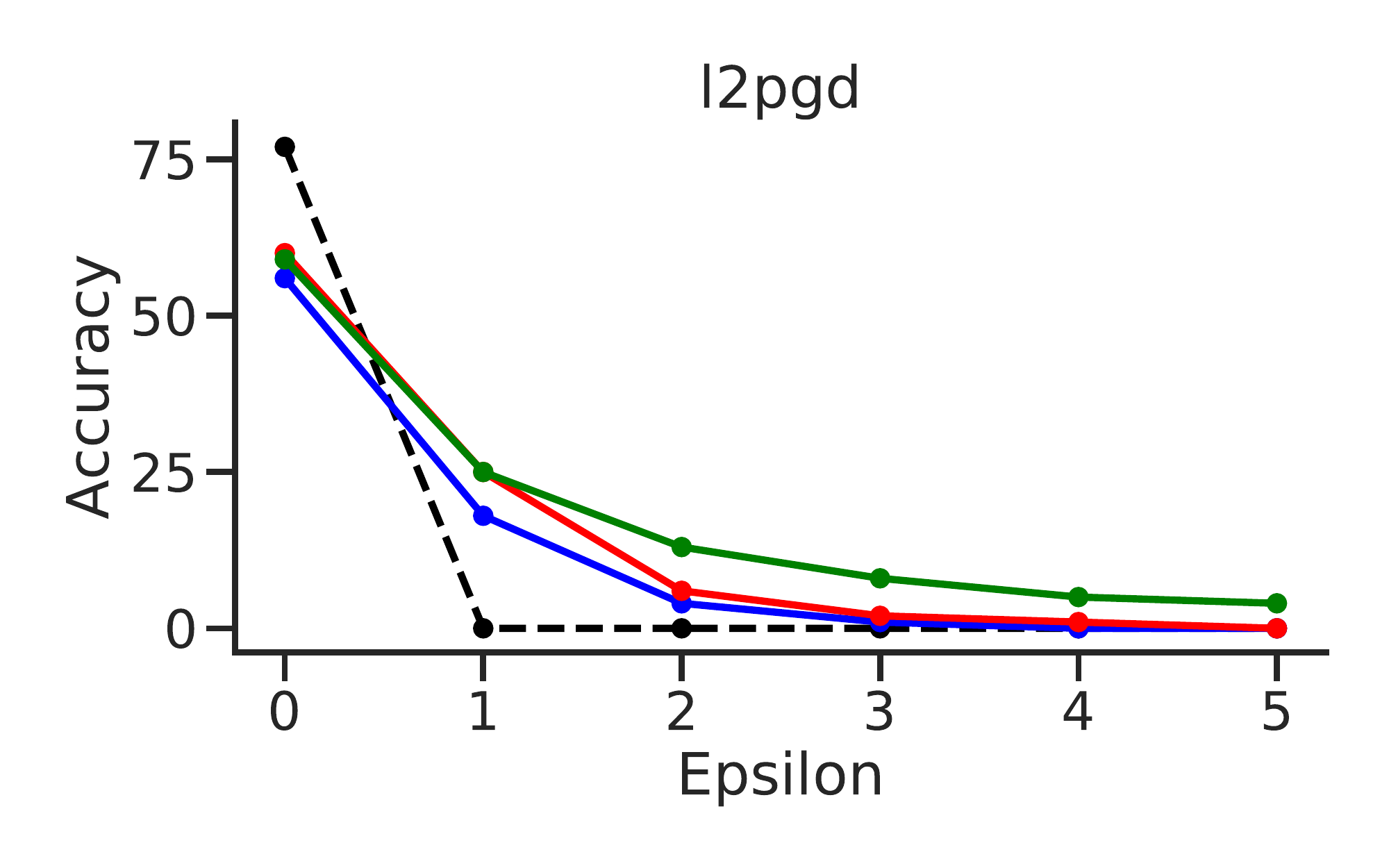}
\includegraphics[width=.24\linewidth]{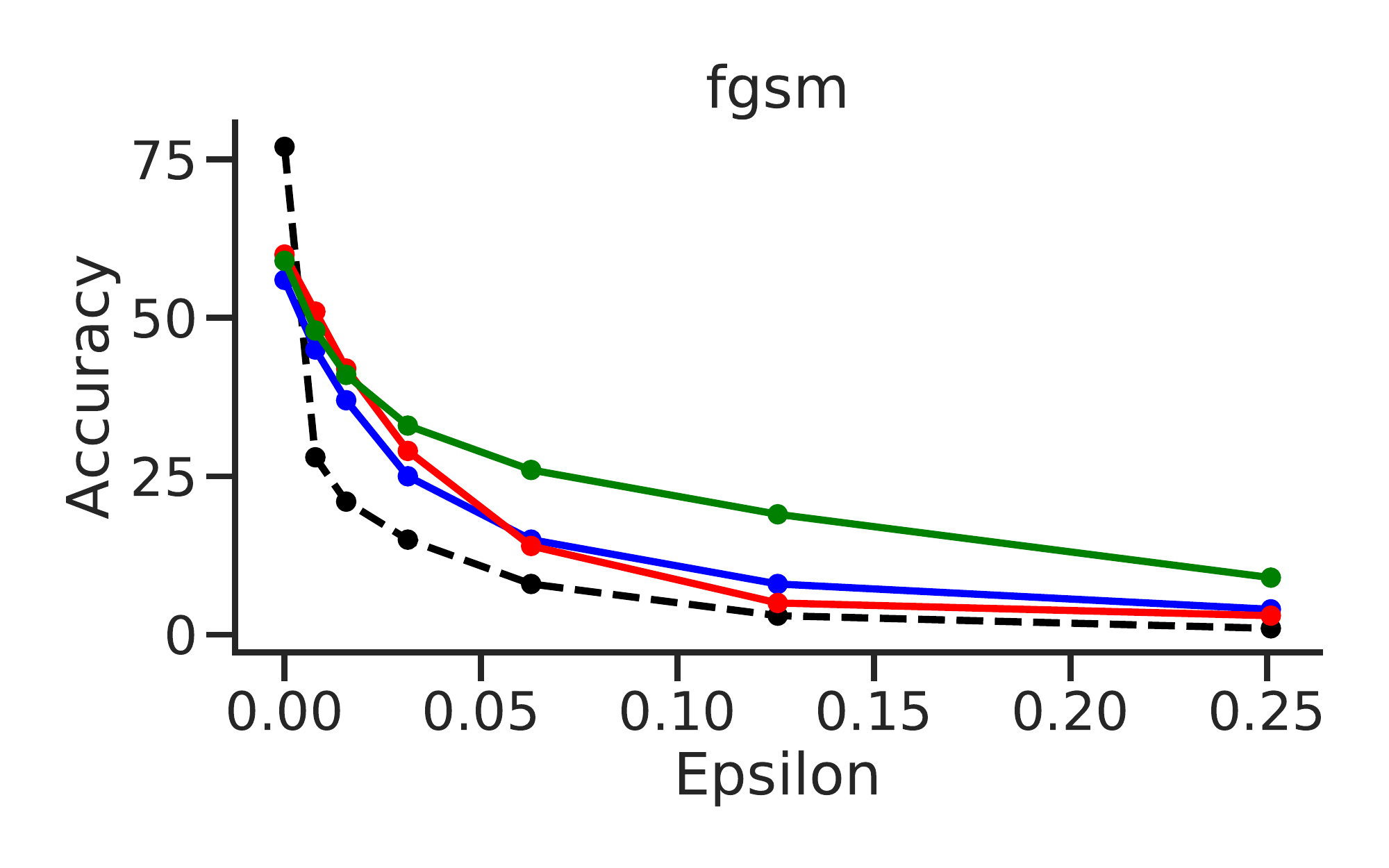}
\includegraphics[width=.24\linewidth]{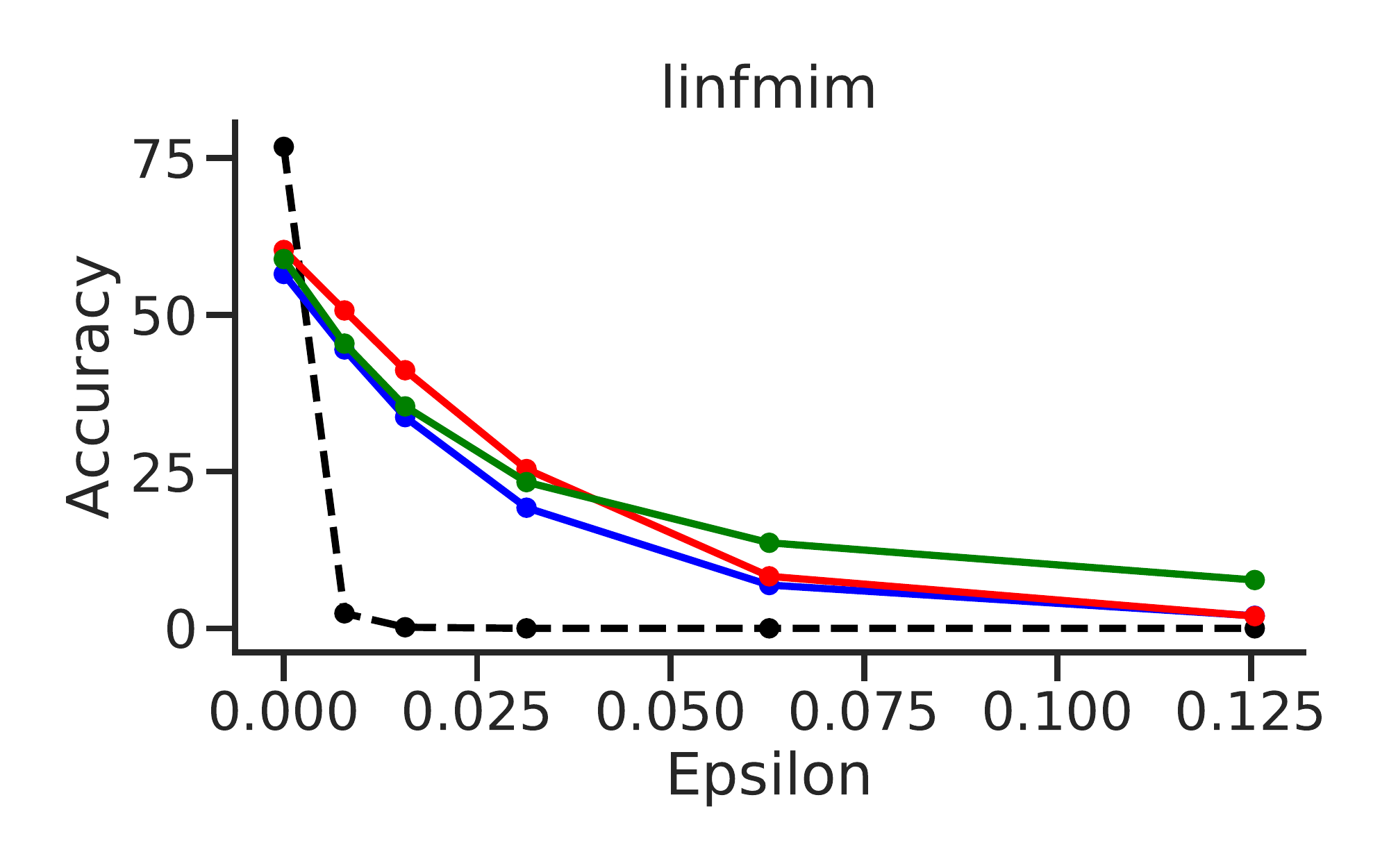}
\includegraphics[width=.24\linewidth]{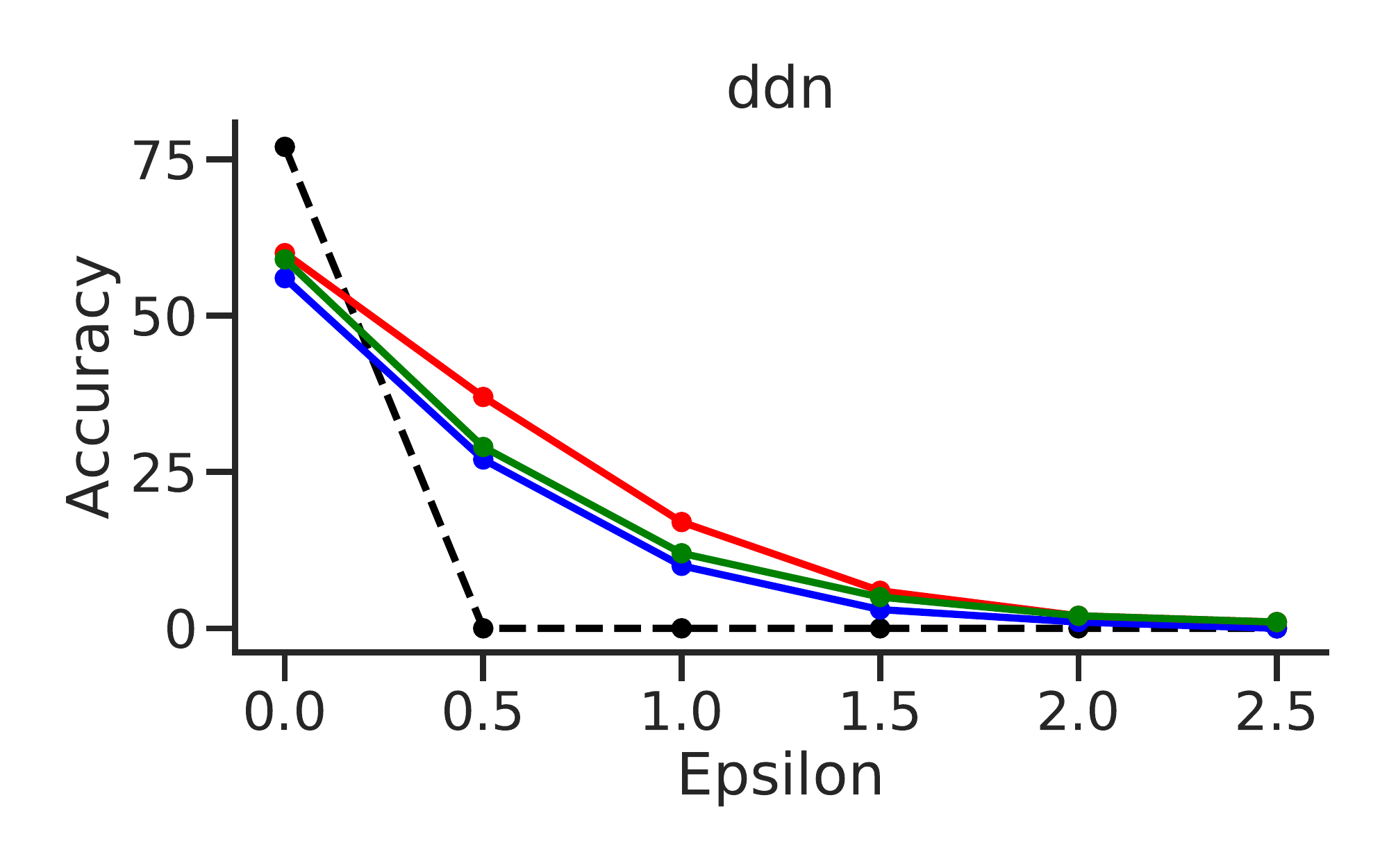}
\includegraphics[width=.24\linewidth]{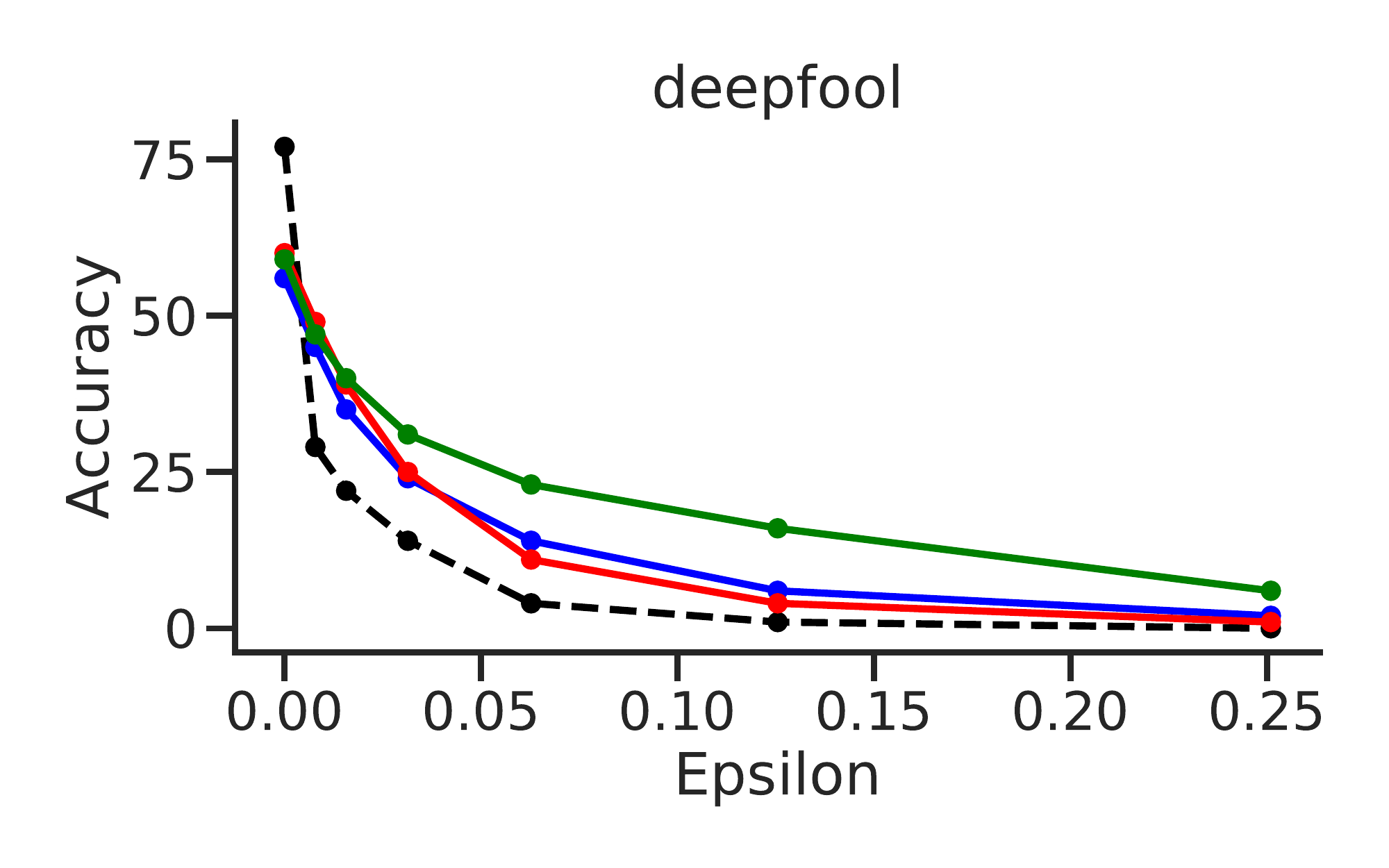}
\includegraphics[width=.24\linewidth]{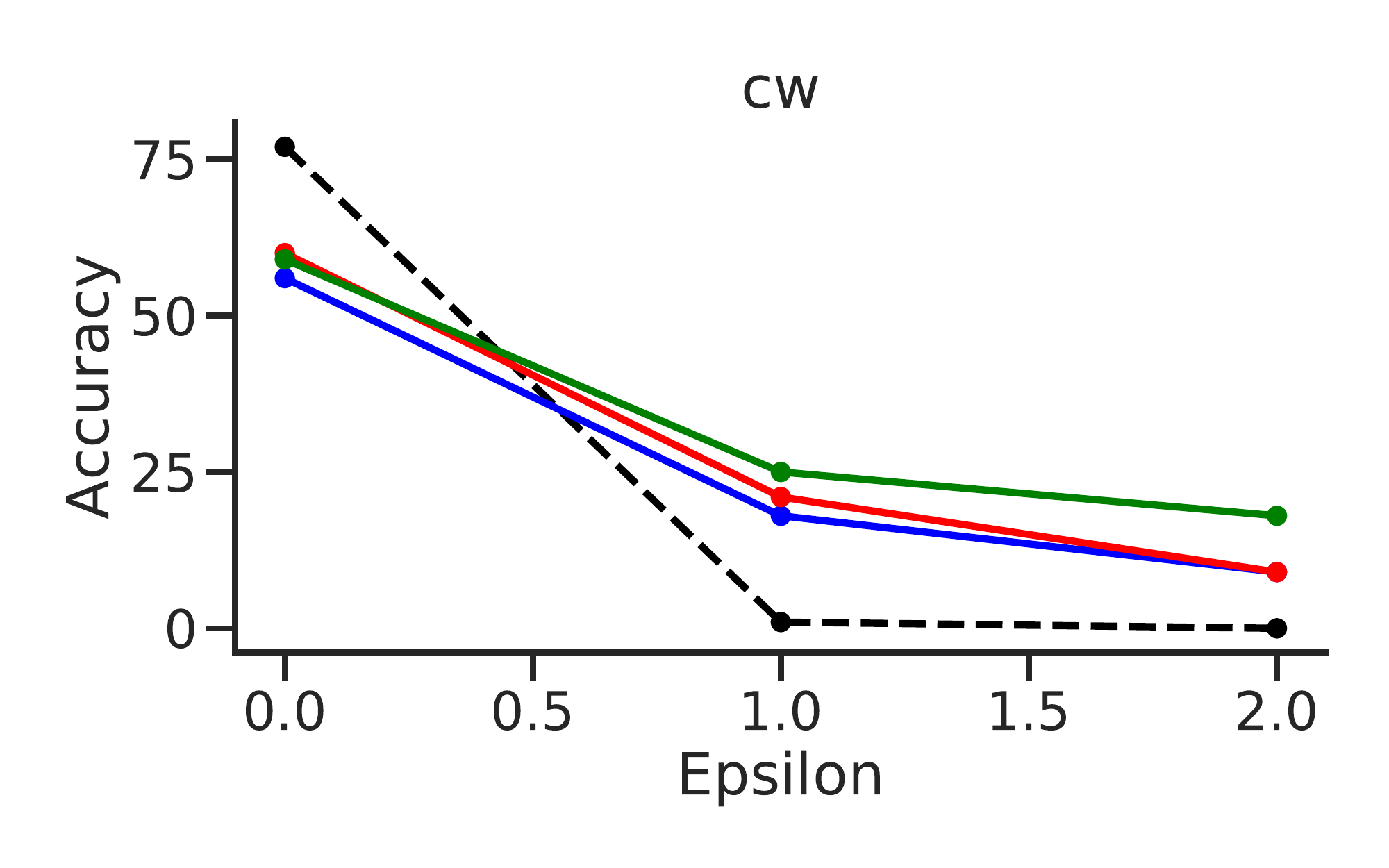}
\caption{Comparison of adversarial accuracy of different methods against white-box attacks on CIFAR100 dataset with ResNet18 architecture.}
\label{fig_supp_cifar100_rn18_eps_lineplots}
\end{figure}

\begin{figure}[hbt!]
\centering
\includegraphics[width=.24\linewidth]{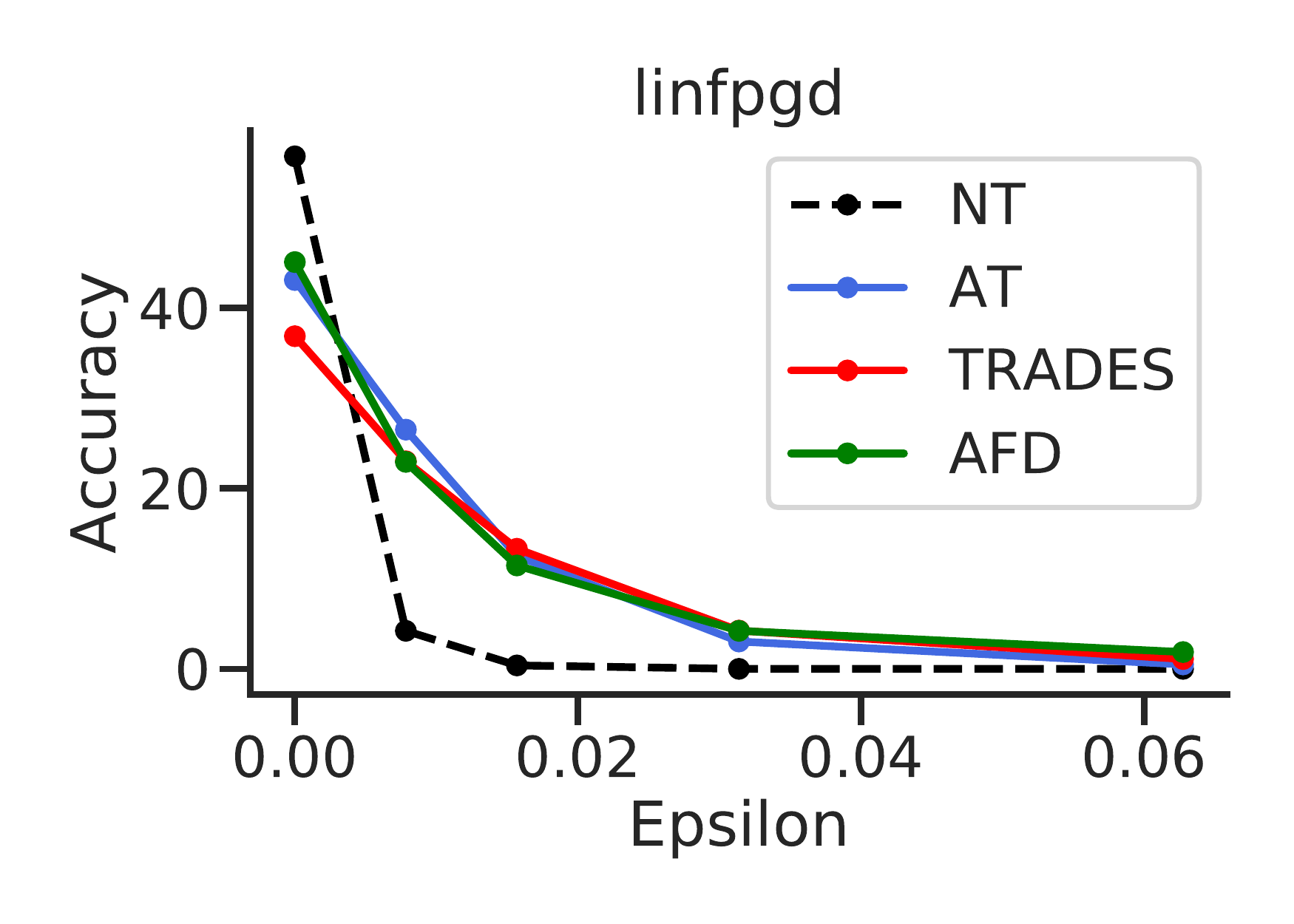}
\includegraphics[width=.24\linewidth]{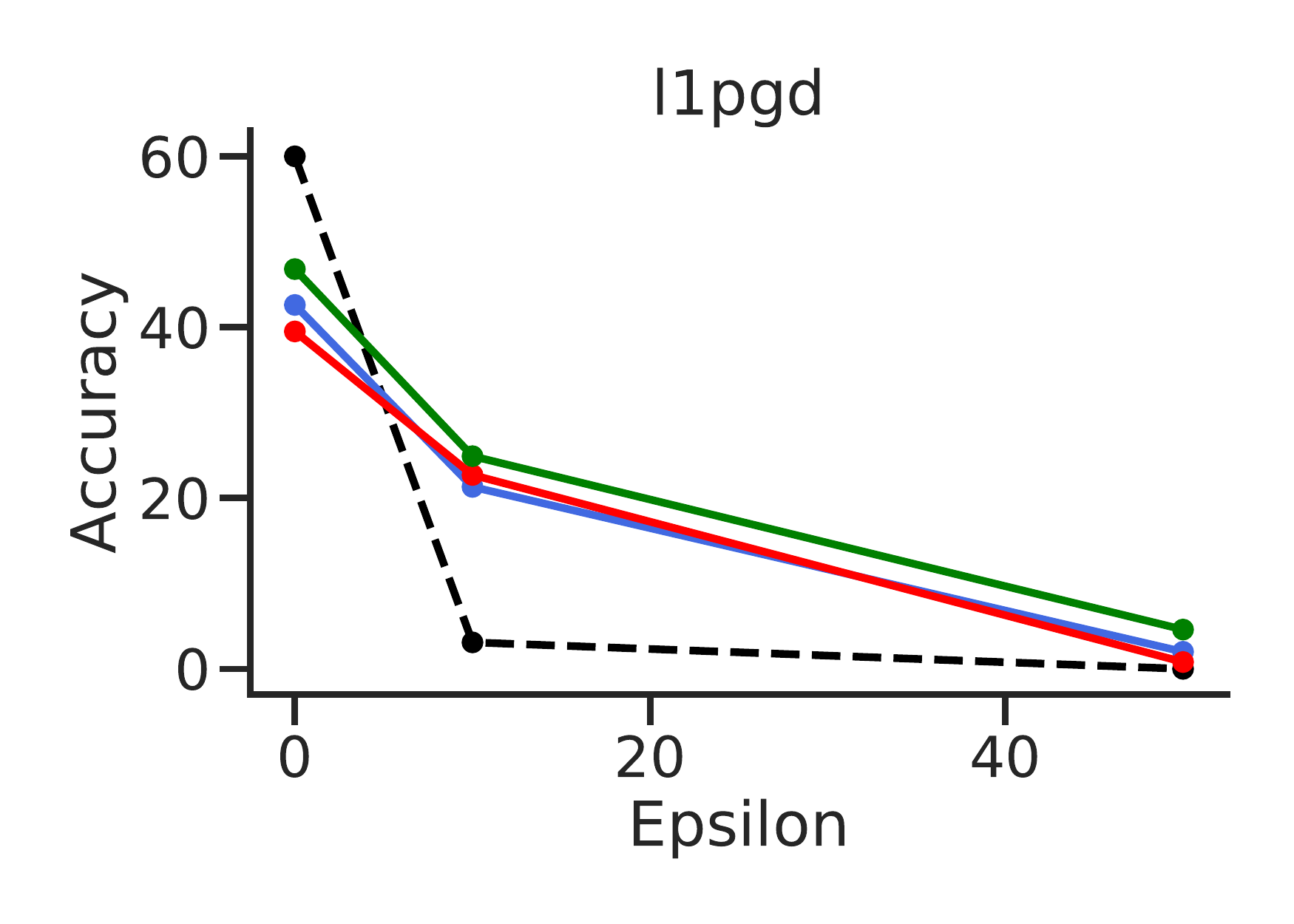}
\includegraphics[width=.24\linewidth]{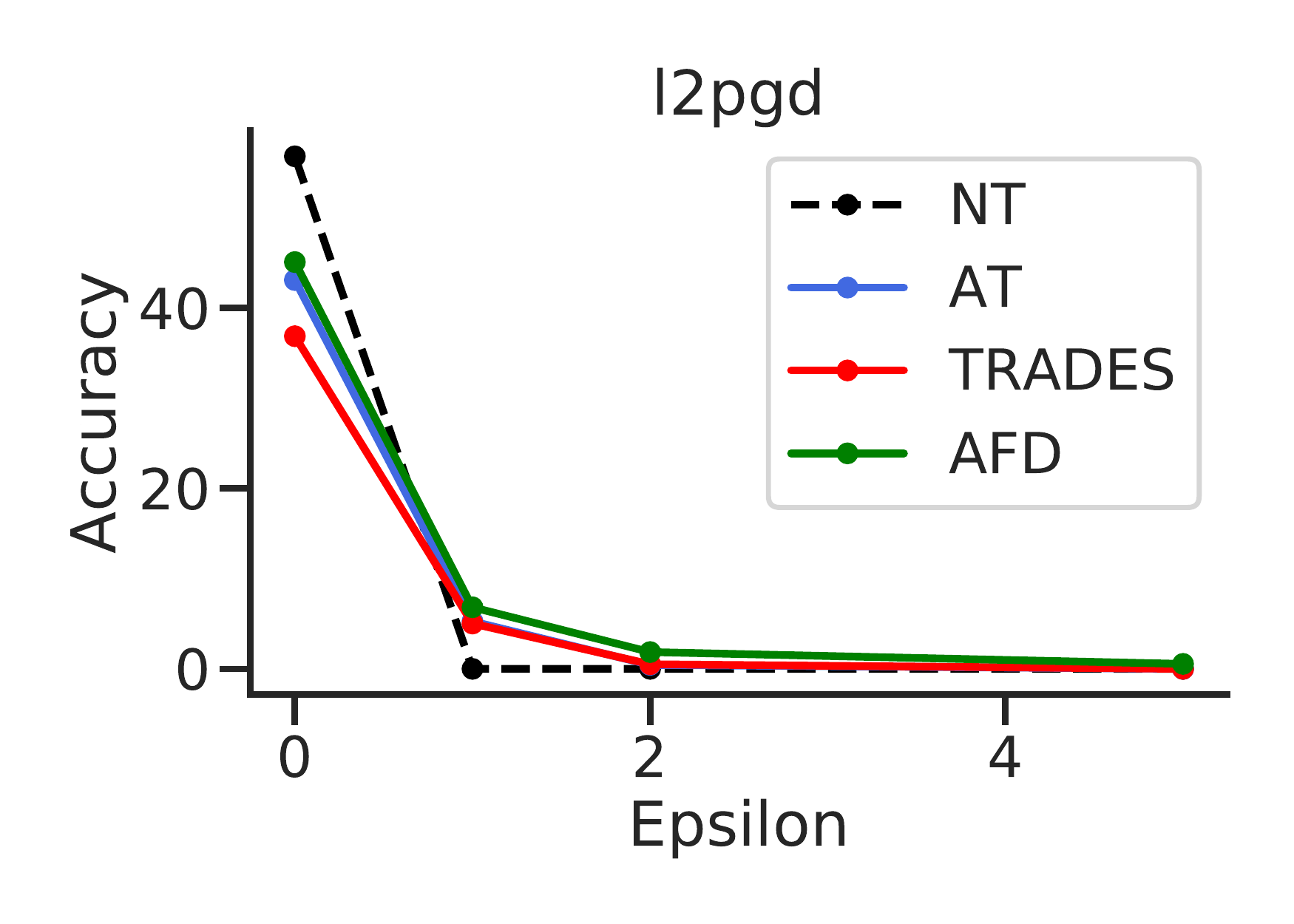}
\includegraphics[width=.24\linewidth]{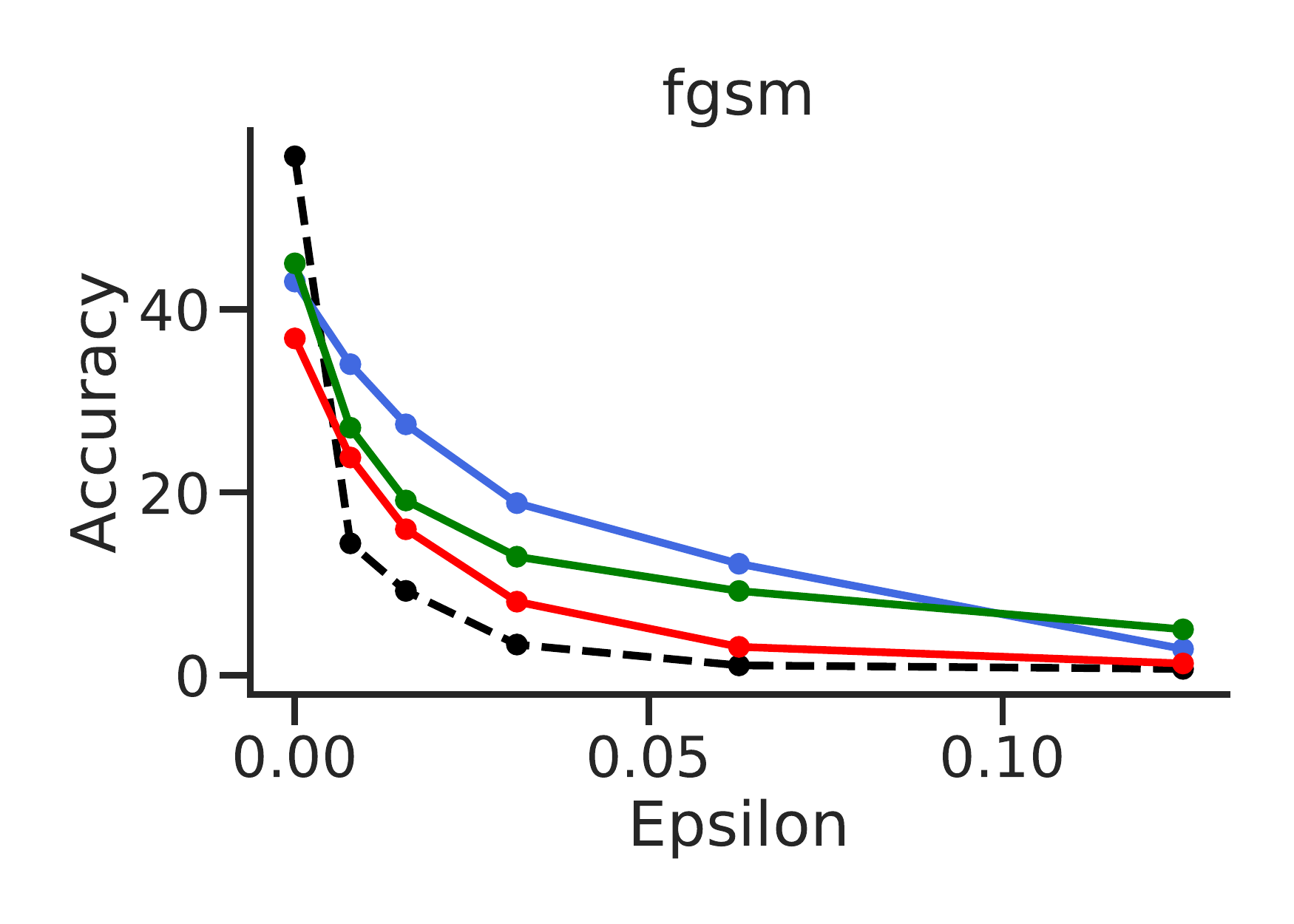}
\includegraphics[width=.24\linewidth]{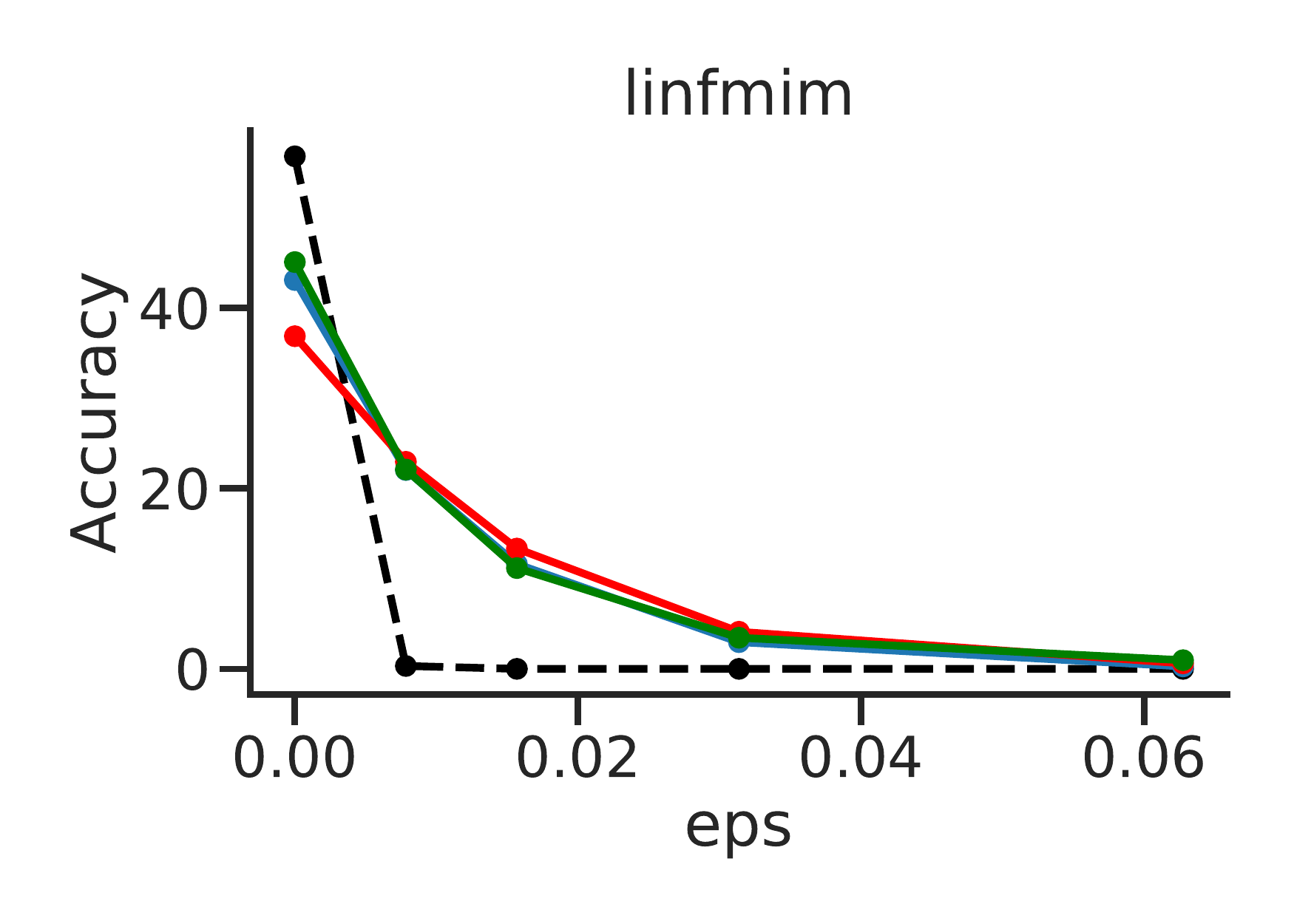}
\includegraphics[width=.24\linewidth]{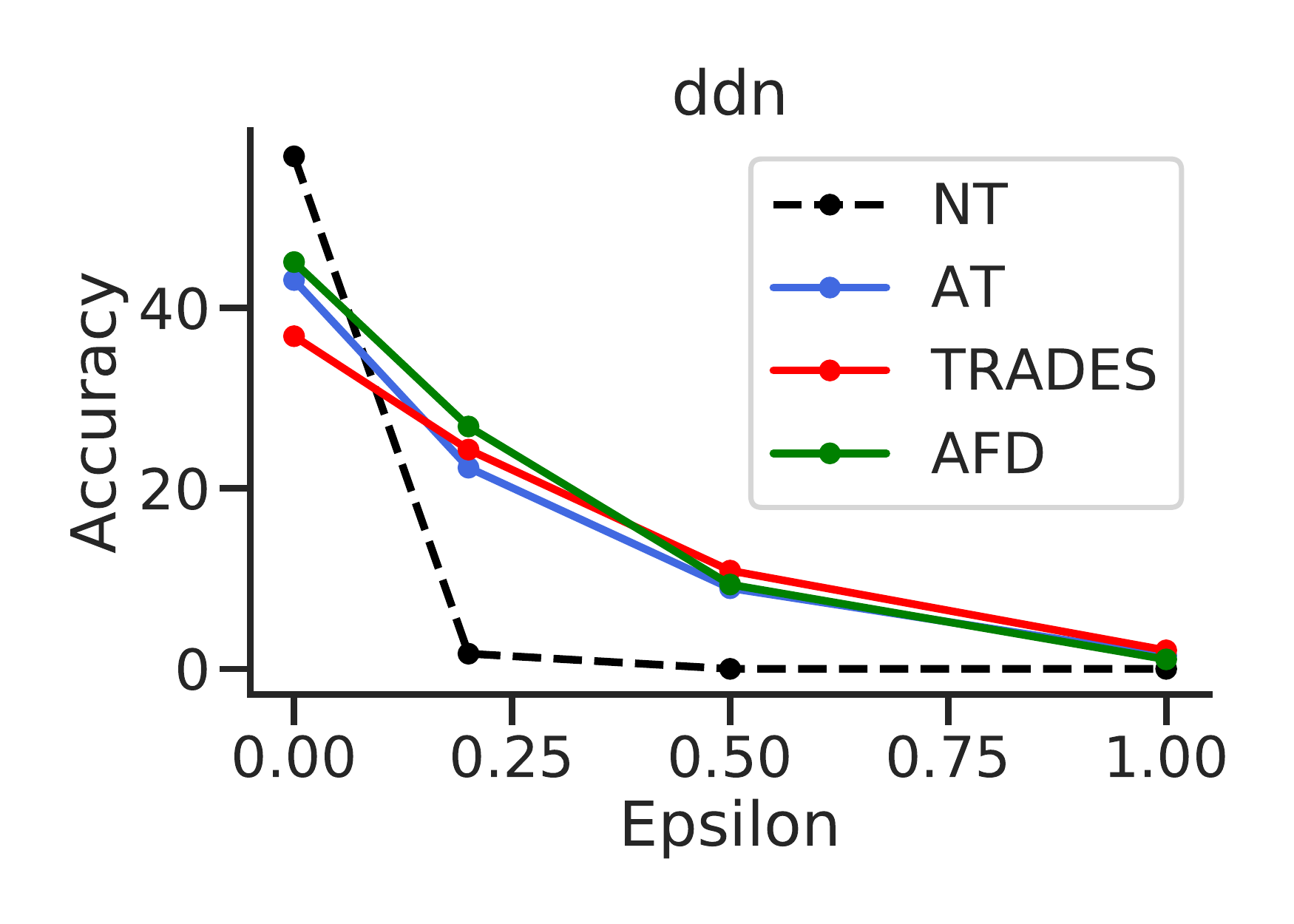}
\includegraphics[width=.24\linewidth]{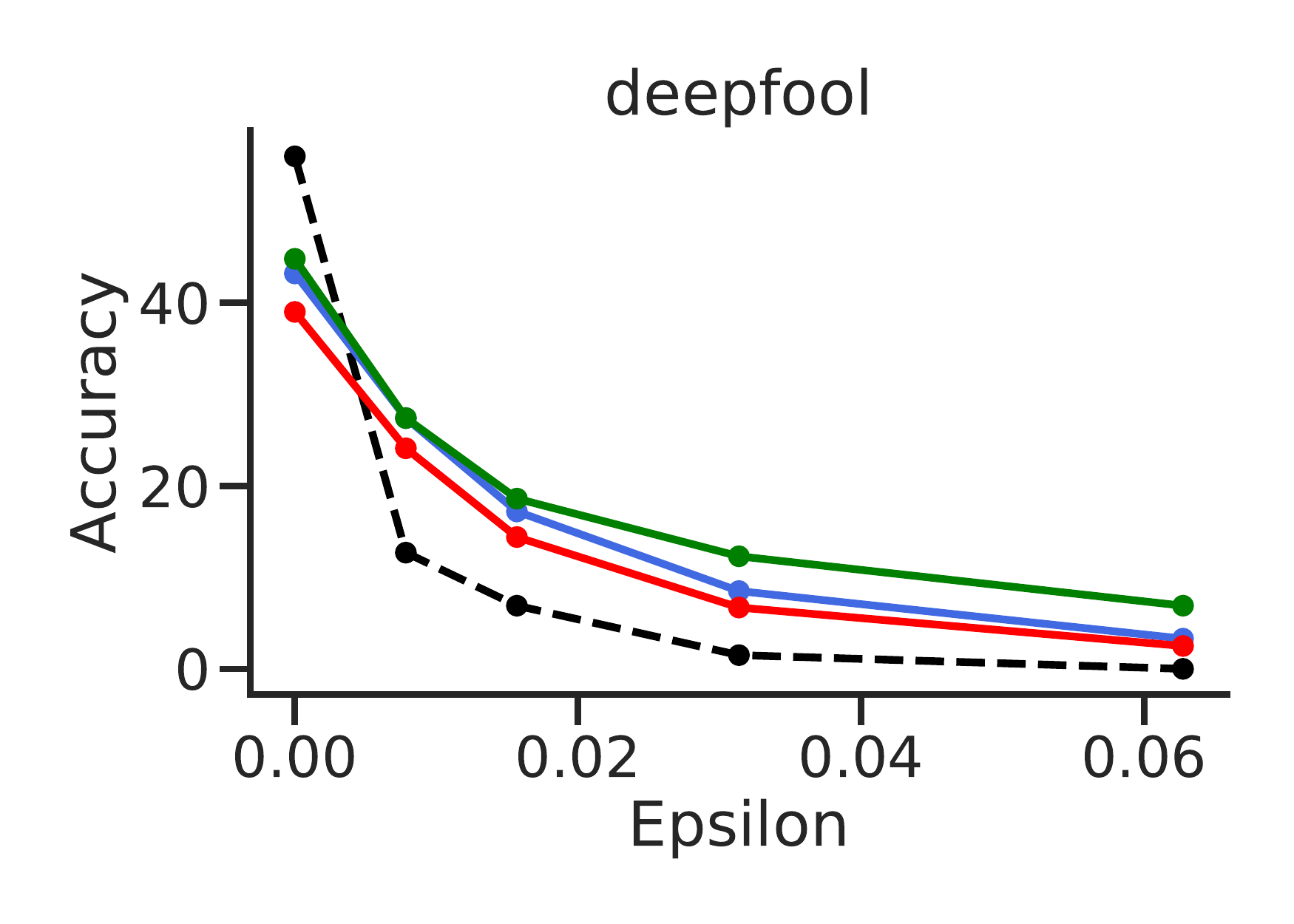}
\includegraphics[width=.24\linewidth]{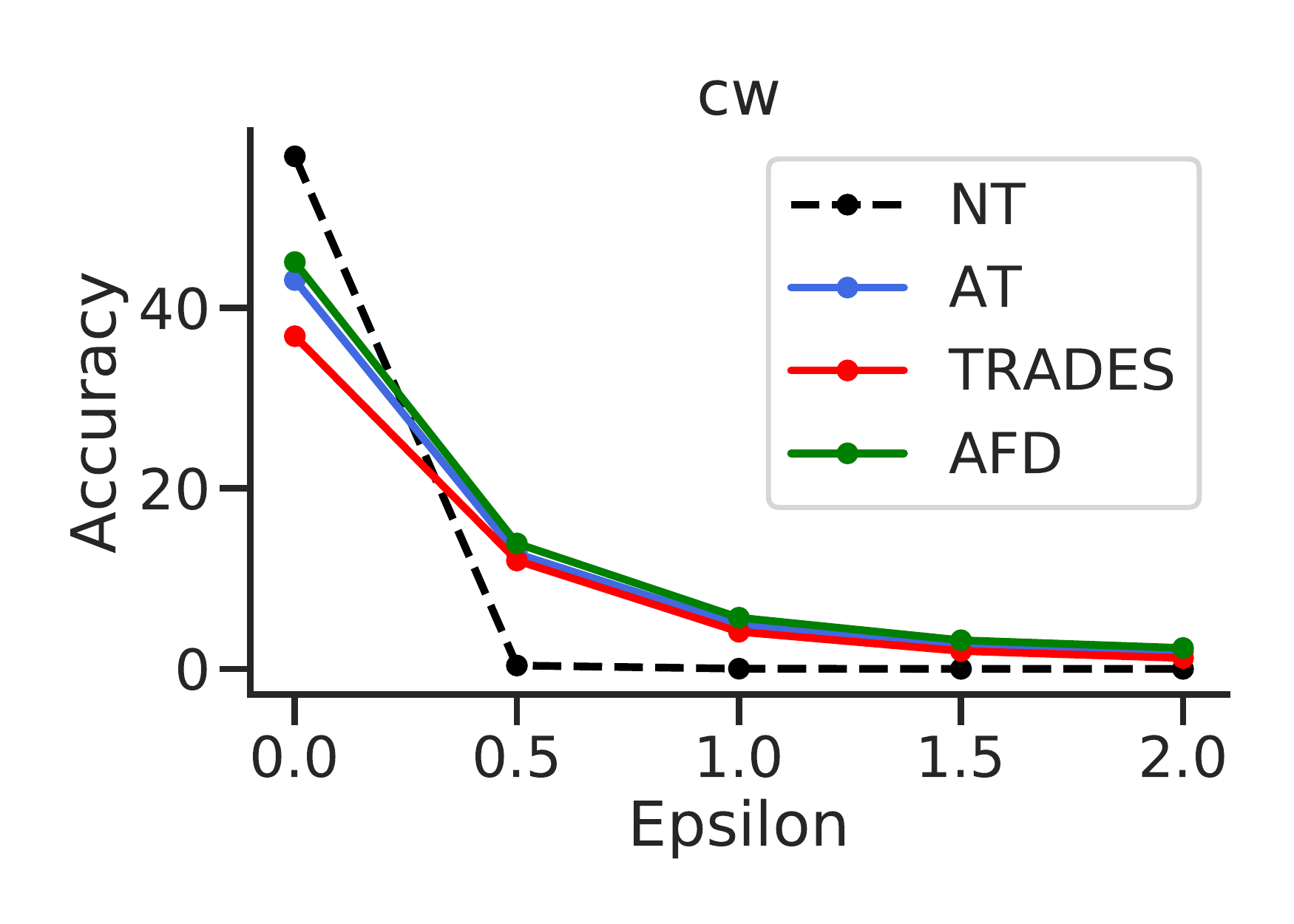}
\caption{Comparison of adversarial accuracy of different methods against white-box attacks on Tiny-Imagenet dataset with ResNet18 architecture.}
\label{fig_supp_cifar100_rn18_eps_lineplots}
\end{figure}

\begin{figure}[h]
\centering
\includegraphics[width=.24\linewidth]{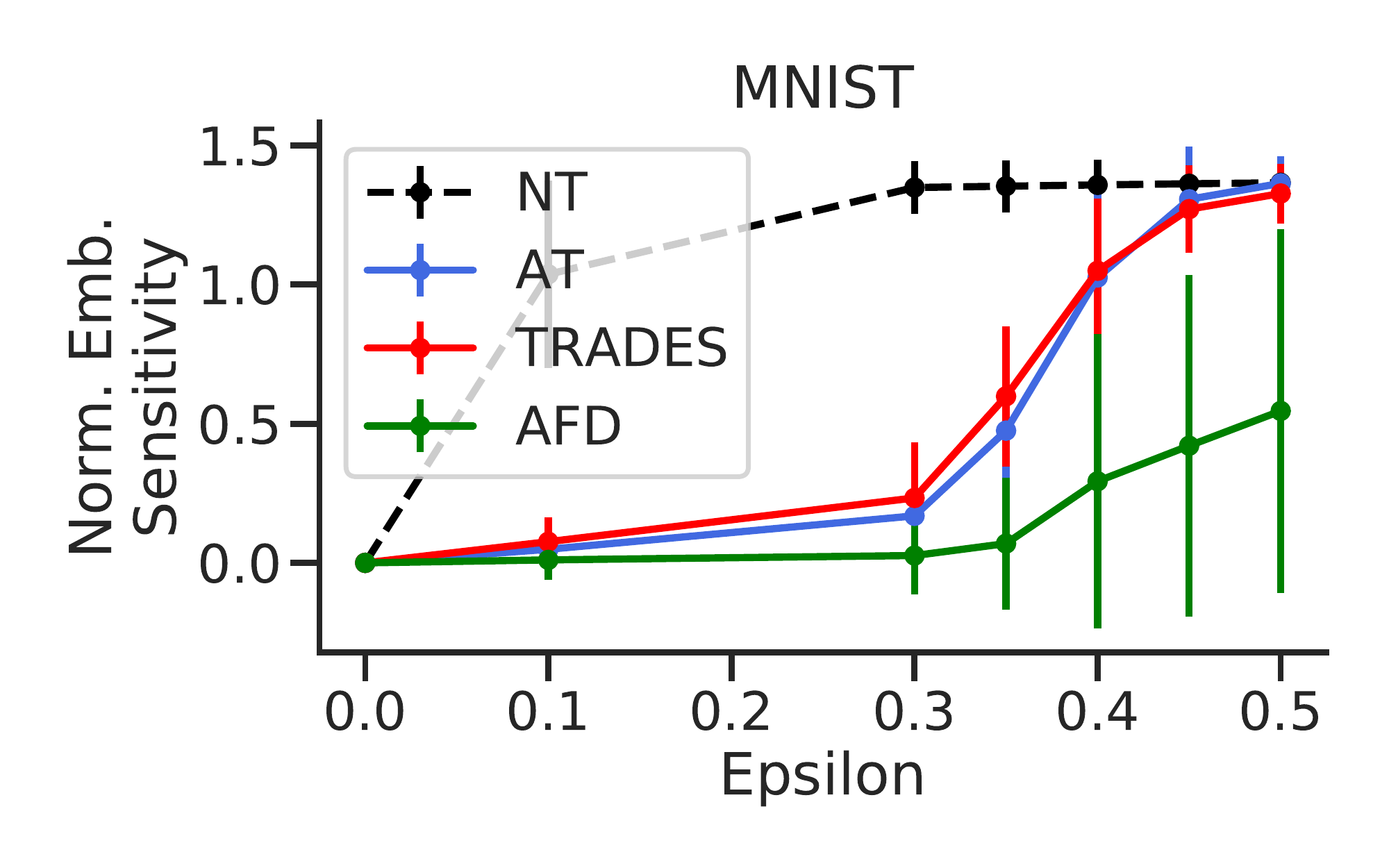}
\includegraphics[width=.24\linewidth]{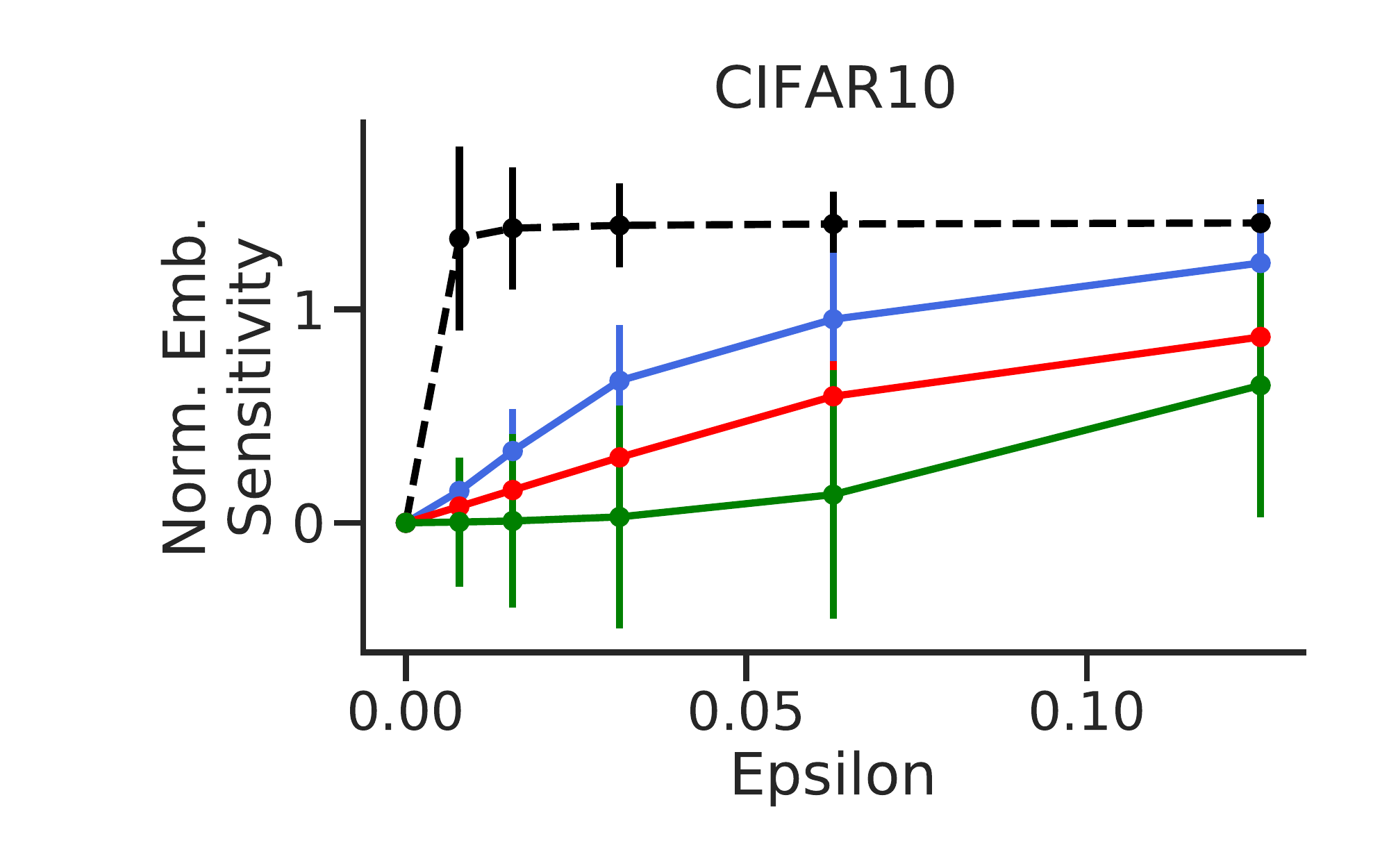}
\includegraphics[width=.24\linewidth]{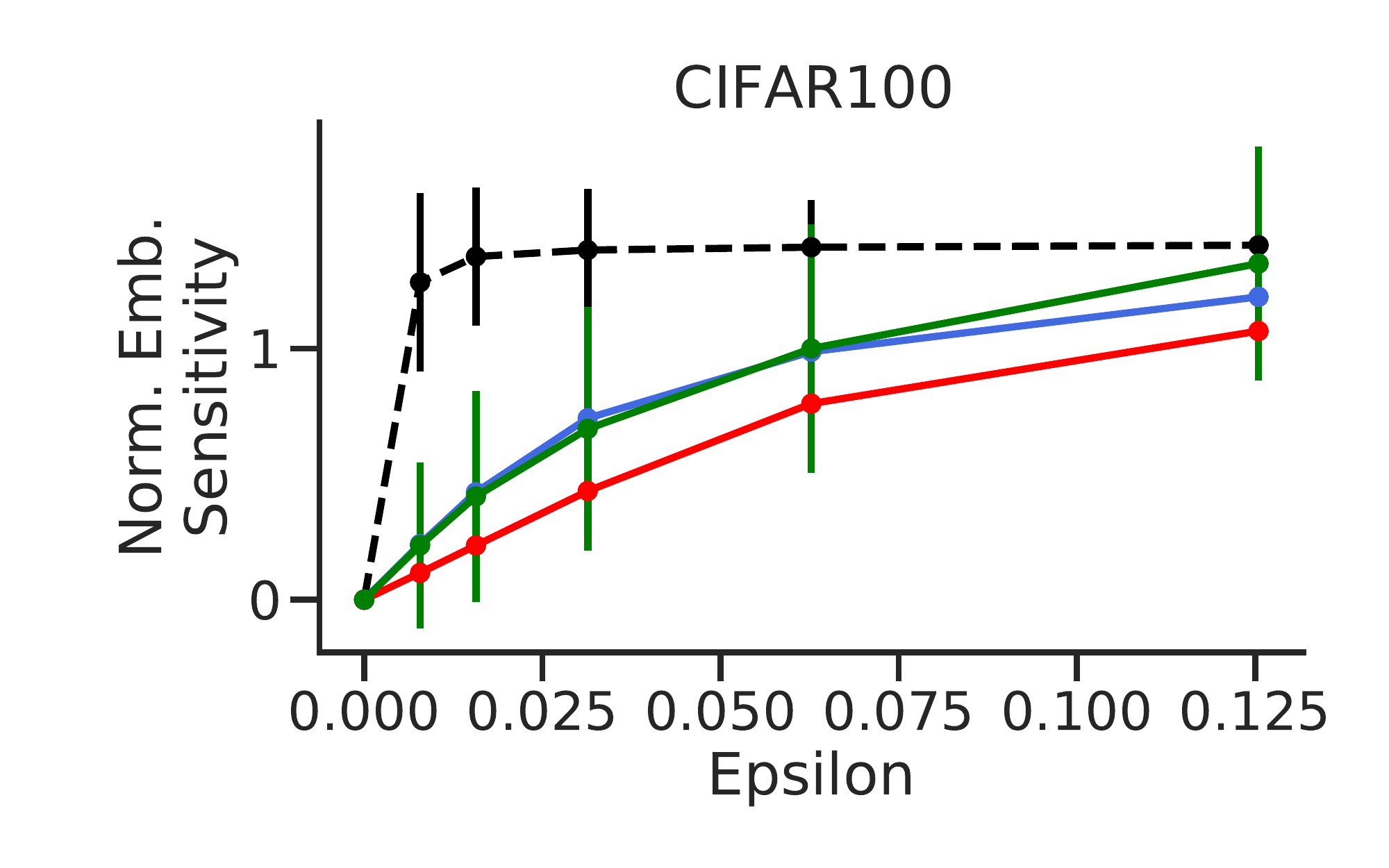}
\includegraphics[width=.225\linewidth]{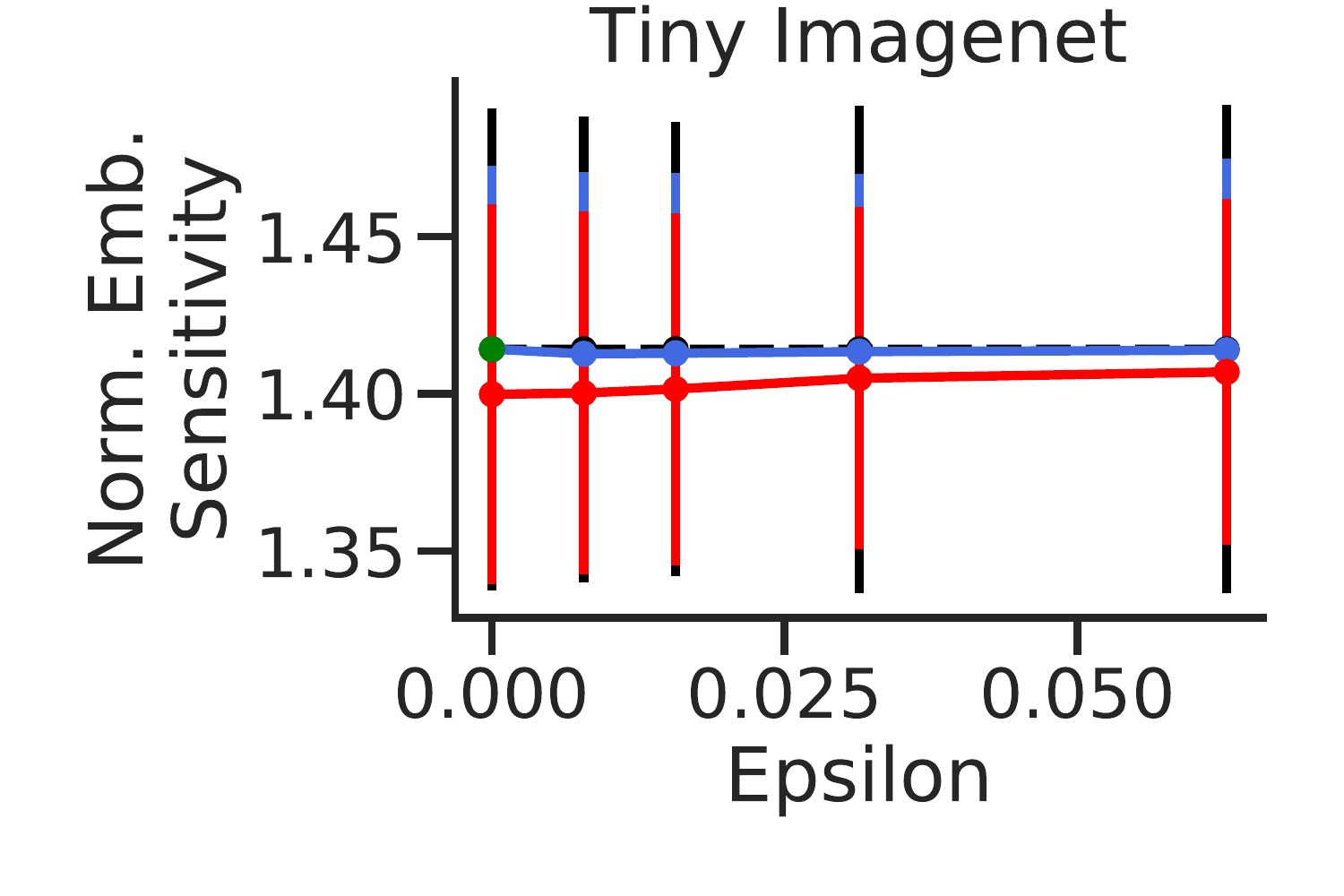}

\caption{Comparison of normalized feature sensitivity on test set of MNIST, CIFAR10, CIFAR100, and Tiny-Imagenet datasets under PGD-$L_\infty$ attack. For each image, we computed the normalized feature sensitivity as $\frac{\norm{F(x)-F(x^\prime)}_2}{\norm{F(x)}_2}$. Plots show the median sensitivity over test-set of each dataset. Error bars correspond to standard deviation. (dashed-black) naturally trained; (blue) adversarially trained; (red) TRADES; (green) AFD.}
\label{fig_supp_emb_sens}
\end{figure}

\begin{figure}[hbt!]
\centering
\includegraphics[width=.22\linewidth]{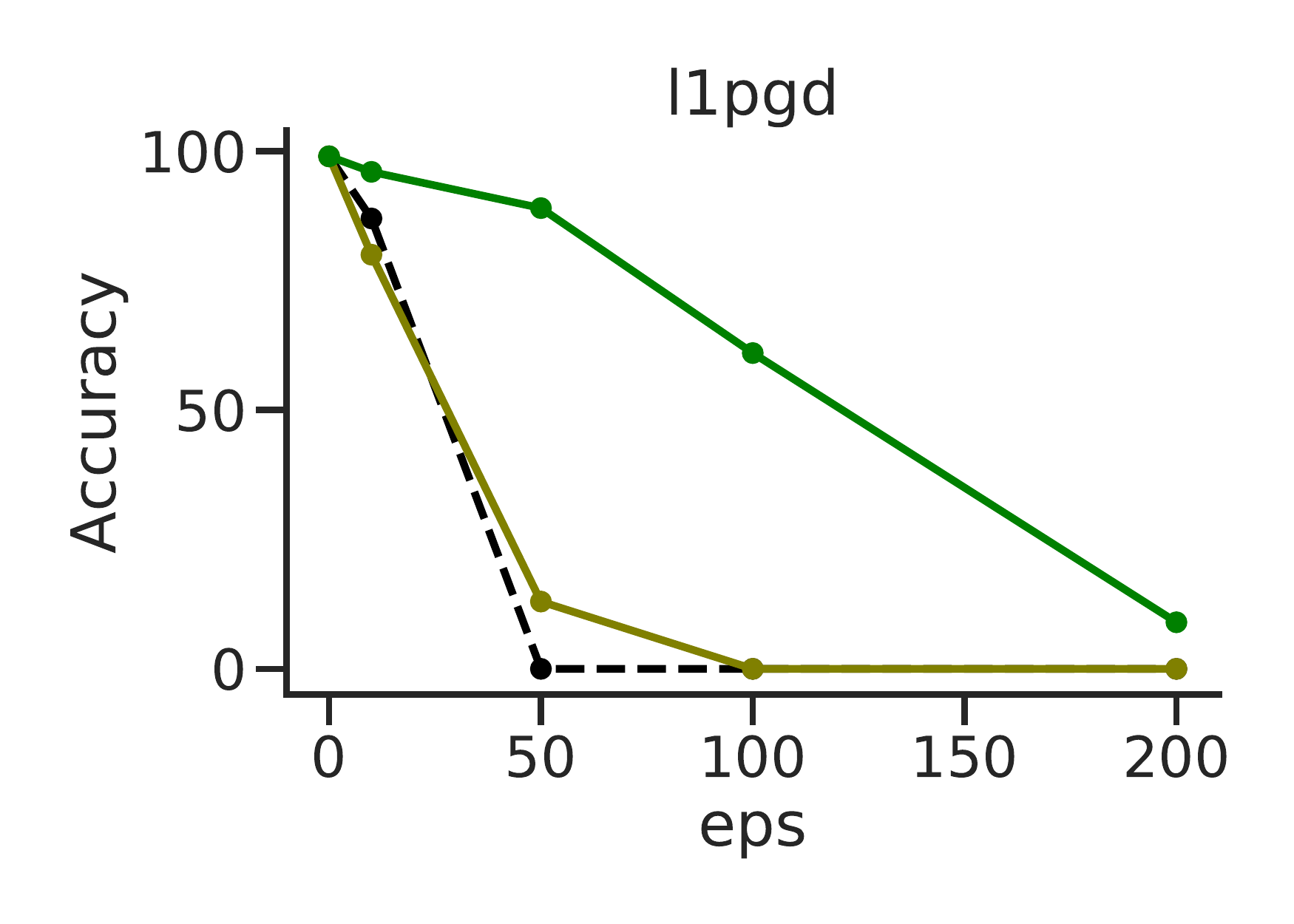}
\includegraphics[width=.22\linewidth]{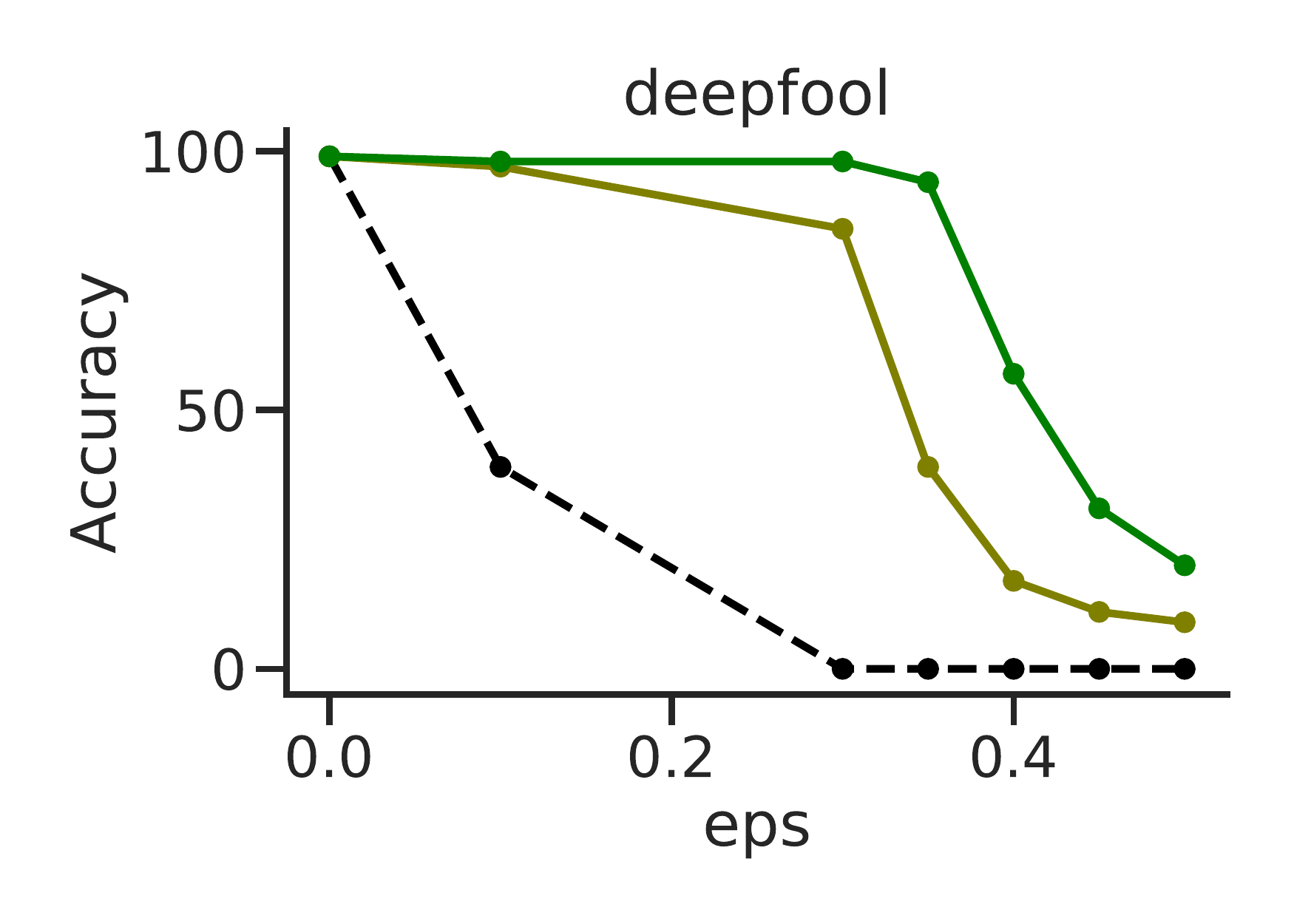}
\includegraphics[width=.23\linewidth]{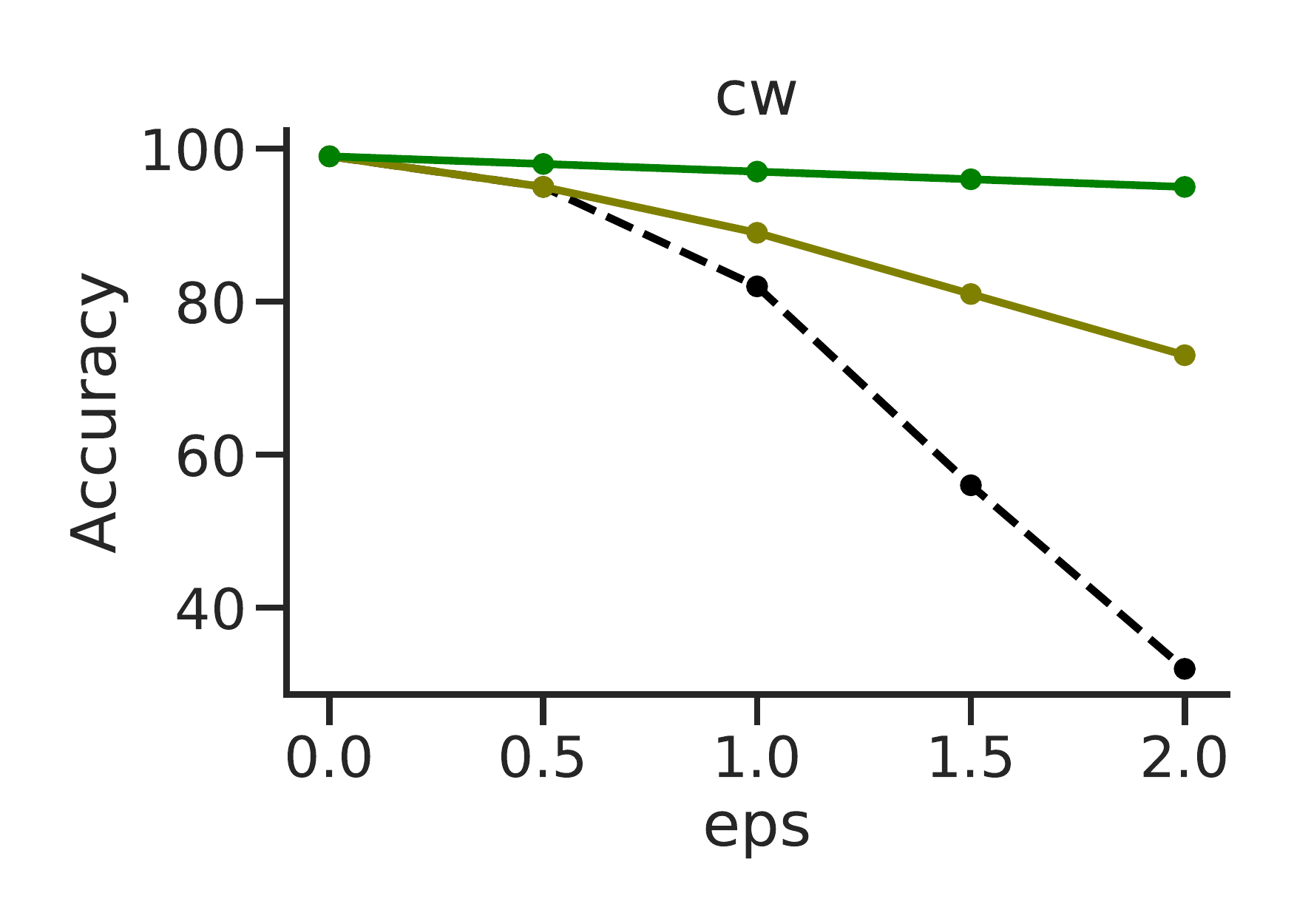}
\includegraphics[width=.3\linewidth]{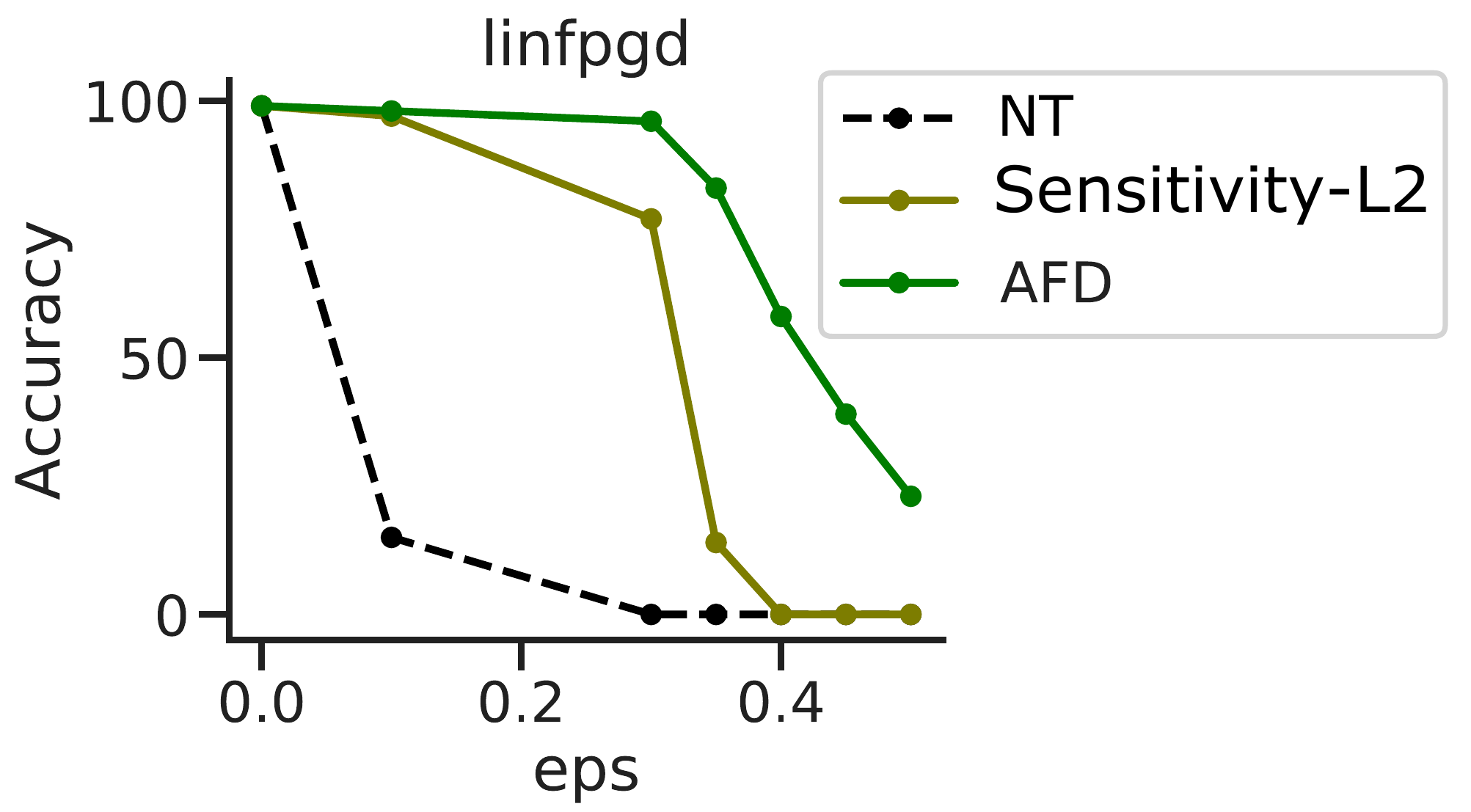}
\caption{Comparison of adversarial accuracy of AFD and representation matching against white-box attacks on MNIST dataset with ResNet18 architecture.}
\label{fig_supp_mnist_eps_lineplots_embmatch}
\end{figure}

\begin{figure}[hbt!]
\centering
\includegraphics[width=1.\linewidth]{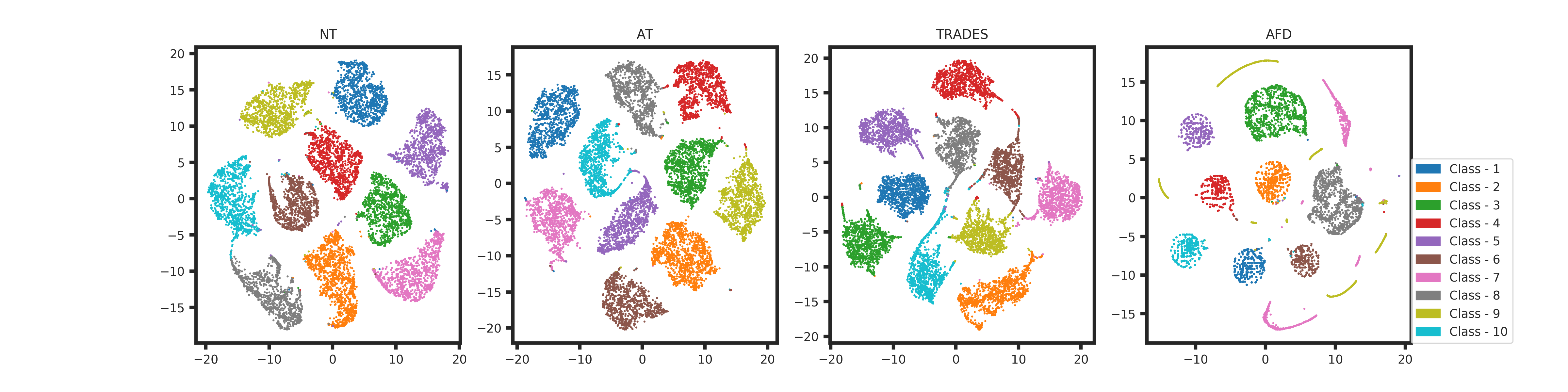}
\includegraphics[width=1.\linewidth]{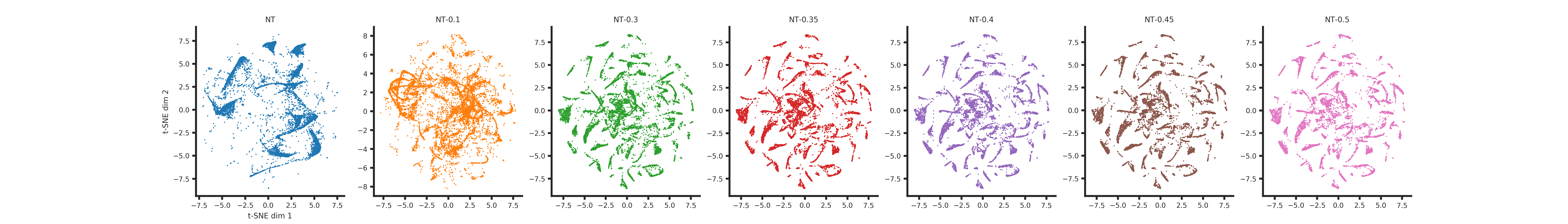}
\includegraphics[width=1.\linewidth]{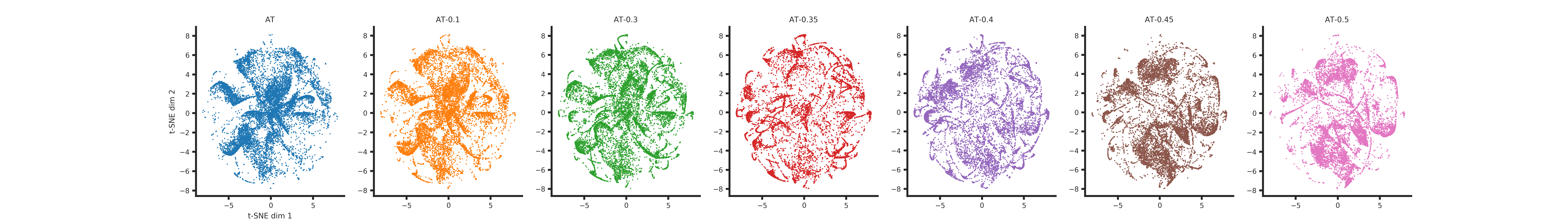}
\includegraphics[width=1.\linewidth]{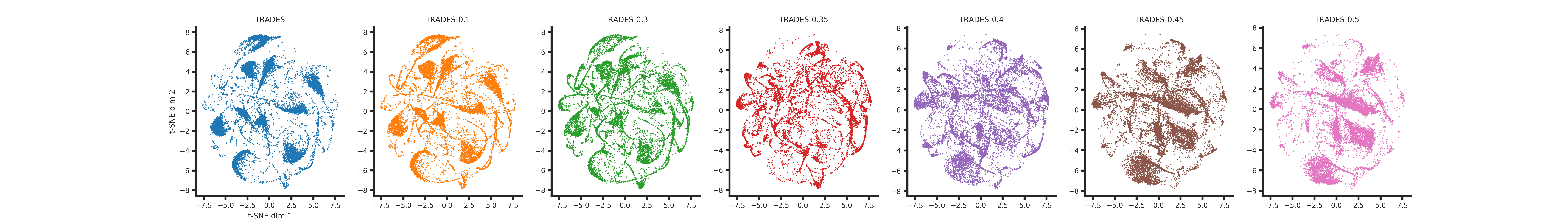}
\includegraphics[width=1.\linewidth]{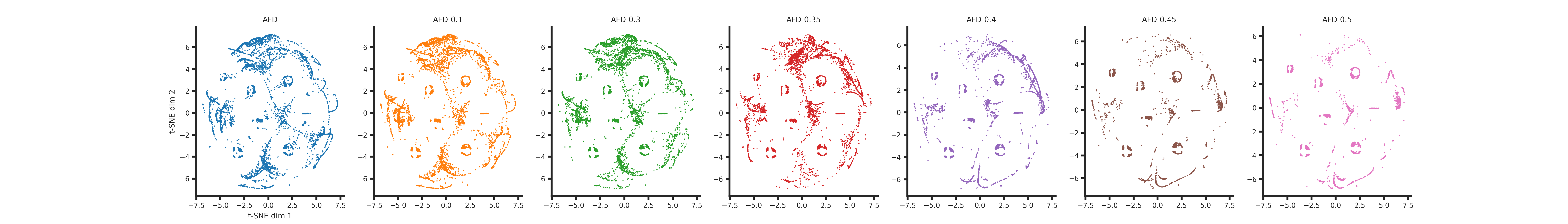}
\caption{Scatter plot of 2-dimensional t-SNE projection \cite{maaten2008visualizing} of the representations derived from training the ResNet18 architecture on MNIST dataset. (top row) t-SNE projection of representations of natural images for networks trained with different methods. Each point corresponds to the representation of one of the images from the MNIST test-set. (rows 2 to 5) t-SNE projection of the representation of the natural and adversarial MNIST test-set images. Columns are sorted from left to right with the strength of the perturbation (left-most column corresponds to natural images and right-most column with highest tested perturbation). Perturbations are generated using PGD-$L_\infty$ attack. NT: naturally trained; AT: adversarially trained\cite{madry2017towards}; TRADES: \cite{Zhang2019}; AFD: adversarial feature desensitization.}
\label{fig_supp_tsne_mnist}
\end{figure}

\begin{figure}[hbt!]
\centering
\includegraphics[width=1.\linewidth]{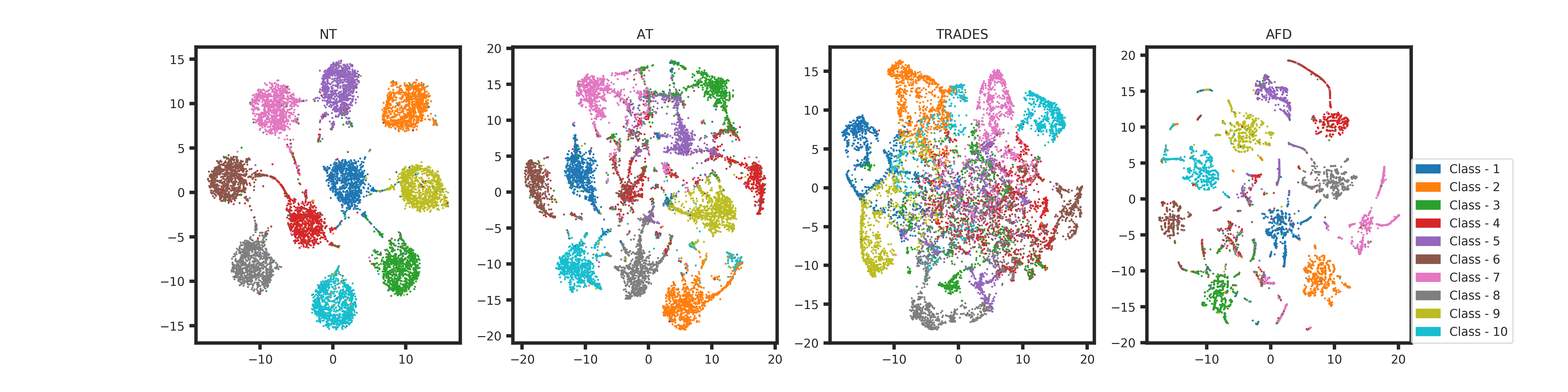}
\includegraphics[width=1.\linewidth]{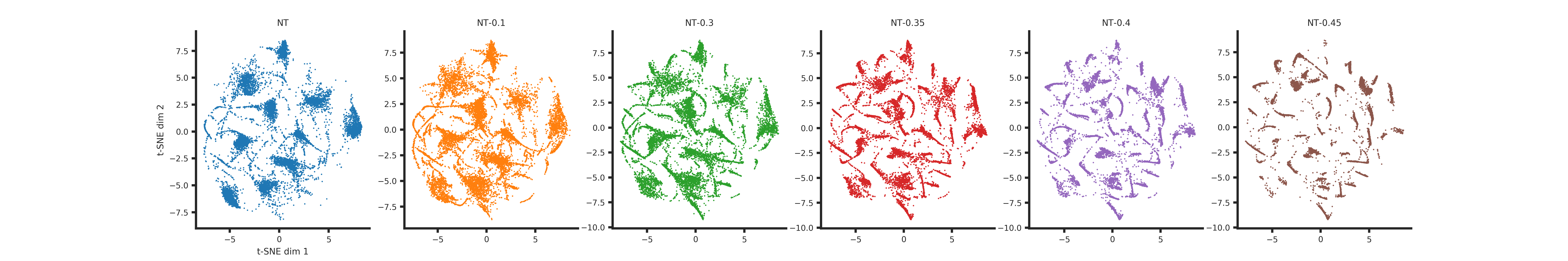}
\includegraphics[width=1.\linewidth]{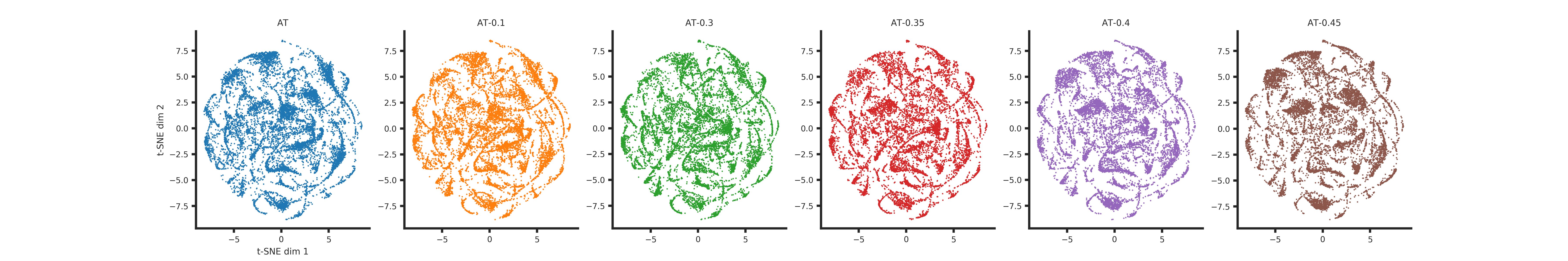}
\includegraphics[width=1.\linewidth]{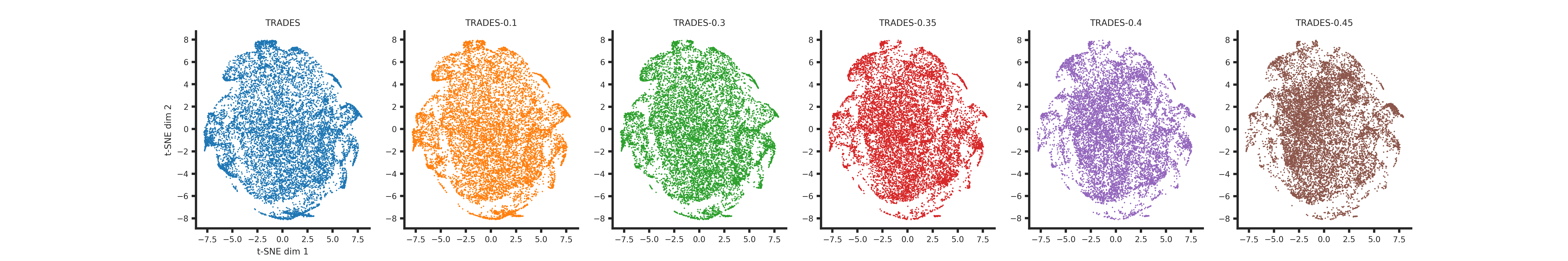}
\includegraphics[width=1.\linewidth]{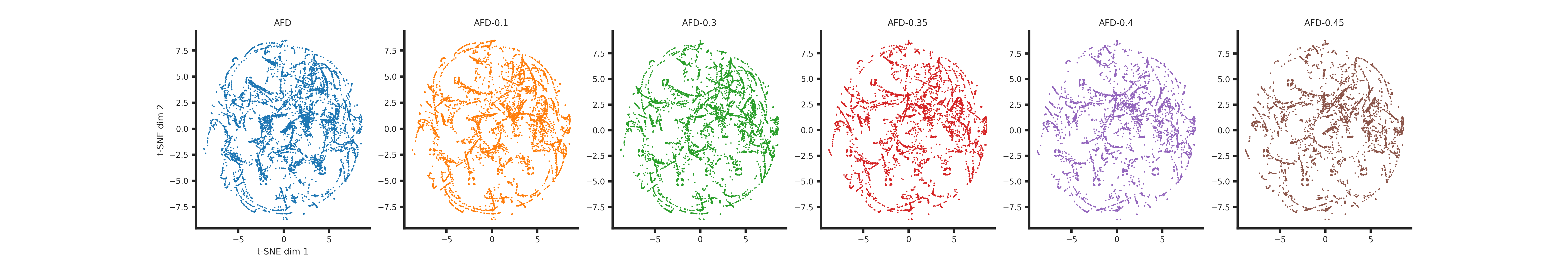}
\caption{Scatter plot of 2-dimensional t-SNE projection \cite{maaten2008visualizing} of the representations derived from training the ResNet5 architecture on CIFAR10 dataset. (top row) t-SNE projection of representations of natural images for networks trained with different methods. Each point corresponds to the representations of one of the images from the CIFAR10 test-set. (rows 2 to 5) t-SNE projection of the representations of natural and adversarial CIFAR10 test-set images. Columns are sorted from left to right with the strength of the perturbation (left-most column corresponds to natural images and right-most column with highest tested perturbation). NT: naturally trained; AT: adversarially trained\cite{madry2017towards}; TRADES: \cite{Zhang2019};AFD: adversarial feature desensitization.}
\label{fig_supp_tsne_cifar10}
\end{figure}

\begin{figure}[hbt!]
\centering
\includegraphics[width=1.\linewidth]{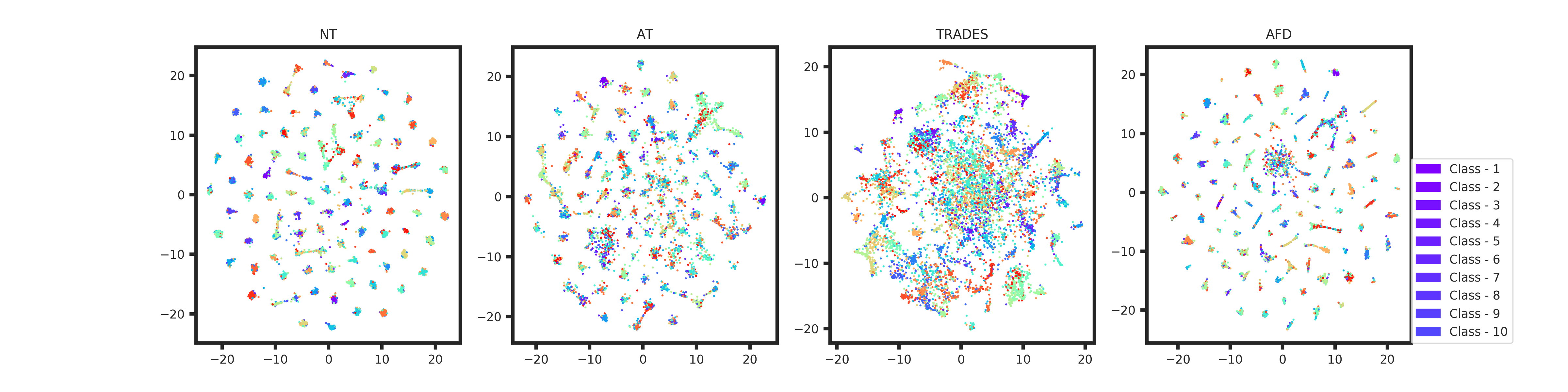}
\includegraphics[width=1.\linewidth]{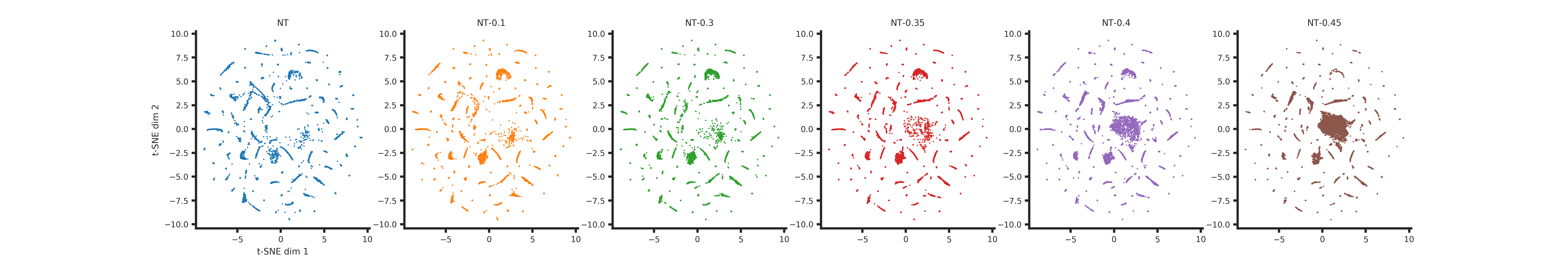}
\includegraphics[width=1.\linewidth]{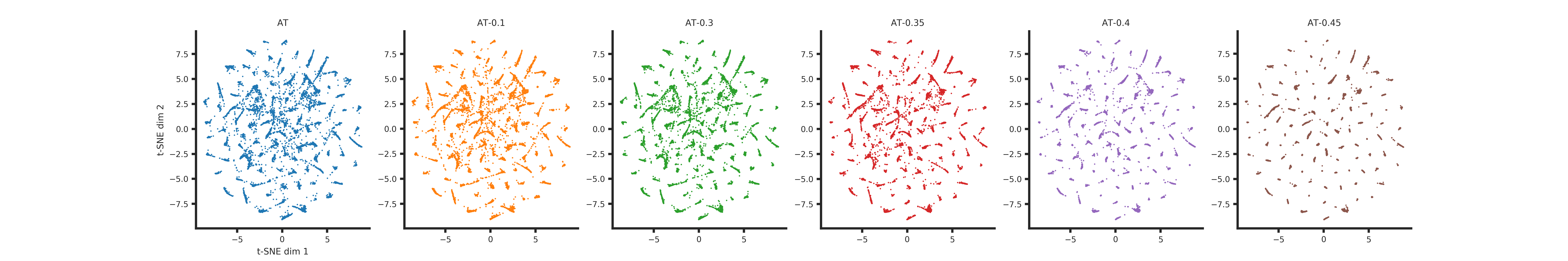}
\includegraphics[width=1.\linewidth]{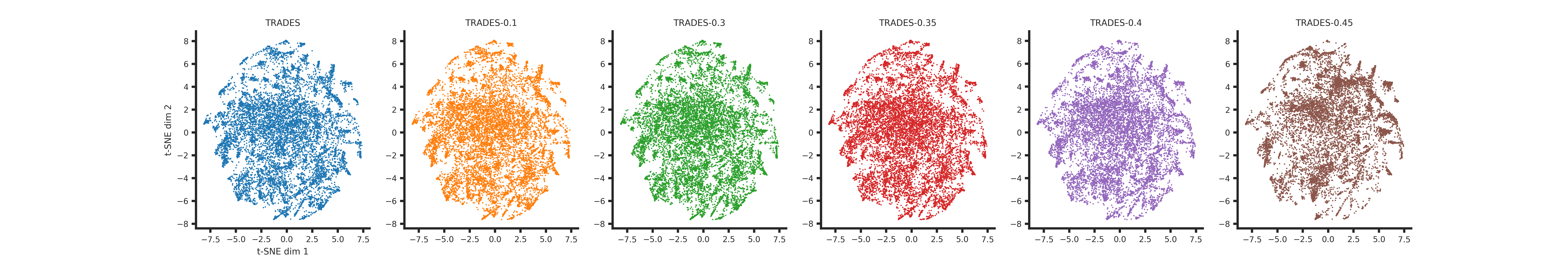}
\includegraphics[width=1.\linewidth]{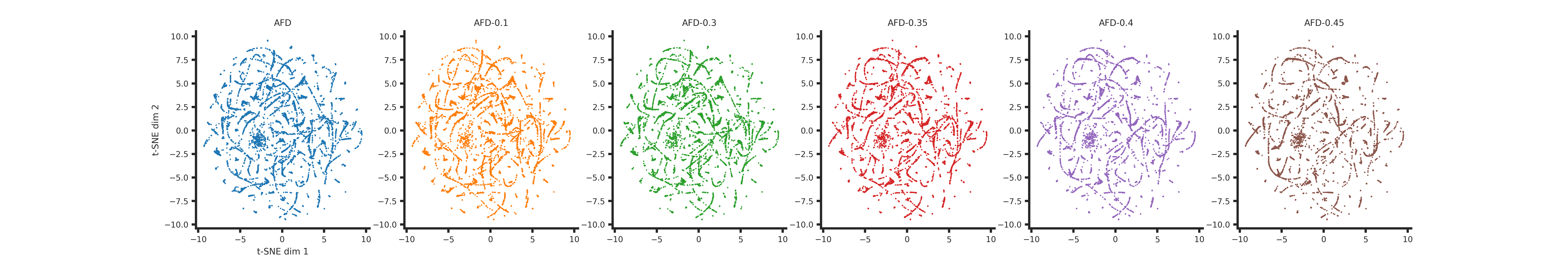}
\caption{Scatter plot of 2-dimensional t-SNE projection \cite{maaten2008visualizing} of the representation derived from training the ResNet5 architecture on CIFAR100 dataset. (top row) t-SNE projection of representations of natural images for networks trained with different methods. Each point corresponds to the representation of one of the images from the CIFAR100 test-set. (rows 2 to 5) t-SNE projection of the representation of the natural and adversarial CIFAR100 test-set images. Columns are sorted from left to right with the strength of the perturbation (left-most column corresponds to natural images and right-most column with highest tested perturbation). NT: naturally trained; AT: adversarially trained \cite{madry2017towards}; TRADES \cite{Zhang2019}; AFD: adversarial feature desensitization.}
\label{fig_supp_tsne_cifar100}
\end{figure}

\clearpage

%%%%%%%%%%%%%%%%%%%%%%%%%%%%%%%%%%%%%%%%%%%%%%%%%%%%%%%%%%%%
\newpage
\section*{Checklist}

%%% BEGIN INSTRUCTIONS %%%
The checklist follows the references.  Please
read the checklist guidelines carefully for information on how to answer these
questions.  For each question, change the default \answerTODO{} to \answerYes{},
\answerNo{}, or \answerNA{}.  You are strongly encouraged to include a {\bf
justification to your answer}, either by referencing the appropriate section of
your paper or providing a brief inline description.  For example:
\begin{itemize}
  \item Did you include the license to the code and datasets? \answerYes{See Section~\ref{gen_inst}.}
  \item Did you include the license to the code and datasets? \answerNo{The code and the data are proprietary.}
  \item Did you include the license to the code and datasets? \answerNA{}
\end{itemize}
Please do not modify the questions and only use the provided macros for your
answers.  Note that the Checklist section does not count towards the page
limit.  In your paper, please delete this instructions block and only keep the
Checklist section heading above along with the questions/answers below.
%%% END INSTRUCTIONS %%%

\begin{enumerate}

\item For all authors...
\begin{enumerate}
  \item Do the main claims made in the abstract and introduction accurately reflect the paper's contributions and scope?
    \answerYes{We provide theoretical motivation for our method in section 3. We then implement and test our approach in section 4.}
  \item Did you describe the limitations of your work?
    \answerYes{We describe the limitations of our work in section 5.}
  \item Did you discuss any potential negative societal impacts of your work?
    \answerNo{Our work is concerned with improving robustness against adversarial attacks. As such, we actually may mitigate some vulnerabilities of machine learning models that could negatively affect society. Arguably, the most likely negative outcome is that providing new defenses may encourage researching stronger attacks.}
  \item Have you read the ethics review guidelines and ensured that your paper conforms to them?
    \answerYes{}
\end{enumerate}

\item If you are including theoretical results...
\begin{enumerate}
  \item Did you state the full set of assumptions of all theoretical results?
    \answerYes{The full set of assumptions are stated in the beginning of Section 3.}
	\item Did you include complete proofs of all theoretical results?
    \answerNo{We do not present any new theories in this work.}
\end{enumerate}

\item If you ran experiments...
\begin{enumerate}
  \item Did you include the code, data, and instructions needed to reproduce the main experimental results (either in the supplemental material or as a URL)?
    \answerYes{The details of the method are fully described in the paper. The code along with model checkpoints are also included in the supplementary material.}
  \item Did you specify all the training details (e.g., data splits, hyperparameters, how they were chosen)?
    \answerYes{This information is provided in the Experiments section.}
	\item Did you report error bars (e.g., with respect to the random seed after running experiments multiple times)?
    \answerNo{We have not reported error bars for experiments on different datasets. The current results include many individual experiments on multiple datasets that were not repeated.}
	\item Did you include the total amount of compute and the type of resources used (e.g., type of GPUs, internal cluster, or cloud provider)?
    \answerYes{The type and number of GPUs used to conduct the experiments were reported in section 4.1.}
\end{enumerate}

\item If you are using existing assets (e.g., code, data, models) or curating/releasing new assets...
\begin{enumerate}
  \item If your work uses existing assets, did you cite the creators?
    \answerYes{This information is provided in the Experiments section.}
  \item Did you mention the license of the assets?  
  \answerYes{In the supplemental material.}
  \item    Did you include any new assets either in the supplemental material or as a URL?
    \answerYes{Yes, code is provided in the supplemental material.}
  \item Did you discuss whether and how consent was obtained from people whose data you're using/curating?
    \answerNo{We did not use any assets that require consent from the creators.}
  \item Did you discuss whether the data you are using/curating contains personally identifiable information or offensive content?
    \answerNo{We only used public benchmarks in our work that do not contain any personally identifiable information or offensive content.}
\end{enumerate}

\item If you used crowdsourcing or conducted research with human subjects...
\begin{enumerate}
  \item Did you include the full text of instructions given to participants and screenshots, if applicable?
    \answerNA{}
  \item Did you describe any potential participant risks, with links to Institutional Review Board (IRB) approvals, if applicable?
    \answerNA{}
  \item Did you include the estimated hourly wage paid to participants and the total amount spent on participant compensation?
    \answerNA{}
\end{enumerate}

\end{enumerate}

%%%%%%%%%%%%%%%%%%%%%%%%%%%%%%%%%%%%%%%%%%%%%%%%%%%%%%%%%%%%

\end{document}